\documentclass[pmlr,twocolumn,10pt]{jmlr}
\usepackage[switch]{lineno}
\usepackage{hyperref}       
\usepackage{booktabs}       
\usepackage{amsfonts}       
\usepackage{nicefrac}       
\usepackage{microtype}      
\usepackage{xcolor}         
\usepackage[font=small]{caption}
\usepackage{subcaption}
\usepackage{enumitem}
\usepackage{titlesec}
\usepackage{mathtools}
\usepackage[english]{babel}
\usepackage{array}
\usepackage[page]{appendix}
\usepackage{tikz}

\newtheorem{assumption}{Assumption}

\jmlrvolume{287}
\jmlryear{2025}
\jmlrsubmitted{} 
\jmlrpublished{} 
\jmlrworkshop{Conference on Health, Inference, and Learning (CHIL) 2025}

\DeclareMathOperator*{\ce}{\mathcal{L}_{CE}}
\DeclareMathOperator*{\ol}{\mathcal{L}_{Y}}

\DeclareMathAlphabet{\pazocal}{OMS}{zplm}{m}{n}
\newcommand{\unif}{\pazocal{U}}

\newcommand*\circled[1]{\tikz[baseline=(char.base)]{
            \node[shape=circle,draw,inner sep=2pt] (char) {#1};}}

\newcommand{\methodname}{LobsterNet}
\title[CFD for CATEA Estimation Under non-adherence]{Conditional Front-door Adjustment for Heterogeneous Treatment Assignment Effect Estimation  Under Non-adherence}

\author{%
 \Name{Winston Chen} \Email{chenwt@umich.com}\\
 \Name{Trenton Chang} \Email{ctrenton@umich.com}\\
 \Name{Jenna Wiens} \Email{wiensj@umich.com}\\
 \addr University of Michigan, Ann Arbor, MI, USA
}

\begin{document}

\maketitle

\begin{abstract}
Estimates of heterogeneous treatment assignment effects can inform treatment decisions.
Under the presence of non-adherence (\emph{e.g.}, patients do not adhere to their assigned treatment), both the standard backdoor adjustment (SBD) and the conditional front-door adjustment (CFD) can recover unbiased estimates of the treatment assignment effects.
However, the estimation variance of these approaches may vary widely across settings, which remains underexplored in the literature. 
In this work, we demonstrate theoretically and empirically that CFD yields lower-variance estimates than SBD when the true effect of treatment assignment is small (\textit{i.e.}, assigning an intervention leads to small changes in patients' future outcome). 
Additionally, since CFD requires estimating multiple nuisance parameters, we introduce \methodname, a multi-task neural network that implements CFD with joint modeling of the nuisance parameters. 
Empirically, \methodname\ reduces estimation error across several semi-synthetic and real-world datasets compared to baselines, even when the true effect of treatment assignment is large. 
Our findings suggest CFD with shared nuisance parameter modeling can improve treatment assignment effect estimation under non-adherence.

\end{abstract}

\paragraph*{Data and Code Availability}
This paper uses IHDP~\cite{hill2011bayesian} and AMR-UTI~\cite{oberst2020amr} datasets. 
Our training and evaluation codes are available at https://github.com/MLD3/LobsterNet.

\paragraph*{Institutional Review Board (IRB)}
This work is not regulated as human subjects research since the data are de-identified and publicly available.

\section{Introduction}\label{sec:introduction}
We study the estimation of heterogeneous treatment assignment effects under treatment non-adherence from observational data.
Treatment non-adherence arises when some individuals' treatment intake does not match their treatment assignment (\textit{e.g.}, patients fail to take their prescribed medication).
This can occur in many real-world applications, such as public health~\citep{tressler2019interventions} and online advertising~\citep{gordon2019comparison}.

In many settings, decision-makers (\textit{e.g.}, clinicians) can only intervene by assigning treatment and cannot enforce adherence. 
To facilitate decisions about individual treatment assignments, decision-makers must estimate the heterogeneous treatment \textit{assignment} effect, which accounts for both the adherence probability and standard treatment effect.
We refer to this quantity as the \textit{conditional average treatment effect of the assignment (CATEA)}, which corresponds to the effect of treatment \emph{assignment} on the outcome of interest (\emph{e.g.}, disease progression)\footnote{Our estimand differs from the standard conditional average treatment effect (CATE), \emph{i.e.}, the effect of adhering to the assigned treatment.}.

To estimate CATEAs, one can use the standard backdoor adjustment (SBD)~\citep{montori2001intention, gupta2011intention}, which estimates potential outcomes while controlling for confounders. 
However, SBD ignores treatment intake information, which is often available. 
We propose to leverage conditional front-door adjustment (CFD)~\citep{pearl2009causality,xu2023causal} as an alternative for estimating the CATEA that leverages treatment intake.
CFD works by combining estimates of three \emph{nuisance parameters:} (1) the probability of treatment assignment, (2) the probability of treatment intake given assignment, and (3) the outcome given treatment assignment and intake via an adjustment formula to form CATEA estimates.

\begin{figure*}[t]
     \centering
     \subfigure[Standard backdoor Adjustment (SBD)]{\includegraphics[width=0.4\linewidth,alt={sbd}]{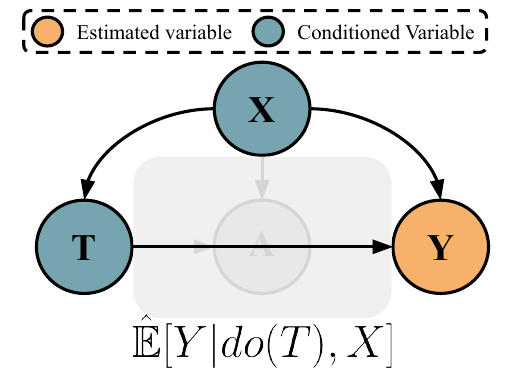}\label{fig:overview:sbd}}
    \subfigure[Conditional Front-door Adjustment (CFD)]{\includegraphics[width=0.56\linewidth,alt={cfd}]{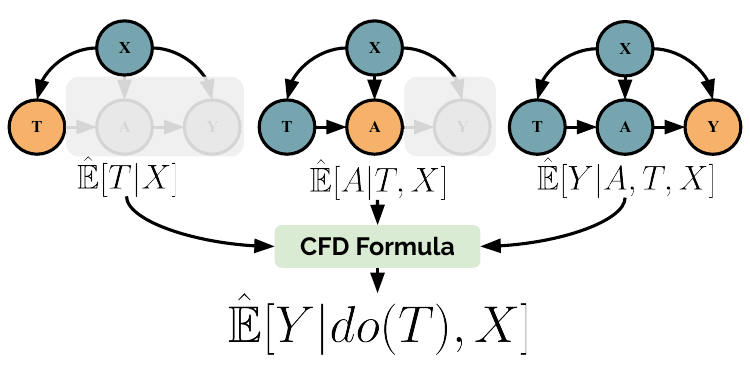}\label{fig:overview:cfd}}
    \caption{Overview of SBD and CFD in causal graphs. Variables $X$, $T$, $A$, and $Y$ indicate confounders, treatment assignment, intake, and outcome. (A) SBD directly estimates the potential outcomes. (B) CFD first estimates three types of nuisance parameters and recovers the potential outcomes by combining the estimates via the CFD formula.}
    \label{fig:overview}
\end{figure*}

The error associated with CATEA estimates decomposes into bias and variance~\citep{domingos2000unified}. 
In this work, we first prove that given sufficient identification assumptions, both CFD and SBD produce unbiased CATEA estimates (Section~\ref{sec:background}).
Thus, one should choose the estimator with the lower variance. 
However, whether CFD or SBD has lower variance is underexplored in the literature (see Section~\ref{sec:related}). 
Intuitively, using treatment intake information in CFD should reduce variance.
However, since CFD estimates more nuisance parameters, variance from each estimate could inflate the overall variance. 
Due to this tension, it is unclear when CFD will yield lower-variance estimates than SBD. 

Thus, we compare the asymptotic variance of CFD to that of SBD and show that CFD has lower variance than SBD when the true CATEA is small.
Small CATEA is common in real-world treatment assignments~\citep{ioannidis2005most, gu2016review, vivalt2020much}, highlighting the practical benefits of CFD.

Building on our theoretical analysis, we propose \methodname, a multi-task neural network architecture that jointly estimates nuisance parameters required by CFD. 
In particular, \methodname\ leverages shared representations between nuisance parameters to improve CFD's performance in finite sample settings. 
In summary, our contributions are as follows:
\begin{itemize}
  \item to the best of our knowledge, we are the first to propose CFD for CATEA estimation and prove its unbiasedness under sufficient assumptions,
  \item we compare the asymptotic variance of  CFD to SBD in CATEA estimation, showing that CFD improves variance under small CATEA,
  \item we propose \methodname, a novel CFD implementation with joint nuisance parameter modeling, improving the empirical performance of CFD compared to existing implementations.
\end{itemize}

\section{Related Work}\label{sec:related}

Under non-adherence, the effect of treatment intake on the outcome, known as the conditional average treatment effect (CATE), is often estimated to understand the impact of treatment intake (\textit{e.g.}, a drug's effectiveness)~\citep{marcus2001estimating,nagelkerke2000estimating,bargagli2022heterogeneous, miladinovic2011instrumental}.
While CATE estimation is important, estimating the treatment assignment effect (\emph{i.e.}, CATEA) could be independently useful in scenarios where decision-makers can only assign treatments but cannot enforce treatment intake
(\textit{e.g.}, encouragement designs~\citep{hirano2000assessing}).

In this work, we are interested in using observational data to estimate CATEAs~\citep{rahier2021individual}, which account for both the individual's adherence probability and the effect of the treatment.  
Knowledge of CATEA can help optimize decisions when assigning treatments. For example, if an individual is unlikely to adhere to a prescription, the overall effect of \textit{assigning} the drug would remain small even for a highly effective drug.

To estimate the CATEA, prior work has applied SBD~\citep{montori2001intention, gupta2011intention}. 
However, such approaches ignore the treatment intake, an intermediate variable that can improve the precision of effect estimates.
Such variables have been used outside the treatment non-adherence setting~\citep{fisher1928statistical, cox1960regression}. 
Most recently, \cite{rahier2021individual} demonstrated that combining CATE estimates with estimated intake probability results in better CATEA estimates than SBD.
However, their approach does not leverage CFD and only applies to one-sided non-adherence (defined in Section~\ref{sec:background:non-adherence}) with randomized control trial (RCT) data. 
In contrast, we leverage the CFD formula to enable CATEA estimation with both one-sided and two-sided non-adherence using observational data.

The CFD formula provides a flexible approach to combine an intermediate variable in causal effect estimation~\citep{pearl2009causality,xu2023causal}.
It has previously been proposed as an alternative to SBD in scenarios outside of treatment non-adherence~\citep{hill2015measuring,bellemare2020paper}.
However, the relative performance of SBD and CFD is poorly understood. \cite{glynn2018front} studied their estimation bias subject to unobserved confounding and demonstrated that CFD has a lower bias in a broad class of applications than SBD.
Similar to our work, \cite{gupta2021estimating} and  \cite{ramsahai2012supplementary} compared SBD and CFD when both approaches are unbiased.
Their findings suggest that under linear Gaussian data generation processes (DGPs), either approach could outperform the other in estimation variance.
In contrast, we compare the estimation variance of SBD and CFD without assuming a specific DGP and characterize scenarios where CFD outperforms SBD.

Beyond ignoring the treatment intake, SBD is also limited by sample inefficiency and sensitivity to model misspecification.
As a result, alternative methods such as the DR-learner~\citep{kennedy2023towards}, targeted maximum likelihood estimation (TMLE)~\citep{van2006targeted}, and double machine learning (DML)~\citep{chernozhukov2017double} have been proposed to augment SBD with more robust and efficient estimators.
However, we focus on comparing CFD and SBD implemented with standard supervised learning. 
Efficient estimators could potentially improve both CFD and SBD, but we aim to first understand how the two approaches perform with standard estimators.

\section{Background}\label{sec:background}
Here, we discuss preliminaries for CATEA estimation under treatment non-adherence.

\subsection{Problem setup}\label{sec:background:setup}
Consider an observational dataset of i.i.d. samples $\{(x_i, t_i, a_i, y_i)\}_{i=1}^{n}$ drawn from a joint distribution $\mathcal{P}(X, T, A, Y)$ described by the causal model in Figure~\ref{fig:overview}. $X$, $T$, $A$, and $Y$ denote confounders, binary treatment assignment, binary treatment intake, and the outcome of interest, respectively.

For every individual $i$, we define $A(t_i, x_i)$ as their potential treatment intake under different treatment assignments, $t_i \in \{0, 1\}$, and $Y(a_i, x_i)$ as their potential outcome under different treatment, intake $a_i \in \{0, 1\}$. 
The fundamental problem of estimating the treatment assignment effect under non-adherence is that we only observe the outcome under one of each individual's treatment assignment and intake combinations.
The observed outcome can be expressed as $y_i=Y(A(t_i, x_i), x_i)$.
We aim to estimate CATEA, $\Phi(x_i)$, defined as:
\begin{align*}
  \Phi(x_i) \coloneqq &\mathbb{E}[Y | do(T=1),x_i] - \mathbb{E}[Y | do(T=0),x_i]
\end{align*}

\subsection{Non-adherence settings}\label{sec:background:non-adherence}
We consider two standard non-adherence settings~\citep{gordon2019comparison}: one-sided and two-sided non-adherence. 
Under one-sided non-adherence, only individuals assigned the treatment ($t_i=1$) can be non-adherent ($a_i \neq t_i$). Individuals not assigned the treatment ($t_i=0$) always adherer; \emph{i.e.}, $A(t_i=0, x_i)=0\ \forall\ i$. 
Under two-sided non-adherence, treatment intake ($A$) may deviate arbitrarily from the treatment assignment ($T$). 
In this paper, we use $\Phi_{\mathrm{os}}(\cdot)$ and $\Phi_{\mathrm{ts}}(\cdot)$ to denote CATEA estimates under one-sided and two-sided non-adherence, respectively.

\subsection{SBD-based approach}\label{sec:background:SBD}
The SBD-based approach ignores individuals' treatment intake ($A$) and only considers treatment assignment ($T$), outcome ($Y$), and confounders ($X$). Its CATEA estimations under both one-sided and two-sided non-adherence are defined as follows:
\begin{align*}
  \hat{\Phi}^{\text{SBD}}(x_i) = \hat{\mathbb{E}}[Y | T=1, x_i] - \hat{\mathbb{E}}[Y | T=0, x_i]
\end{align*}
$\hat{\mathbb{E}}[\cdot]$ indicates the empirical estimate of the expectation. In practice, SBD is often implemented with meta-learners such as T-learner~\citep{kunzel2019metalearners}, which estimates $\hat{\mathbb{E}}[Y | T=0, x_i]$ and $\hat{\mathbb{E}}[Y | T=1, x_i]$ with two separate models. When all the potential confounders between $T$, $A$, and $Y$ are included in $X$, SBD provides unbiased CATEA estimation because it removes any effect from confounders by conditioning them. (Proof of SBD's unbiasedness is in Appendix~\ref{sec:appendix:id-proofs:sbd})

\subsection{CFD-based approach}\label{sec:background:CFD}
Our proposed CFD-based approach first estimates three sets of nuisance parameters from the data:
\begin{align*}
  \hat{\pi}(x_i) &= \hat{\mathbb{E}}[T| x_i] \\
  \hat{A}(t_i, x_i) &= \hat{\mathbb{E}}[A |T=t_i, x_i] \\
  \hat{Y}(a_i, t_i, x_i) &= \hat{\mathbb{E}}[Y |A=a_i, T=t_i, x_i]
\end{align*}

Each of these nuisance parameters can be estimated via standard supervised learning techniques. 
CATEA is then recovered by combining these estimates with the CFD formula~\citep{pearl2009causality,xu2023causal}. Under one-sided non-adherence, the formula is defined as:
\begin{align*}
    \hat{\Phi}^{\text{CFD}}_{\text{os}} = 
  \bigg[&\left(\hat{Y}(1,1,x_i) - \hat{Y}(0,0,x_i)\right)(1-\hat{\pi}(x_i)) + \\
  &\left(\hat{Y}(1,1,x_i) - \hat{Y}(0,1,x_i)\right)\hat{\pi}(x_i)\bigg]\hat{A}(1, x_i)
\end{align*}

The CFD formula under two-sided non-adherence is defined as follows:
\begin{align*}
  &\hat{\Phi}^{\text{CFD}}_{\text{ts}}(x_i) = \bigg[\left(\hat{Y}(1,0,x_i) - \hat{Y}(0,0,x_i)\right)(1-\hat{\pi}(x_i)) + \\
  &\left(\hat{Y}(1,1,x_i) - \hat{Y}(0,1,x_i)\right)\!\hat{\pi}(x_i)\bigg]\!\cdot\! \left[\hat{A}(1, x_i)-\hat{A}(0, x_i)\right]
\end{align*}

Leveraging \cite{pearl2009causality}'s do calculus, we prove that CFD leads to unbiased CATEA estimates under both one-sided and two-sided non-adherence (details in Appendix~\ref{sec:appendix:id-proofs:cfd-two-sided} and \ref{sec:appendix:id-proofs:cfd-one-sided}).
Intuitively, these formulas compute the effect of intake ($a_i$) on the outcome ($y_i$) under different assignments ($t_i$) and then combine them via a weighted sum, where the weights are defined by the probability of treatment assignment ($\hat{\pi}(\cdot)$) and the effect of treatment assignment on intake. 
CFD only uses treatment intake to estimate nuisance parameters; at test time, both SBD and CFD only use confounders (X).
The derivations for the above formulas are in Appendix~\ref{sec:appendix:cfd-derivation}. 

\subsection{Identifiability assumptions}\label{sec:background:id-assumptions}
To make the CATEA identifiable under our problem setup, we assume consistency and positivity, as well as no unobserved confounders between any pair of $T$, $A$, and $Y$ (ignorability), which are standard assumptions in causal effect estimation~\citep{pearl2009causality}. 
Additionally, we assume full mediation, a standard assumption under treatment non-adherence~\citep{rahier2021individual, gordon2019comparison}. 
See Appendix~\ref{sec:appendix:assumptions} for detailed definitions of these assumptions.

\section{Theoretical Analysis}\label{sec:theory}
In this section, we explore when the CFD-based approach is guaranteed to have lower estimation variances than the SBD-based approach. 

\subsection{Setup}\label{sec:theory:setup}
We compare the asymptotic variances of $\hat{\Phi}^{\text{SBD}}$ and $\hat{\Phi}^{\text{CFD}}$ in the single-stratum setting, which focuses on estimation for a single value $x_0 \in X$, for which we assume to observe $n$ samples $\{(x_0, t_i, a_i, y_i)\}_{1\le i\le n}$. We simplify our notation as follows:
\begin{align*}
    \begin{split}
        A_{t=0}&\sim P\left(a_i | t_i=0, x_0\right) \\
        Y_{a=0}&\sim P\left(y_i | a_i=0, x_0\right)
    \end{split}
    \begin{split}
        A_{t=1}&\sim P\left(a_i | t_i=1, x_0\right) \\
        Y_{a=1}&\sim P\left(y_i | a_i=1, x_0\right)
    \end{split}
\end{align*}

$\mathbb{E}(\cdot)$ and $\mathbb{V}(\cdot)$ denote the expectation and variance of the corresponding variable.
We also define $\Delta_A$ and $\Delta_Y$ as the effect of assignment on intake and the effect of intake on the outcome. 
They are decompositions of the CATEA (\textit{i.e.}, $\Phi(x_o) = \Delta_A\Delta_Y$):
\begin{align*}
    \Delta_A &= \mathbb{E}\left(A_{t=1}\right)-\mathbb{E}\left(A_{t=0}\right)\\
    \Delta_Y &= \mathbb{E}\left(Y_{a=1}\right)-\mathbb{E}\left(Y_{a=0}\right)
\end{align*}

Because CFD requires estimating outcomes conditioned on each assignment and intake combination, we define $\rho$ as the minimum probability that the individual experiences one of the possible combinations. 
Formally, $\rho$ is defined as below:
\begin{align*}
    \rho &= \begin{cases}
    \min\limits_{\substack{t',a' \in \{0,1\} \\ (a', t') \neq (1,0)}}\left\{ P(a', t' \mid x_0) \right\} 
    & \parbox[t]{5cm}{if one-sided \\ non-adherence} \\[2mm]
    \min\limits_{t',a' \in \{0,1\}}\left\{ P(a', t' \mid x_0) \right\} 
    & \parbox[t]{5cm}{if two-sided \\ non-adherence}
    \end{cases}
\end{align*}
Later, $\rho$ will help simplify the bound on the variance of CFD by representing the fewest observations among all possible treatment assignment/intake combinations.

\begin{figure*}
  \centering
  \includegraphics[width=1.0\linewidth,alt={variance}]{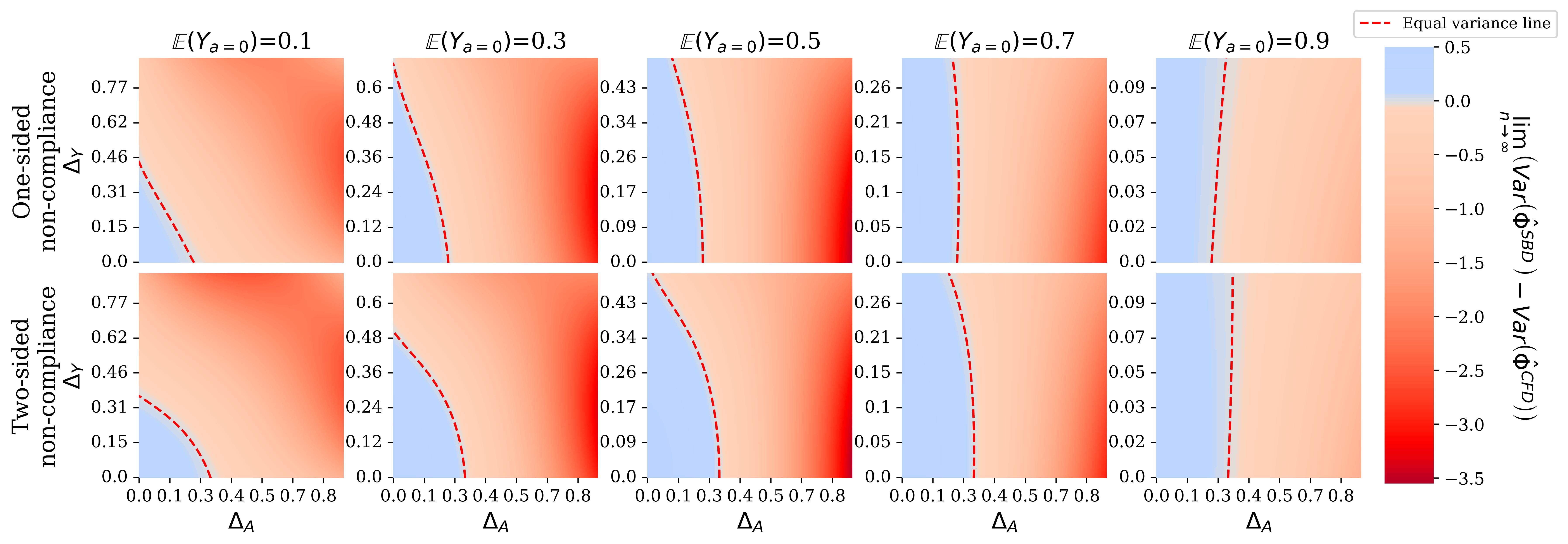}\vspace{-2mm}
  \captionof{figure}{Numerical visualization of lower bounds on CFD's asymptotic variance reduction over SBD.
  More blue colors indicate greater variance reduction by CFD. 
  The red dotted line indicates contours where the reduction equals zero. 
  CFD is guaranteed to reduce variance when the effect size (either $\Delta_A$ or $\Delta_Y$) is small.}
  \label{fig:variance}
\end{figure*}

\subsection{Asymptotic Variance Bounds}\label{sec:theory:bounds}
Here, we present asymptotic variance \emph{lower} bounds for SBD-based CATEA estimation and \emph{upper} bounds for CFD-based CATEA estimation.
Our derivation follows the analysis framework proposed by~\cite{rahier2021individual}, and are described in Appendix~\ref{sec:appendix:variance:bound}.

\begin{proposition}
Under one-sided and two-sided non-adherence, the asymptotic variance of SBD-based CATEA estimation is lower-bounded by:
\begin{align*}
    &\lim_{n\rightarrow \infty}n\mathrm{Var}\left(\hat{\Phi}^{\mathrm{SBD}}_{\mathrm{os}}\right) > \mathbb{V}\left(Y_{a=0}\right)\big(2-\mathbb{E}\left(A_{t=1}\right)\big) + \\
    &\ \mathbb{V}\left(Y_{a=1}\right)\mathbb{E}\left(A_{t=1}\right) + \mathbb{V}\left(A_{t=1}\right)\left(\Delta_Y\right)^2 \\
    &\lim_{n\rightarrow \infty}n\mathbb{V}(\hat{\Phi}^{\mathrm{SBD}}_{\mathrm{ts}}) >  \mathbb{V}\left(Y_{a=0}\right)\big(2-\mathbb{E}\left(A_{t=0}\right)-\mathbb{E}\left(A_{t=1}\right)\big) \\ 
     &\ + \mathbb{V}\left(Y_{a=1}\right)\big(\mathbb{E}\left(A_{t=0}\right) + \mathbb{E}\left(A_{t=1}\right)\big) \\
     &\ + \big(\mathbb{V}\left(A_{t=0}\right) + \mathbb{V}\left(A_{t=1}\right)\big)\left(\Delta_Y\right)^2. \\
\end{align*}
\end{proposition}\label{prop:two-sided-cfd-bounds}

\begin{proposition}
Under one-sided and two-sided non-adherence, the asymptotic variance of CFD-based CATEA estimation is upper-bounded by:
\begin{align*}
    &\lim_{n\rightarrow \infty}n\mathrm{Var}(\hat{\Phi}^{\mathrm{CFD}}_{\mathrm{os}}) < \frac{1}{\rho}\bigg(\big(\mathbb{V}\left(Y_{a=0}\right) + \mathbb{V}\left(Y_{a=1}\right)\big) \Delta_A^2 \\ 
    &\quad +  \mathbb{V}\left(A_{t=1}\right)\Delta_Y^2\bigg) \\
    &\lim_{n\rightarrow \infty}n\mathrm{Var}(\hat{\Phi}^{\mathrm{CFD}}_{\mathrm{ts}}) < \frac{1}{\rho}\bigg(\big(\mathbb{V}\left(Y_{a=0}\right) + \mathbb{V}\left(Y_{a=1}\right)\big) \Delta_A^2 \\ 
    &\quad + \big(\mathbb{V}\left(A_{t=0}\right) + \mathbb{V}\left(A_{t=1}\right)\big)\Delta_Y^2\bigg).
\end{align*}
\end{proposition}

\begin{figure*}[t]
     \centering
     \subfigure[T-learner for SBD]{\includegraphics[width=0.48\linewidth,alt={tlearner_sbd}]{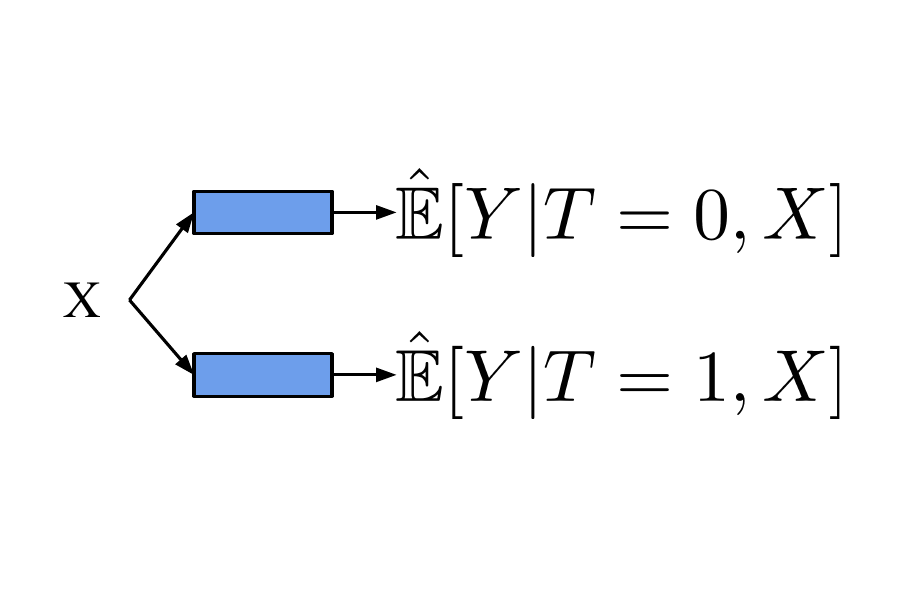}\label{fig:tlearner:sbd}}
    \subfigure[T-learner for CFD]{\includegraphics[width=0.48\linewidth,alt={tlearner_cfd}]{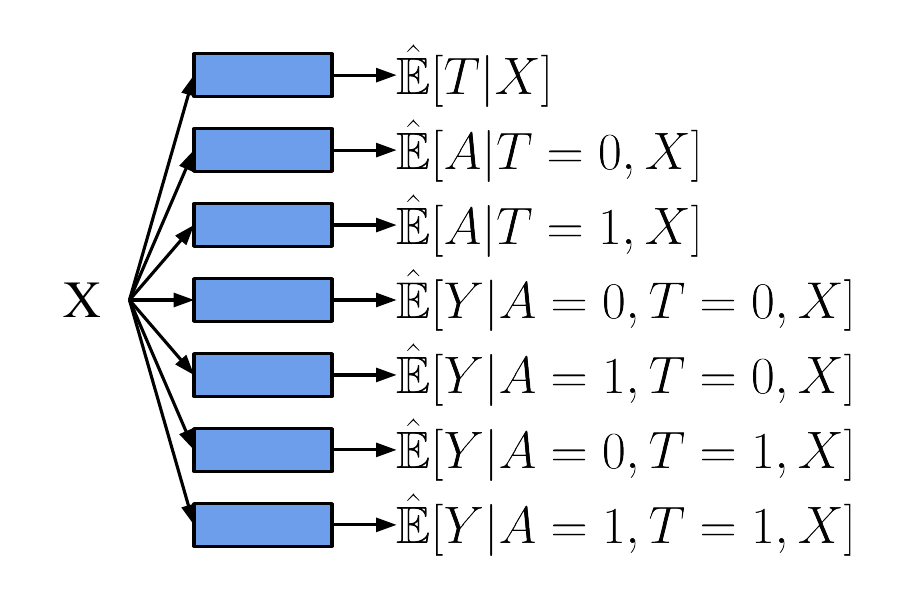}\label{fig:tlearner:cfd}}
    \caption{Illustrations of the T-learner method for implementing both SBD and CFD. Because CFD requires estimating more nuisance parameters than SBD, we extend the T-learner by introducing additional independent models.}
    \label{fig:tlearner}
\end{figure*}

\subsection{Analytical Comparison}
\label{sec:theory:analytical}
From the upper bounds on CFD's variance and lower bounds on SBD's variance we can obtain a \textit{lower} bound on CFD's variance reduction over SBD (\textit{i.e.}, $\lim\limits_{n\rightarrow \infty}\left(n\text{Var}\left(\Phi^{\text{SBD}}\right) - n\text{Var}\left(\Phi^{\text{CFD}}\right)\right)$).

To simplify this comparison, we assume that each potential outcome or intake shares the same variance, which we denote with $\mathrm{V}_\mathrm{Y}$ and $\mathrm{V}_\mathrm{A}$ respectively:
\begin{align*}
    \mathbb{V}\left(Y_{a=0}\right) &= \mathbb{V}\left(Y_{a=1}\right) = \mathrm{V}_\mathrm{Y}, \\
    \mathbb{V}\left(A_{t=0}\right) &= \mathbb{V}\left(A_{t=1}\right) = \mathrm{V}_\mathrm{A}.
\end{align*}
This is reasonable because real-world potential outcomes often share the same generation processes, leading to similar variances~\citep{curth2021inductive}.
Then, we obtain the following simplified variance reduction lower-bound (details in Appendix~\ref{sec:appendix:variance:compare}):

\begin{proposition}
    CFD's asymptotic variance reduction over SBD is lower-bounded by:
    \begin{align*}
         &\lim_{n\rightarrow \infty}n\left(\mathrm{Var}(\hat{\Phi}^{\mathrm{SBD}})-\mathrm{Var}(\hat{\Phi}^{\mathrm{CFD}})\right)  > \\
         &\quad \mathcal{O}\left(V_Y-\frac{V_Y\Delta_A^2}{\rho} - \frac{(1-\rho)V_A\Delta_Y^2}{\rho}\right).
    \end{align*}
\end{proposition}

This variance reduction bound applies to both one-sided and two-sided non-adherence.
Intuitively, \textbf{decreasing either $\Delta_A^2$ and $\Delta_Y^2$ will increase the overall bound, giving CFD a greater advantage over SBD}.
Additionally, $\rho$ also influences the bound.
Because $0<\rho<1$, \textbf{decreasing $\rho$ will reduce the overall bound, making CFD less advantageous}.

\subsection{Numerical Comparison}\label{sec:theory:visual}
In Figure~\ref{fig:variance}, we visualize the variance reduction lower bound in settings where variance may not be equal between potential outcomes and inputs.
Without loss of generality, we consider that $\Delta_A, \Delta_Y\geq0$ and set $\rho=0.1$. $\mathbb{E}\left(Y_{a=0}\right)$ is set to five different values between 0 and 1. 
Under one-sided non-adherence, $\mathbb{E}\left(A_{a=0}\right)$ is zero by definition, and under two-sided non-adherence, $\mathbb{E}\left(A_{a=0}\right)$ is set to 0.1.

The smaller the reduction lower bound (more blue in Figure~\ref{fig:variance}), the larger is the gap between the CFD variance versus the SBD variance in favor of CFD.
Across all values of $\mathbb{E}\left(Y_{a=0}\right)$, CFD-based estimation has increasing advantage as $\Delta_{A}$ and $\Delta_{Y}$ decrease. Intuitively, CFD enjoys an advantage over SBD when the true CATEA is small, \emph{i.e.}, when either $\Delta_A$ or $\Delta_Y$ is small, supporting our analytical finding.
As $\mathbb{E}\left(Y_{a=0}\right)$ increases, the area where CFD is guaranteed to achieve lower variances also increases. This is because the feasible value of $\Delta_Y$ decreases as $\mathbb{E}\left(Y_{a=0}\right)$ increases, reducing CATEA and giving more CFD advantages. 
More details about the numerical visualization can be found in Appendix~\ref{sec:appendix:num_viz}.

\subsection{Limitations of CFD}\label{sec:theory:limits}
Our analysis also highlights regions where CFD does not guarantee a reduced variance compared to SBD. 
For example, the advantage of CFD over SBD in terms of asymptotic variance diminishes when $\rho$ (the minimum probability for individuals to experience one of the possible assignment and intake combinations) is small, as shown via an analytical comparison (Section~\ref{sec:theory:analytical}).
Furthermore, Figure~\ref{fig:variance} assumes a conservative $\rho$ value (0.1), making it hard to guarantee CFD's advantage under many settings.
While lower CFD variance can be achieved with greater $\rho$, this requires individuals having similar probabilities for experiencing different assignment and intake combinations, which may not hold in practice~\citep{khan2024treatment}.

To mitigate CFD's need for having sufficient samples across all assignment and intake combinations, we draw inspiration from the hypothesis that the true potential outcome functions may have shared structures in practice~\citep{curth2021inductive}. 
In Section~\ref{sec:method}, we introduce a multi-task neural network architecture designed to leverage shared representations when learning CFD's nuisance parameters. 
When nuisance parameters are learned via shared representations, $\rho$ is virtually increased because every sample contributes to estimating nuisance parameters conditioned on each assignment and intake combination.

\begin{figure*}
  \centering
  \includegraphics[width=1.0\linewidth,alt={lobster}]{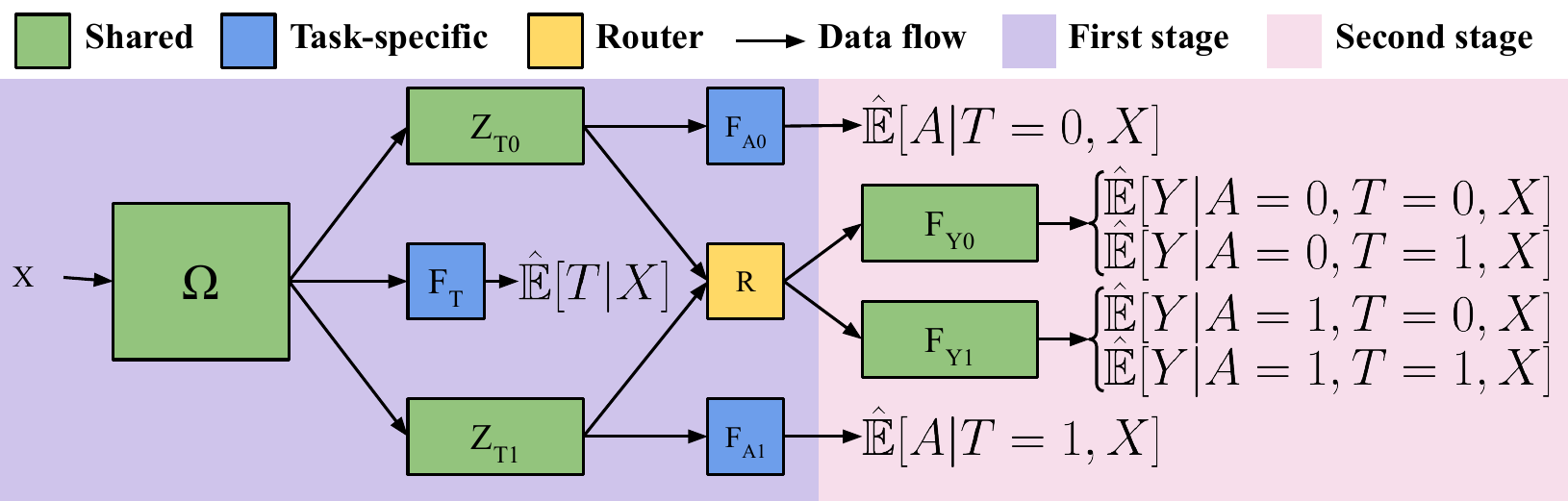}
  \captionof{figure}{Illustration of our proposed \methodname\ architecture,which jointly models all required nuisance parameters}\vspace{-4mm}
  \label{fig:cfd:lobster}
\end{figure*}

\section{Implementing CFD in Practice}\label{sec:method}
We first discuss implementing CFD by extending the T-learner, an existing method for implementing SBD. 
Then, we introduce \methodname, a multi-task neural network architecture tailored for CFD-based CATEA estimation.

\subsection{CFD with T-learner}
T-learner (short for Two-model learner) is a meta-algorithm commonly used to implement SBD~\citep{kunzel2019metalearners}.
It leverages two independent models to estimate potential outcomes required by SBD (Figure~\ref{fig:tlearner:sbd}).
We implement CFD by extending the T-learner. 
Because CFD requires estimating more than two nuisance parameters, we introduce additional models to independently estimate all nuisance parameters (Figure~\ref{fig:tlearner:cfd}).
We provide implementation details of CFD with T-learner in Appendix Section~\ref{sec:appendix:implementation:tlearner}.
Despite its popularity, the T-learner may be sample inefficient because it does not leverage potential similarities between estimates. 
When applied to implement CFD, the T-learner approach can have high variance when certain nuisance parameters have a small number of observed samples.
Thus, we propose a new approach to overcome this challenge by estimating the nuisance parameters jointly.

\subsection{CFD with \methodname}
We propose \methodname\ to simultaneously estimate all nuisance parameters required for CFD via shared representations and a joint training objective. 
\methodname\ extends DragonNet~\citep{shi2019adapting} by introducing a two-stage structure: stage one models treatment assignment and intake, and stage two leverages the representations learned in stage one to predict potential outcomes (Figure~\ref{fig:cfd:lobster}). However, in contrast to DragonNet, our proposed approach enables sharing across all CFD-required estimation tasks.

The first stage starts with a backbone network, $\Omega(\cdot)$, which produces a shared representation to predict treatment assignment and 
conditional treatment intake via networks: $F_T(\cdot)$, $F_{A0}(Z_{T0}(\cdot))$, and $F_{A1}(Z_{T1}(\cdot))$. In the second stage, \methodname\ predicts outcomes conditioned on both treatment assignment and intake. 
The conditioning on intake is achieved by running predictions with separate networks, $F_{Y0}(\cdot)$ and $F_{Y1}(\cdot)$. 
To condition on different assignments, \methodname\ leverages a router module, $R(\cdot)$, which selects output from either $Z_{T0}(\cdot)$ or $Z_{T1}(\cdot)$ as the input to $F_{Y0}(\cdot)$ and $F_{Y1}(\cdot)$.

Let $\hat{T}(x)$, $\hat{A}(t, x)$ and $\hat{Y}(t, a, x)$ denote \methodname's 
estimates of treatment assignment, intake, and outcome, respectively. We optimize it  by minimizing:
\begin{align*}
  \mathcal{L}&(\hat{Y}, \hat{A}, \hat{T}) = \frac{1}{N}\sum_i^N[
    \ol(\hat{Y}(t_i, a_i, x_i), y_i)+ \\ 
    & \alpha\ce(\hat{A}(t_i, x_i), a_i)+\beta\ce(\hat{T}(x_i), t_i)] 
\end{align*}

where $\alpha, \beta\in\mathbb{R}^+$ are hyperparameters balancing each loss component. 
$\ce$ is cross entropy loss, and $\ol$ is the outcome ($y_i$) prediction loss (\emph{e.g.}, mean squared error).
During inference, \methodname\ outputs $\hat{T}(x)$, $\hat{A}(t, x)$ and $\hat{Y}(t, a, x)$, then plugs them into the CFD formula to estimate CATEA (see Section~\ref{sec:background:CFD}).
We provide additional implementation details about \methodname\ in Appendix Section~\ref{sec:appendix:implementation:lobster}.

Through joint optimization with shared neural network representations,  \methodname\ leverages shared components between each nuisance parameter. We hypothesize that this sharing can lead to improved CFD-based CATEA estimation. 
In Section~\ref{sec:experiments}, we empirically validate this hypothesis using semi-synthetic and real-world datasets.

\section{Experiments \& Results}\label{sec:experiments}
In this section, we empirically validate our theoretical insights that CFD reduces estimation error compared to SBD when $\Delta_A$ and $\Delta_Y$ are small in finite sample settings. Additionally, we benchmark \methodname\ against T-learner-based CFD estimators. 

\subsection{Datasets}\label{sec:experiments:datasets}

Building on our theoretical analysis, we design two synthetic datasets. \textbf{Synthetic dataset A} evaluates CATEA estimation under different $\Delta_A$: 

\begin{align*}
    &x_i \sim \mathcal{N}(0,\,\mathbf{I}_{30 \times 30}), \quad
    t_i|x_i = \text{Bern}\left(\sigma\left(\frac{w x_i}{30}\right)\right) \\
    &y_i | a_i, x_i = \text{Bern}\left(\sigma\left((1-a_i) \frac{w_{a0} x_i}{30} + a_i \frac{w_{a1} x_i}{30}\right)\right) 
    \\
    &p_i^{\text{nc}} | t_i, x_i = \frac{\sigma((1-t_i)w_{t0} x_i + t_iw_{t1} x_i)}{\sum_{i=1}^n \sigma((1-t_i)w_{t0} x_i + t_iw_{t1} x_i)} \\
    &\{w, w_{t0}, w_{t1}, w_{a0}, w_{a1}\} \in \mathbb{R}^{30} \sim \unif(-10, 10)
\end{align*}

We sample 1000 data points following the above. To simulate different $\Delta_A$, we generate $a_i|t_i, x_i$ with various non-adherence rate $\gamma$. Lower $\Delta_A$ corresponds to $\gamma \rightarrow 100\%$ under one-sided non-adherence and $\gamma \rightarrow 50\%$ under two-sided non-adherence. $a_i$ is generated via weighted sampling: $1000\times\gamma$ non-adherers are selected with $p_i^{\text{nc}}$ as sampling probability. $a_i$ is set to $1-t_i$ if sampled as non-adherer and $t_i$ if otherwise. 

\begin{figure*}[t]
     \centering
     \subfigure[Synthetic Dataset A]{\includegraphics[width=0.48\linewidth,alt={syntetich_a}]{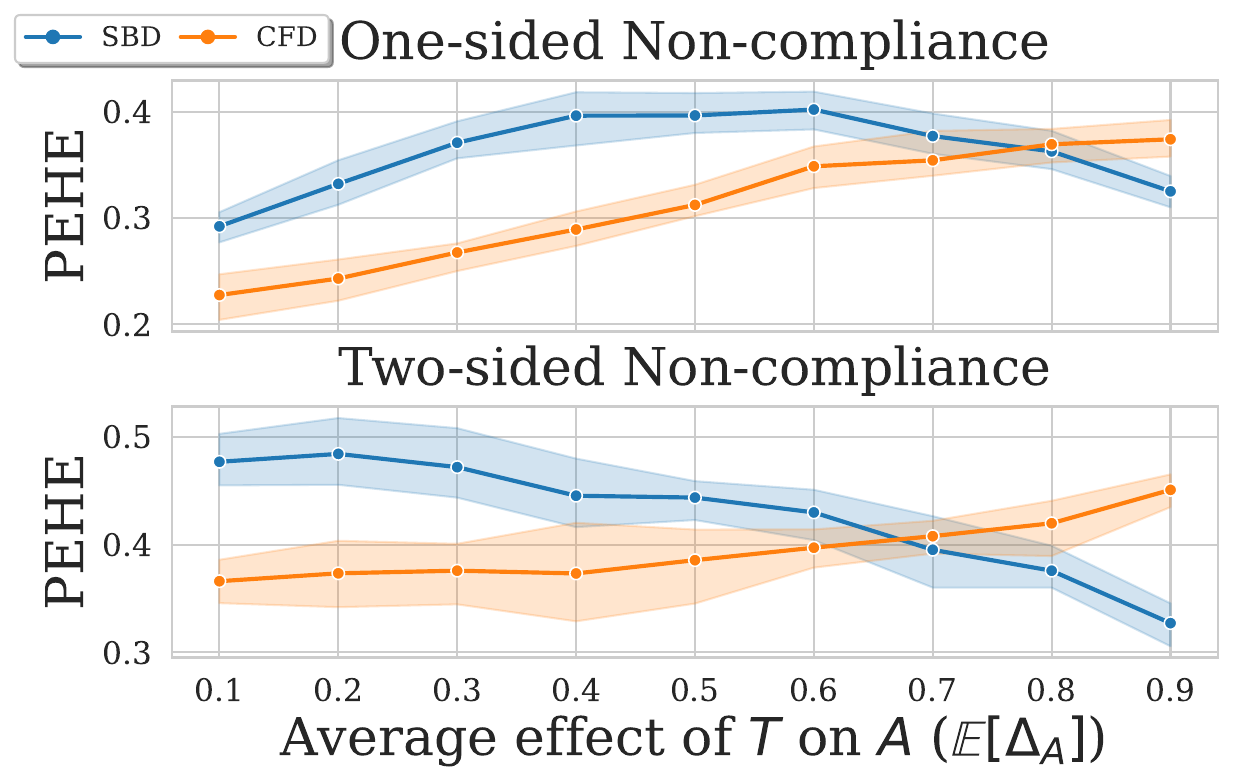}\label{fig:sim:synthetic-a}}
    \subfigure[Synthetic Dataset B]{\includegraphics[width=0.48\linewidth,alt={synthetic_b}]{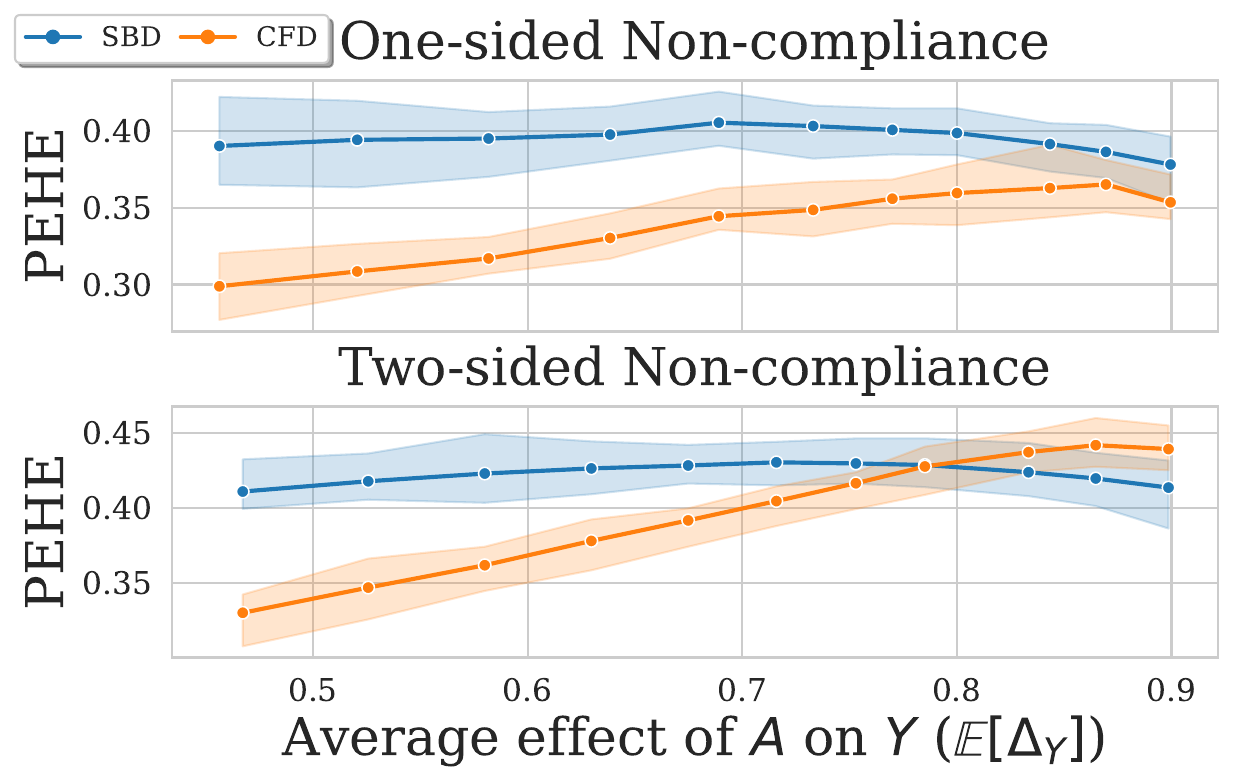}\label{fig:sim:synthetic-b}}
    \caption{CATEA estimation PEHE achieved by SBD and CFD on synthetic datasets. CFD outperforms SBD in most settings. The overall trend of CFD's performance improvement over SBD also supports the theoretical result.}
    \label{fig:sim}
\end{figure*}

\textbf{Synthetic dataset B} evaluates CATEA estimation under different $\Delta_Y$. $x_i$ and $t_i$ generation follows synthetic dataset A. $a_i$ and $y_i$ are generated with:
\begin{align*}
    &a_i|t_i, x_i = \text{Bern}\left(\sigma\left((1-t_i) \frac{w_{t0} x_i}{30} + t_i \frac{w_{t1} x_i}{30}\right)\right) \\
    &y_i|a_i, x_i = \text{Bern}\left((1-a_i)0.1 + a_i\sigma\left(\frac{w_{a1} x_i}{30} + \eta \right)\right)
\end{align*}

$\Delta_Y$ is controlled via $\eta \in \mathbb{R}$, the average effect of treatment intake on the log-odds of $y_i$, with higher $\eta$ giving higher average $\Delta_Y$. $w_{t0}$, $w_{t1}$, and $w_{a1}$ are sampled identically to the synthetic dataset A, except that under one-sided non-adherence, $w_{t0}$ is zero.

Next, we consider the \textbf{IHDP} (Infant Health and Development Program) dataset ($n = 747$)~\citep{hill2011bayesian}, a semi-synthetic dataset created from an experiment that studied the effect of high-quality childcare and home visits on future cognitive test scores. It contains 25 features ($X$) and a binary treatment assignment ($T$). We simulate treatment intake ($A$) identically as the synthetic datasets and adopt the same outcome simulation function proposed by~\cite{hill2011bayesian}, except that we replace $T$ with $A$ as the conditioning variable.

To evaluate our approach with real potential outcomes, we also examine \textbf{AMR-UTI} ($n=15,806$), a dataset containing antimicrobial resistance for patients with urinary tract infections (UTIs)~\citep{oberst2020amr}. 
AMR-UTI covers four types of antimicrobial treatments ($T$) and uniquely provides ground-truth antimicrobial resistance labels for each treatment obtained via microbiology testing, such that the labels are representative of real-world counterfactual outcomes. We adapt AMR-UTI to one-sided and two-sided non-adherence. Under one-sided non-adherence, the treatment assignment indicates whether Nitrofurantoin was prescribed. 
Only patients prescribed Nitrofurantoin can be non-adherent. Under two-sided non-adherence, treatment assignment represents a prescription for one of two antimicrobial treatments (Nitrofurantoin and Ciprofloxacin). 
In this toy setup, patients prescribed either treatment may not adhere to their treatment by taking the other, unassigned treatment.
In both settings, we simulate non-adherence behavior identically to the synthetic datasets. 
We define whether a patient recovers from the UTI as the outcome of interest ($Y$).
Only patients infected with a pathogen susceptible (\emph{i.e.}, non-resistant) to the antimicrobial treatment actually taken will recover. 
The goal is to estimate the individual-level difference in recovery rates between the two treatment assignments.

For IHDP and AMR-UTI, we do not vary $\Delta_Y$ because the potential outcome generation processes are pre-specified. 
 $\Delta_A$ is controlled identically to the synthetic datasets.
For all datasets, we evaluate the quality of CATEA estimation using the precision in the estimation of heterogeneous effect (PEHE) metric~\citep{hill2011bayesian}.

\subsection{Implementation details}\label{sec:experiments:implementation}
We use neural networks to implement CFD and SBD. 
For each dataset, we perform an 80:20 train-test split.
We perform early stopping on a validation split sampled from 20\% of the training split. 
To quantify uncertainty, we repeat data generation and model training 20 times and report the median and inter-quartile range (IQR) of the results. 
Full implementation details are in Appendix~\ref{sec:appendix:implementation}.

\begin{figure*}[t]
     \centering
     \subfigure[IHDP]{\includegraphics[width=0.48\linewidth,alt={ihdp}]{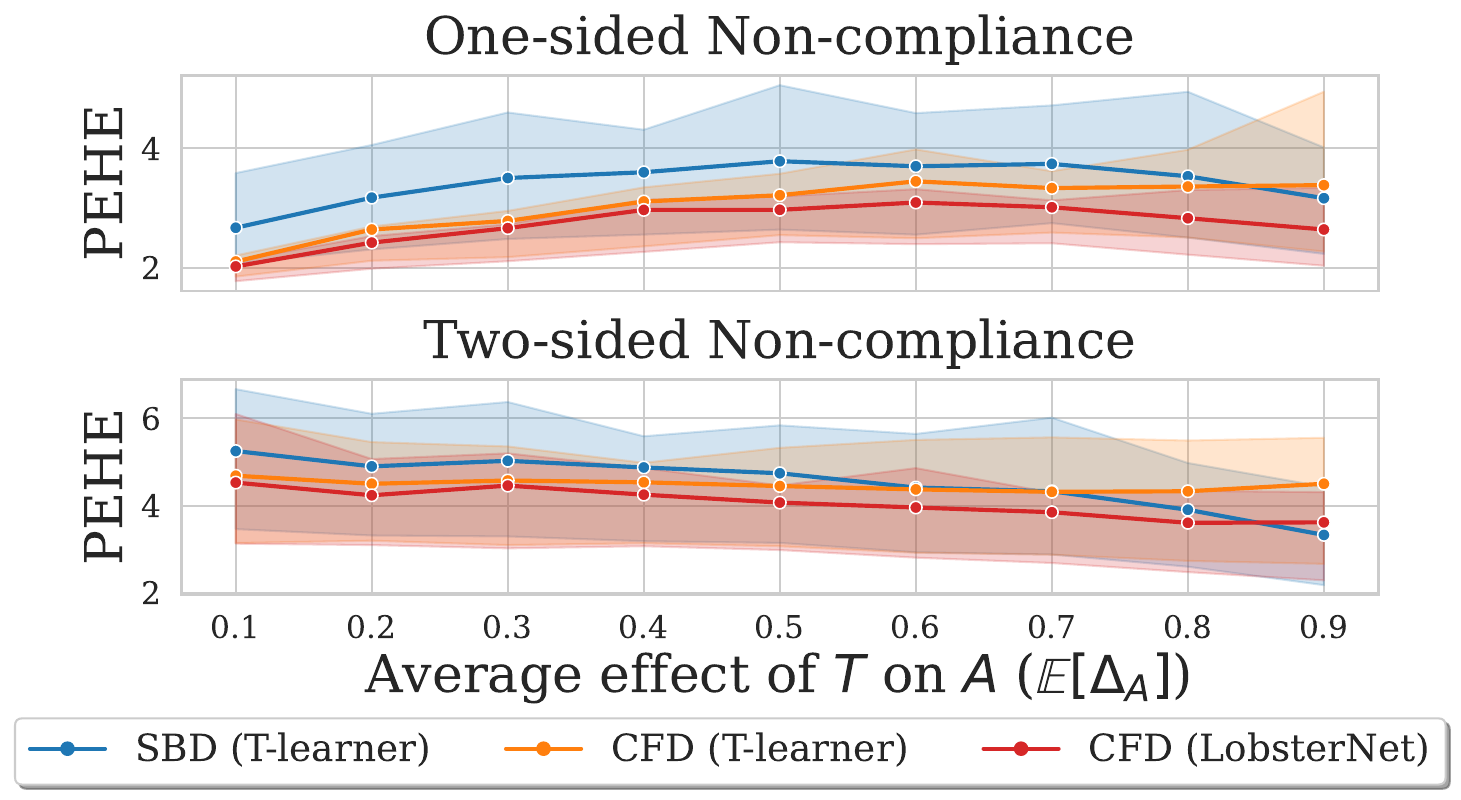}\label{fig:real:ihdp_a}}
    \subfigure[AMR-UTI]{\includegraphics[width=0.48\linewidth,alt={amruti}]{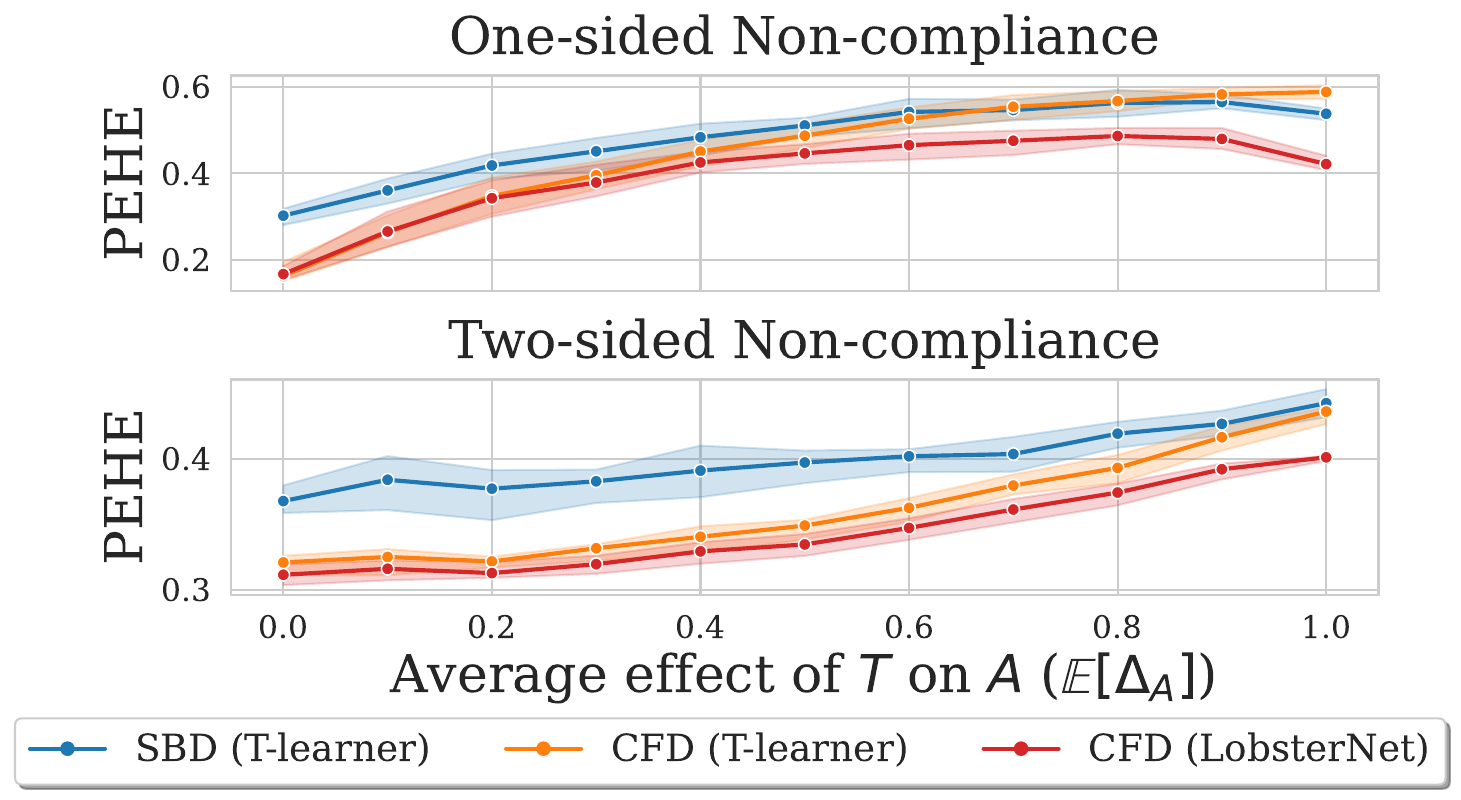}\label{fig:real:amruti}}
    \caption{CATEA estimation PEHE achieved by SBD, CFD, and \methodname\ on IHDP and AMR-UTI datasets. 
    CFD-based methods overall outperform SBD, and \methodname\ yields additional improvements over baseline CFD.}
    \label{fig:real}
\end{figure*}

\subsection{Results on Synthetic Datasets}\label{sec:experiments:cfd-sbd-comp}
\paragraph{Overall, CFD achieves lower PEHE than SBD on both synthetic datasets A and B}. Figure~\ref{fig:sim} directly compares CFD and SBD on the synthetic datasets, with both approaches implemented via T-learners~\citep{kunzel2019metalearners}.

With synthetic dataset A (Figure~\ref{fig:sim:synthetic-a}), CFD achieves lower average PEHE than SBD under both one-sided (CFD: 0.309, 95\%CI: 0.300-0.318 vs. SBD: 0.362, 95\%CI: 0.355-0.368) and two-sided non-adherence (CFD:  0.394, 95\%CI: 0.336-0.345 vs. SBD: 0.428, 95\%CI: 0.422-0.435) across all levels of average $\Delta_A$ ($\mathbb{E}[\Delta_A]$).
Similarly, with synthetic dataset B (Figure~\ref{fig:sim:synthetic-b}), CFD shows an advantage over SBD in average PEHE under both one-sided (CFD: 0.341, 95\%CI: 0.336-0.345  vs. SBD: 0.395, 95\%CI: 0.390-0.400) and two-sided non-adherence (CFD:  0.398, 95\%CI: 0.393-0.405 vs. SBD: 0.423, 95\%CI: 0.419-0.426) across different average $\Delta_Y$ ($\mathbb{E}[\Delta_Y]$). 

\paragraph{CFD's advantage increases as $\Delta_A$ decreases.} 
As shown in Figure~\ref{fig:sim:synthetic-a}, CFD's performance improvement over SBD increases as the $\mathbb{E}[\Delta_A]$ decreases. 
This trend is consistent across both one-sided and two-sided non-adherence, validating our theoretical finding that CFD has lower variance than SBD when $\Delta_A$ is small.

\paragraph{CFD's advantage inversely correlates with $\Delta_Y$.}
Given our theoretical result that CFD has lower variance than SBD when $\Delta_Y$ is smaller, we expect the lift of CFD over SBD to increase as $\mathbb{E}[\Delta_Y]$ decreases. 
This is precisely what we observe in Figure~\ref{fig:sim:synthetic-b}. 
Under both one-sided and two-sided non-adherence, CFD's PEHE improvement over SBD increases as $\mathbb{E}[\Delta_Y]$ decreases, consistent with the theory. 

\subsection{Results on IHDP and AMR-UTI}\label{sec:experiments:lobsternet-comp}
\paragraph{CFD outperforms SBD, and \methodname\ yields additional gains across IHDP and AMR-UTI datasets.}
In Figure~\ref{fig:real}, we compare SBD and CFD with \methodname\ in terms of CATEA estimation PEHE. 
On average, SBD (in blue) produces lower-quality CATEA estimates than either of the CFD-based approaches. 
Trends in relative performance between SBD and CFD in the IHDP and AMR-UTI datasets align with results in the two synthetic datasets: 
CFD provides greater advantages over SBD when $\mathbb{E}[\Delta_A]$ is small, validating our theoretical finding.

\paragraph{\methodname\ additionally improves CATEA estimation compared to CFD.} 
In Figure~\ref{fig:real}, \methodname\ yields a small improvement on CATEA estimation PEHE compared to CFD with a T-learner across all $\mathbb{E}[\Delta_A]$ settings.
This result is consistent in both the IHDP and AMR-UTI datasets, illustrating the empirical benefit of shared representations when modeling CFD nuisance parameters. 

\paragraph{\methodname\ obtains greater improvements over CFD as $\mathbb{E}[\Delta_A]$ increases and is empirically advantageous even when $\mathbb{E}[\Delta_A]$ is large.}
As shown in Figure~\ref{fig:real}, across both IHDP and AMR-UTI datasets, \methodname's relative advantages over CFD consistently correlate with $\mathbb{E}[\Delta_A]$. 
This is because as $\mathbb{E}[\Delta_A]$ increases, individuals' treatment intake conditioned on treatment assignment becomes more deterministic (\textit{i.e.}, adherence improves), leaving fewer observations for non-adherent outcomes.
In such settings, \methodname's shared modeling can efficiently estimate under-observed outcomes.
More importantly, the additional boost from employing \methodname\ allows CFD to outperform or match SBD in PEHE across all settings, suggesting that CFD with \methodname\ is a desirable choice of CATEA estimation even when $\Delta_A$ is large.

\section{Conclusion}\label{sec:conclusion}
We studied CATEA estimation in a setting without perfect treatment adherence. Specifically, we proposed CFD as a framework for incorporating individual treatment intake information. 
We showed that CFD-based approaches have lower asymptotic variance than conventional SBD-based approaches when the effect of treatment assignment is small. However, CFD requires estimating multiple nuisance parameters, which limits its estimation quality in settings where the distribution of treatment intake behavior is skewed. To overcome this limitation, we proposed \methodname, a neural network architecture that empirically surpasses existing methods due to its unified approach to modeling nuisance parameters.

\textbf{Limitations.} 
The CFD-based approach depends on treatment adherence information, which may be unavailable or unreliable~\citep{evans1983problem}. Additionally, our work assumes binary treatments, but real-world treatments could be categorical~\citep{lopez2017estimation}, continuous~\citep{bahadori2022end}, and graph-valued~\citep{kaddour2021causal}. Modeling non-adherence in these challenging settings is a valuable future direction. 

Nonetheless, researchers aiming to estimate CATEAs in settings with non-adherence should consider utilizing CFD-based approaches. 
This approach is theoretically proven to be advantageous when the true CATEA is small, which is often the case in real-world treatment assignments~\citep{ioannidis2005most, gu2016review, vivalt2020much}. 
Empirically, by leveraging our proposed \methodname\ architecture, CFD can reduce estimation error regardless of the true CATEA size.
In real-world healthcare scenarios, choosing CFD with \methodname over the existing SBD can result in better CATEA estimate, potentially improving downstream treatment recommendations. 

\bibliography{references}

\newpage
\onecolumn
\begin{appendix}

\section{Identifiability Assumptions}\label{sec:appendix:assumptions}
\begin{assumption}[Consistency]\label{asup:consistency}
    For every individual $i$, the potential treatment intake and potential outcome under the observed treatment assignment should equal the observed treatment intake and observed outcome, respectively: $A_i(t_i) = a_i, Y_i(A_i(t_i)) = y_i \ \forall i$.
\end{assumption}
\begin{assumption}[Positivity]\label{asup:positivity}
    Every individual has a non-zero probability of receiving either treatment assignment: $0<P(t_i=1 | x_i)<1\ \forall i$.
\end{assumption}
\begin{assumption}[Full Mediation]\label{asup:full-mediation}
    The effect of treatment assignment on the outcome is fully mediated by treatment intake: $\mathbb{E}[Y | do(T=t), x_i] = \sum_{a'\in \{0,1\}}\mathbb{E}[Y | do(A=a'), x_i]P(a' | do(T=t), x_i)$. 
\end{assumption}
\begin{assumption}[Assignment-Outcome Ignorability]\label{asup:assign-outcome-ignore}
    For every individual $i$, the potential outcomes given treatment assignment, $Y_i(A_i(0))$ and $Y_i(A_i(1))$, are conditionally independent of the observed treatment assignment given their pre-treatment covariates: $Y_i(A_i(0)), Y_i(A_i(1)) \perp\!\!\!\!\perp t_i | x_i \ \forall i$.
\end{assumption}
\begin{assumption}[Assignment-Intake Ignorability]\label{asup:assign-intake-ignore}
    For every individual $i$, the potential treatment intakes given assignment, $A_i(0)$ and $A_i(1)$, are conditionally independent of the observed treatment assignment given their pre-treatment covariates: $A_i(0), A_i(1) \perp\!\!\!\!\perp t_i | x_i \ \forall i$.
\end{assumption}
\begin{assumption}[Intake-Outcome Ignorability]\label{asup:intake-outcome-ignore}
    For every individual $i$, the potential outcomes given treatment intake, $Y_i(0)$ and $Y_i(1)$, are conditionally independent of the observed treatment intake given pre-treatment covariates: $Y_i(0), Y_i(1) \perp\!\!\!\!\perp a_i | x_i\ \forall i$.
\end{assumption}

\section{Identifiability 
Proofs}\label{sec:appendix:id-proofs}
\subsection{Proof of SBD's identifiability}\label{sec:appendix:id-proofs:sbd}
We show that $\mathbb{E}[Y | do(T=t), X]$ can be expanded as following:
\begin{align*}
  \mathbb{E}&[Y | do(T=t), X=x_i] \\
    &\stackrel{(1)}{=} \sum_{a'\in\{0,1\}}\mathbb{E}[Y | 
        A=a', do(T=t), X=x_i]P(A=a' | do(T=t), X=x_i) \\
    &\stackrel{(2)}{=} \sum_{a'\in\{0,1\}}\mathbb{E}[Y | 
        A=a', do(T=t), X=x_i]P(A=a' | T=t, X=x_i) \\
    &\stackrel{(3)}{=} \sum_{a'\in\{0,1\}}\mathbb{E}[Y | 
        A=a', T=t, X=x_i]P(A=a' | T=t, X=x_i) \\
    &\stackrel{(4)}{=} \mathbb{E}[Y | T=t, X=x_i]
\end{align*}

Equality (1) factorizes the conditional expectation. Equality (2) applies assignment-intake ignorability (Assumption~\ref{asup:assign-intake-ignore}). Equality (3) applies the assignment-outcome ignorability (Assumption~\ref{asup:assign-outcome-ignore}) and the fact that any unobserved backdoors between treatment intake and outcome are blocked by conditioning on $A$. Equality (4) marginalizes the expectations to complete the proof.
\subsection{Proof of identifiability of CFD under two-sided non-adherence}\label{sec:appendix:id-proofs:cfd-two-sided}
We show that $\mathbb{E}[Y | do(T=t), X=x_i]$ can be expanded as following:
\newpage
\begin{equation}
\begin{aligned}
  \mathbb{E}&[Y | do(T=t), X=x_i] \\
    &\stackrel{(1)}{=} \sum_{a'\in\{0,1\}}\mathbb{E}[Y | 
        A=a', do(T=t), X=x_i]P(A=a' | do(T=t), X=x_i) \\
    &\stackrel{(2)}{=} \sum_{a'\in\{0,1\}}\mathbb{E}[Y | 
        do(A=a'), X=x_i]
      P(A=a' | do(T=t), X=x_i) \\
    &\stackrel{(3)}{=} \sum_{a'\in\{0,1\}} \left(
      \mathbb{E}[Y | do(A=a'), X] P(A=a' | T=t, X=x_i)
    \right) \\
    &\stackrel{(4)}{=} \sum_{a'\in\{0,1\}} \bigg(
      \sum_{t'\in\{0,1\}}\left(
        \mathbb{E}[Y | T=t', do(A=a'), X=x_i]
        P(T=t' | do(A=a'), X=x_i)
      \right) \\
    & \quad\quad\quad\quad\ \quad P(A=a' | T=t, X=x_i)
    \bigg) \\
    &\stackrel{(5)}{=} \sum_{a'\in\{0,1\}} \left(
      \sum_{t'\in\{0,1\}}\left(
        \mathbb{E}[Y | T=t', do(A=a'), X=x_i]
        P(T=t' | X=x_i)
      \right) P(A=a' | T=t, X=x_i)
    \right) \\
    &\stackrel{(6)}{=} \sum_{a'\in\{0,1\}} \left(
      \sum_{t'\in\{0,1\}}\left(
        \mathbb{E}[Y | T=t', A=a', X=x_i]
        P(T=t' | X=x_i)
      \right) P(A=a' | T=t, X=x_i)
    \right) \\
    &\stackrel{(7)}{=} \sum_{t',a'\in\{0,1\}}
        \mathbb{E}[Y | T=t', A=a', X=x_i]P(T=t' | X=x_i)P(A=a' | T=t, X=x_i)
\end{aligned}
\label{eq:proof-two-sided}
\end{equation}
Equality $(1)$ factorizes the conditional expectation. 
Equality $(2)$ applies full mediation (Assumption~\ref{asup:full-mediation}).
Equality $(3)$ is allowed because of assignment-intake ignorability (Assumption~\ref{asup:assign-intake-ignore}).
Equality $(4)$ performs another factorization of conditional expectation. 
Equality $(5)$ applies the definition of $do(\cdot)$ intervention on causal DAG assumed by our problem setup (\textit{i.e.} $A$ does not influence $T$), intervening on $A$ does not affect the probability of $T$.
Equality $(6)$ applies intake-outcome ignorability (Assumption~\ref{asup:intake-outcome-ignore}), and any unobserved backdoor between $T$ and $Y$ is blocked by conditioning on $T$. 
Finally, equality $(7)$ combines the two summations to complete the proof. 

\subsection{Proof of CFD's identifiability under one-sided non-adherence}\label{sec:appendix:id-proofs:cfd-one-sided}

For identifying $\mathbb{E}[Y|do(T=0), X=x_i]$, we start with equality (3) in Equation~\ref{eq:proof-two-sided}:
\begin{align*}
    \mathbb{E}&[Y | do(T=0), X=x_i] = \sum_{a'\in\{0,1\}} 
      \mathbb{E}[Y | do(A=a'), X=x_i] P(A=a' | T=0, X=x_i)
\end{align*}

Applying assumptions for one-sided non-adherence in Section~\ref{sec:background:non-adherence} (\emph{i.e.} $P(A | T=0, X)=0$), we simplify:
\begin{align*}
    \mathbb{E}&[Y | do(T=0), X=x_i] = \mathbb{E}[Y | do(A=0), X=x_i]
\end{align*}

Then, applying the same techniques as equality (4), (5), and (6) in Equation~\ref{eq:proof-two-sided} and assumptions for one-sided non-adherence in Section~\ref{sec:background:non-adherence} (\emph{i.e.} $1-P(A | T=0, X=x_i)=1$), we can get:
\begin{align*}
    \mathbb{E}&[Y | do(T=0), X=x_i] = \sum_{t'\in\{0,1\}}
        \mathbb{E}[Y | A=0, T=t', X=x_i]
        P(T=t' | X=x_i)
\end{align*}

For identifying $\mathbb{E}[Y|do(T=1), X=x_i]$, we start with equality (4) in Equation~\ref{eq:proof-two-sided}.
\begin{align*}
    \mathbb{E}&[Y | do(T=1), X=x_i] \\
    & = \sum_{a'\in\{0,1\}} \bigg(
      \sum_{t'\in\{0,1\}}\left(\mathbb{E}[Y | T=t', do(A=a'), X=x_i]P(T=t' | do(A=a'), X=x_iP\right) \\ 
    &\qquad\qquad\qquad P(A=a' | T=1, X=x_i)\bigg) \\
    & = \sum_{t'\in\{0,1\}}\left(\mathbb{E}[Y | T=t', do(A=0), X]P(T=t' | do(A=0), X=x_i)\right) P(A=0 | T=1, X=x_i) \\
    & \quad+ \sum_{t'\in\{0,1\}}\left(\mathbb{E}[Y | T=t', do(A=1), X=x_i]P(T=t' | do(A=1), X=x_i)\right) P(A=1 | T=1, X=x_i) 
\end{align*}
Applying definitions for one-sided non-adherence in Section~\ref{sec:background:non-adherence} (\emph{i.e.} $P(T=0|do(A=1), X)=0$), we get:
\begin{align*}
    \mathbb{E}&[Y | do(T=1), X=x_i] \\
    &=P(A=0 | T=1, X=x_i)\sum_{t'\in\{0,1\}}\mathbb{E}[Y | T=t', do(A=0), X=x_i]p[T=t' | do(A=0), X=x_i] \\ & \quad + \mathbb{E}[Y | T=1, do(A=1), X=x_i] P(A=1 | T=1, X=x_i)
\end{align*}
Finally, following the same technique as equality (5) and (6) in Equation~\ref{eq:proof-two-sided}, we get the final form:
\begin{align*}
    \mathbb{E}[Y | do(T=1), X=x_i] 
    &= P(A=0 | T=1, X=x_i)\sum_{t'\in\{0,1\}}\mathbb{E}[Y | A=0, T=t', X]P(T=t' | X=x_i)  \\
    & \quad + \mathbb{E}[Y | A=1, T=1, X=x_i] P(A=1 | T=1, X=x_i)
\end{align*}

\section{CFD-based CATEA Estimation Derivation}\label{sec:appendix:cfd-derivation}
Given CFD's identifiability proof under one-sided non-adherence (Appendix~\ref{sec:appendix:id-proofs:cfd-one-sided}), we have its potential outcome estimation defined as follows:
\begin{align*}
    \mathbb{E}[Y | do(T=0), X=x_i] = 
        &\mathbb{E}[Y | A=0, T=0, X=x_i]
        p(T=0 | X=x_i) + \\ 
        &\mathbb{E}[Y | A=0, T=1, X=x_i]
        P(T=1 | X=x_i) \\
    \mathbb{E}[Y | do(T=1), X=x_i] 
    = &P(A=0 | T=1, X=x_i)(\mathbb{E}[Y | A=0, T=0, X=x_i]P(T=0 | X=x_i) + \\ 
        &\mathbb{E}[Y | A=0, T=1, X=x_i]P(T=1 | X=x_i))  + \\ 
        &\mathbb{E}[Y | A=1, T=1, X=x_i] P(A=1 | T=1, X=x_i)
\end{align*}

Substitute in the following simplified notations:
\begin{equation}
\begin{aligned}
  \hat{\pi}(x_i) = \hat{\mathbb{E}}[T | x_i]\quad
  \hat{A}(t_i, x_i) = \hat{\mathbb{E}}[A |T=t_i, x_i]\quad
  \hat{Y}(a_i, t_i, x_i) = \hat{\mathbb{E}}[Y |A=a_i, T=t_i, x_i]
\end{aligned}
\label{eq:cfd-nuisance}
\end{equation}

We get CFD's CATEA estimation under one-sided non-adherence as:
\begin{align*}
    \hat{\Phi_{\text{OS}}^{\text{CFD}}} = &(1-\hat{A}(T=1, x_i))\underbrace{\left(\hat{Y}(A=0, T=0, x_i)(1-\hat{\pi}(x_i)) + \hat{Y}(A=0, T=1, x_i)\hat{\pi}(x_i)\right)}_{\text{same}}
    + \\
    &\hat{Y}(A=1, T=1, x_i)\hat{A}(T=1, x_i) - \underbrace{\left(\hat{Y}(A=0, T=0, x_i)
        (1-\hat{\pi}(x_i)) + \hat{Y}(A=0, T=1, x_i)
        \hat{\pi}(x_i)\right)}_{\text{same}} \\
    =& -\underbrace{\hat{A}(T=1, x_i)}_{\text{same}}\left(\hat{Y}(A=0, T=0, x_i)(1-\hat{\pi}(x_i)) + \hat{Y}(A=0, T=1, x_i)\hat{\pi}(x_i)\right) + \hat{Y}(A=1, T=1, x_i)\underbrace{\hat{A}(T=1, x_i)}_{\text{same}} \\
    =& \left(\underbrace{\hat{Y}(A=1, T=1, x_i)}_{\text{Split into two parts}} - \hat{Y}(A=0, T=0, x_i)(1-\hat{\pi}(x_i)) - \hat{Y}(A=0, T=1, x_i)\hat{\pi}(x_i)\right)\hat{A}(T=1, x_i) \\
    =& \left((\hat{Y}(A=1, T=1, x_i) - \hat{Y}(A=0, T=0, x_i))(1-\hat{\pi}(x_i)) \right.+ \\
    &\ \  \left.(\hat{Y}(A=1, T=1, x_i) - \hat{Y}(A=0, T=1, x_i))\hat{\pi}(x_i)\right)\hat{A}(T=1, x_i)
\end{align*}

Given CFD's identifiability proof under two-sided non-adherence (Appendix~\ref{sec:appendix:id-proofs:cfd-two-sided}), we have its potential outcome estimation defined as follows:
\begin{align*}
    \mathbb{E}[Y | do(T=t), X=x_i] = 
        &\mathbb{E}[Y | A=0, T=0, X=x_i]P(T=0 | X=x_i)P(A=0 | T=t, X=x_i) + \\
        &\mathbb{E}[Y | A=0, T=1, X=x_i]P(T=1 | X=x_i)P(A=0 | T=t, X=x_i) + \\
        &\mathbb{E}[Y | A=1, T=0,  X=x_i]P(T=0 | X=x_i)P(A=1 | T=t, X=x_i) + \\
        &\mathbb{E}[Y | A=1, T=1, X=x_i]P(T=1 | X=x_i)P(A=1 | T=t, X=x_i)
\end{align*}

Substituting in notations from Equation~\ref{eq:cfd-nuisance}, we get CFD's CATEA estimation under two-sided non-adherence as:
\begin{align*}
    \hat{\Phi_{\text{TS}}^{\text{CFD}}} = &\hat{Y}(A=0, T=0, x_i)\left(1-\hat{\pi}(x_i)\right)\left(1-\hat{A}(T=1, x_i)\right) + \hat{Y}(A=0, T=1, x_i)\hat{\pi}(x_i)\left(1-\hat{A}(T=1, x_i)\right) + \\
    &\hat{Y}(A=1, T=0, x_i)\left(1-\hat{\pi}(x_i)\right)\hat{A}(T=1, x_i) + \hat{Y}(A=1, T=1, x_i)\hat{\pi}(x_i)\hat{A}(T=1, x_i) - \\
    &\hat{Y}(A=0, T=0, x_i)\left(1-\hat{\pi}(x_i)\right)\left(1-\hat{A}(T=0, x_i)\right) - \hat{Y}(A=0, T=1, x_i)\hat{\pi}(x_i)\left(1-\hat{A}(T=0, x_i)\right) - \\
        &\hat{Y}(A=1, T=0, x_i)(1-\hat{\pi}(x_i))\hat{A}(T=0, x_i) - \hat{Y}(A=1, T=0, x_i)\hat{\pi}(x_i)\hat{A}(T=0, x_i) \\
    = & - \hat{Y}(A=0, T=0, x_i)(1-\hat{\pi}(x_i))\left(\hat{A}(T=1, x_i) - \hat{A}(T=0, x_i)\right) \\
    & -\hat{Y}(A=0, T=1, x_i)\hat{\pi}(x_i)\left(\hat{A}(T=1, x_i) - \hat{A}(T=0, x_i)\right) \\
    & + \hat{Y}(A=1, T=0, x_i)(1-\hat{\pi}(x_i))\left(\hat{A}(T=1, x_i) - \hat{A}(T=0, x_i)\right) \\
    & + \hat{Y}(A=1, T=1, x_i)\hat{\pi}(x_i)\left(\hat{A}(T=1, x_i) - \hat{A}(T=0, x_i)\right) \\
    = & \left(\hat{Y}(A=1, T=0, x_i)(1-\hat{\pi}(x_i)) + \hat{Y}(A=1, T=1, x_i)\hat{\pi}(x_i) \right. \\ 
    &\left. -\hat{Y}(A=0, T=0, x_i)(1-\hat{\pi}(x_i)) - \hat{Y}(A=0, T=1, x_i)\hat{\pi}(x_i)
    \right)\left(\hat{A}(T=1, x_i) - \hat{A}(T=0, x_i)\right) \\
    = & \bigg(\left(\hat{Y}(A=1, T=0, x_i) - \hat{Y}(A=0, T=0, x_i)\right)\left(1-\hat{\pi}(x_i)\right) + \\ 
    & \ \  \left(\hat{Y}(A=1, T=1, x_i) - \hat{Y}(A=0, T=1, x_i)\right)\hat{\pi}(x_i)
    \bigg)\bigg(\hat{A}(T=1, x_i) - \hat{A}(T=0, x_i)\bigg)
\end{align*}

\section{Asymptotic Variance Derivation}\label{sec:appendix:variance}
We derive and compare the asymptotic variances of two estimators: 1) the standard backdoor adjustment (SBD) estimator that only considers treatment ($T$) and 2) the conditional front-door adjustment (CFD) estimator that considers both treatment ($T$) and action ($A$). The variance bounds for SBD under one-sided non-adherence with binary outcomes ($Y$) are first derived in \cite{rahier2021individual}; 
We follow their framework to develop variance bounds for SBD and CFD under both one-sided and two-sided non-adherence without assuming binary outcomes.

Our variance results are derived under:

\textbf{Single-stratum setting.} We focus on CATEA estimation for a single value $x_0 \in X$, for which we assume to observe $n$ \textit{i.i.d.} samples $\{(x_0, t_i, a_i, y_i)\}_{1\le i\le n}$.

\textbf{Full ignorability setting.} Because SBD and CFD require slightly different ignorability assumptions, for a fair comparison, we assume that all ignorability assumptions required by SBD and CFD are met. In order words, we assume Assumption~\ref{asup:assign-outcome-ignore}, \ref{asup:assign-intake-ignore}, and~\ref{asup:intake-outcome-ignore} hold.

\textbf{Notations.} We first define compact notation for the propensity score 
\begin{equation}
\begin{aligned}
\pi = \mathbb{E}[t | x_0]
\end{aligned}
\label{eq:compact-propensity}
\end{equation}

Then, we define the probability of four different treatment assignments and intake combinations (or non-adherence behaviors):
\begin{equation}
\begin{aligned}
\begin{split}
        \omega_{a=0,t=0} &= P(a=0,t=0 | x_0) \\
        \omega_{a=0,t=1} &= P(a=0,t=1 | x_0)
      \end{split}
    \quad\quad
      \begin{split}
        \omega_{a=1,t=0} &= P(a=1,t=0 | x_0) \\
        \omega_{a=1,t=1} &= P(a=1,t=1 | x_0)
      \end{split} \\
\end{aligned}
\label{eq:compact-comb-prob}
\end{equation}

Next, we define compact notations for treatment intakes generated under different treatment assignments:
\begin{align*}
    \begin{split}
        A_{t=0} \sim P\left(a | t=0, x_0\right)
    \end{split}
    \quad\quad
    \begin{split}
        A_{t=1} \sim P\left(a | t=1, x_0\right)
  \end{split} 
\end{align*}

We also define compact notations for outcomes generated under different treatment assignments or intakes.
\begin{align*}
    \begin{split}
        Y_{t=0} &\sim P\left(y | t=0, x_0\right) \\
        Y_{a=0} &\sim P\left(y | a=0, x_0\right)
    \end{split}
    \quad\quad
    \begin{split}
        Y_{t=1} &\sim P\left(y | t=1, x_0\right) \\
        Y_{a=1} &\sim P\left(y | 1=1, x_0\right)
  \end{split} 
\end{align*}

Finally, we use $\mathbb{E}(\cdot)$ and $\mathbb{V}(\cdot)$ to denote expectation and variance of the corresponding variable.
For instance:
\begin{align*}
      \begin{split}
        \mathbb{E}\left(Y_{t=0}\right) &= \mathbb{E}[y | t=0, x_0] \\
      \end{split}
    \quad\quad
      \begin{split}
        \mathbb{V}\left(Y_{t=1}\right) &= \mathrm{Var}(y | t=0, x_0) \\
      \end{split} 
\end{align*}

For notational simplicity, we also define $\Delta_A$ and $\Delta_Y$ as the effect of assignment on intake and the effect of intake on the outcome, which are decompositions of the CATEA (\textit{i.e.} $\Phi(x_o) = \Delta_A\Delta_Y$):
\begin{align*}
\begin{split}
    \Delta_A &= \mathbb{E}\left(A_{t=1}\right)-\mathbb{E}\left(A_{t=0}\right)\\
\end{split}
\begin{split}
    \Delta_Y &= \mathbb{E}\left(Y_{a=1}\right)-\mathbb{E}\left(Y_{a=0}\right)
\end{split}
\end{align*}

The derivation is organized in four steps:
\begin{enumerate}
  \item Define maximum-likelihood CATEA estimators (Appendix~\ref{sec:appendix:variance:estimators-define})
  \item Derive variance of SBD and CFD estimators (Appendix~\ref{sec:appendix:variance:estimators-derive})
  \item Derive asymptotic variance upper and lower bounds (Appendix~\ref{sec:appendix:variance:bound})
  \item Compare SBD and CFD's variance bounds directly. (Appendix~\ref{sec:appendix:variance:compare})
\end{enumerate}

\subsection{Define maximum-likelihood and CATEA estimators.}\label{sec:appendix:variance:estimators-define}
Given $n$ \textit{i.i.d.} samples $\{(t_i, a_i, y_i)\}_{1\le i \le n}$, we define maximum-likelihood (MLE) estimators of notations described above.

The MLE estimator of the propensity score (Equation~\ref{eq:compact-propensity}) is: 
\begin{align*}
\hat{\pi} &= \frac{\sum\limits_{i=1}^n t_i}{n}
\end{align*}

The MLE estimators of the four different treatment assignments and intake combinations (Equation~\ref{eq:compact-comb-prob}) are:
\begin{align*}
    \begin{split}
        \hat{\omega}_{a=0,t=0} &= \frac{\sum\limits_{i=1}^n(1-a_i)(1-t_i)}{n} \\
        \hat{\omega}_{a=0,t=1} &= \frac{\sum\limits_{i=1}^n(1-a_i)t_i}{n}
    \end{split}
    \quad\quad
    \begin{split}
        \hat{\omega}_{a=1,t=0} &= \frac{\sum\limits_{i=1}^na_i(1-t_i)}{n} \\
        \hat{\omega}_{a=1,t=1} &= \frac{\sum\limits_{i=1}^na_it_i}{n}
    \end{split} \\
\end{align*}

The MLE estimators of the potential treatment intakes are defined as follows:
\begin{align*}
      \begin{split}
        \hat{\mathbb{E}}\left(A_{t=0}\right) &= \frac{\sum\limits^n_{i=1}(1-t_i)a_i}{\sum\limits_{i=1}^{n}(1-t_i)} \\
      \end{split}
    \quad\quad
      \begin{split}
        \hat{\mathbb{E}}\left(A_{t=1}\right) &= \frac{\sum\limits^n_{i=1}t_ia_i}{\sum\limits_{i=1}^{n}t_i} \\
      \end{split} \\
\end{align*}

The MLE estimators of the potential outcomes are defined as follows:
\begin{align*}
      \begin{split}
        \hat{\mathbb{E}}\left(Y_{t=0}\right) &= \frac{\sum\limits^n_{i=1}(1-t_i)y_i}{\sum\limits_{i=1}^{n}(1-t_i)} \\
        \hat{\mathbb{E}}\left(Y_{a=0, t=0}\right) &= \frac{\sum\limits^n_{i=1}(1-a_i)(1-t_i)y_i}{\sum\limits_{i=1}^{n}(1-a_i)(1-t_i)} \\
        \hat{\mathbb{E}}\left(Y_{a=0, t=1}\right) &= \frac{\sum\limits^n_{i=1}(1-a_i)t_iy_i}{\sum\limits_{i=1}^{n}(1-a_i)t_i}
      \end{split}
    \quad\quad
      \begin{split}
        \hat{\mathbb{E}}\left(Y_{t=1}\right) &= \frac{\sum\limits^n_{i=1}t_iy_i}{\sum\limits_{i=1}^{n}t_i} \\
        \hat{\mathbb{E}}\left(Y_{a=1, t=0}\right) &= \frac{\sum\limits^n_{i=1}a_i(1-t_i)y_i}{\sum\limits_{i=1}^{n}a_i(1-t_i)} \\
        \hat{\mathbb{E}}\left(Y_{a=1, t=1}\right) &= \frac{\sum\limits^n_{i=1}a_it_iy_i}{\sum\limits_{i=1}^{n}a_it_i} 
      \end{split} \\
\end{align*}

Using the MLE estimator notations, we define the SBD estimator under both one-sided and two-sided non-adherence settings:
\begin{equation}
\begin{aligned}
    \hat{\Phi}^{\text{SBD}} = \hat{\mathbb{E}}\left(Y_{t=1}\right) - \hat{\mathbb{E}}\left(Y_{t=0}\right)
\end{aligned}
\label{eq:sbd-def}
\end{equation}
Under one-sided non-adherence, the CFD estimator is defined as:
\begin{equation}
\begin{aligned}
    \hat{\Phi}^{\text{CFD}}_{\text{os}} = (\hat{\mathbb{E}}\left(Y_{a=1,t=1}\right) - \hat{\mathbb{E}}\left(Y_{a=0,t=0}\right)(1-\hat{\pi}) - \hat{\mathbb{E}}\left(Y_{a=0,t=1}\right)\hat{\pi})\hat{\mathbb{E}}\left(A_{t=1}\right)
\end{aligned}
\label{eq:one-sided-cfd-def}
\end{equation}
Under two-sided non-adherence, the CFD estimator is defined as:
\begin{equation}
\begin{aligned}
    \hat{\Phi}^{\text{CFD}}_{\text{ts}} = ((\hat{\mathbb{E}}\left(Y_{a=1, t=0}\right) - \hat{\mathbb{E}}\left(Y_{a=0, t=0}\right))(1-\hat{\pi}) + (\hat{\mathbb{E}}\left(Y_{a=1, t=1}\right) - \hat{\mathbb{E}}\left(Y_{a=0, t=1}\right))\hat{\pi})(\hat{\mathbb{E}}\left(A_{t=1}\right)-\hat{\mathbb{E}}\left(A_{t=0}\right))
\end{aligned}
\label{eq:two-sided-cfd-def}
\end{equation}

\subsection{Derive variance of SBD and CFD estimators}\label{sec:appendix:variance:estimators-derive}
\subsubsection{SBD estimator's variance derivation derivation}\label{sec:baseline-variance}
We start by applying the law of total variance to get:
\begin{align*}
    \text{Var}\left(\hat{\Phi}^{\text{SBD}}\right) = \underbrace{\mathbb{E}\left[\text{Var}\left(\hat{\Phi}^{\text{SBD}} \mid \{t_k\}_k\right)\right]}_{\circled{A}} + \underbrace{\text{Var}\left(\mathbb{E}\left[\hat{\Phi}^{\text{SBD}} \mid \{t_k\}_k\right]\right)}_{\circled{B}}
\end{align*}
where $\{t_k\}_k \coloneqq \{t_1, \cdots, t_n\}$. \\

\textbf{Computation of $\circled{A}$} \\
Given definition from Equation~\ref{eq:sbd-def}, term $\circled{A}$ decomposes as:
\begin{equation}
\begin{aligned}
    \mathbb{E}&\left[\text{Var}\left(\hat{\Phi}^{\text{SBD}} \mid \{t_k\}_k\right)\right] \\
    &= 
    \mathbb{E}\left[
        \underbrace{\text{Var}\left(\hat{\mathbb{E}}\left(Y_{t=1}\right) \mid \{t_k\}_k\right)}_{\circled{A1}} + 
        \underbrace{\text{Var}\left(\hat{\mathbb{E}}\left(Y_{t=0}\right) \mid \{t_k\}_k\right)}_{\circled{A2}} - 2\underbrace{\text{Cov}\left(\hat{\mathbb{E}}\left(Y_{t=1}\right), \hat{\mathbb{E}}\left(Y_{t=0}\right) \mid \{t_k\}_k\right)}_{\circled{A3}}
    \right]
\label{eq:eq-A}
\end{aligned}
\end{equation}

For $\circled{A1}$, we write:
\begin{equation}
\begin{aligned}
    \text{Var}&\left(\hat{\mathbb{E}}\left(Y_{t=1}\right) \mid \{t_k\}_k\right) = \text{Var}\left(\frac{\sum\limits^n_{i=1}t_iy_i}{\underbrace{\sum\limits_{i=1}^{n}t_i}_{\text{Conditional Independence}}} \mid \{t_k\}_k\right) = \left(\frac{1}{\sum\limits_{i=1}^{n}t_i}\right)^2\text{Var}\left(\sum\limits^n_{i=1} \underbrace{t_iy_i}_{\textit{i.i.d}} \mid \{t_k\}_k\right) \\
    &= \left(\frac{1}{\sum\limits_{i=1}^{n}t_i}\right)^2\sum\limits^n_{i=1}\text{Var}\left(t_iy_i \mid \underbrace{\{t_k\}_k}_{t_i\ \text{since samples are}\ \textit{i.i.d}}\right) = \left(\frac{1}{\sum\limits_{i=1}^{n}t_i}\right)^2\sum\limits^n_{i=1}\underbrace{\text{Var}\left(t_iy_i | t_i\right)}_{t_i\text{Var}\left(y_i | t_i\right)\ (t_i\ \text{is binary})} \\
    &= \left(\frac{1}{\sum\limits_{i=1}^{n}t_i}\right)^2\sum\limits^n_{i=1}t_i\underbrace{\text{Var}\left(y_i | t_i\right)}_{\text{Var}(y | t=1)\ (\text{since } \textit{i.i.d. } \text{and multiplied by } t_i)} = \left(\frac{1}{\sum\limits_{i=1}^{n}t_i}\right)^2 \text{Var}\left(y \mid t=1\right) \sum\limits^n_{i=1}t_i = \frac{\mathbb{V}\left(Y_{t=1}\right)}{n\hat{\pi}}
\end{aligned}
\label{eq:cfd-a1}
\end{equation}

For $\circled{A2}$, we analogously get:
\begin{align*}
    \text{Var}(\hat{\mathbb{E}}\left(Y_{t=0}\right) \mid \{t_k\}_k) = \frac{\mathbb{V}\left(Y_{t=0}\right)}{n\left(1-\hat{\pi}\right)}
\end{align*}

For $\circled{A3}$, we start by applying bi-linearity of covariance:
\begin{equation}
\begin{aligned}
    \text{Cov}&\left(\hat{\mathbb{E}}\left(Y_{t=1}\right), \hat{\mathbb{E}}\left(Y_{t=0}\right) \mid \{t_k\}_k\right) = \text{Cov}\left(
        \frac{\sum\limits^n_{i=1}t_iy_i}{\sum\limits_{i=1}^{n}t_i}, 
        \frac{\sum\limits^n_{i=j}(1-t_j)y_j}{\sum\limits_{j=1}^{n}(1-t_j)}
        \mid \{t_k\}_k\right) = \frac{\text{Cov}\left(\sum\limits^n_{i=1}t_iy_i, \sum\limits^n_{j=1}(1-t_j)y_j \mid \{t_k\}_k\right)}{\sum\limits_{i=1}^{n}t_i\sum\limits_{j=1}^{n}(1-t_j)} \\
        &= \frac{\sum\limits^n_{i=1} \sum\limits^n_{j=1}t_i(1-t_j) \underbrace{\text{Cov}\left(y_i, y_j| \{t_k\}_k\right)}_{\neq 0 \text{ only if } i=j \text{ (since } \textit{i.i.d.})}}{\sum\limits_{i=1}^{n}t_i\sum\limits_{j=1}^{n}(1-t_j)} = \frac{\sum\limits^n_{i=1}\underbrace{t_i(1-t_i)}_{=0}\text{Var}\left(y_i \mid \{t_k\}_k\right)}{\sum\limits_{i=1}^{n}t_i \sum\limits_{j=1}^{n}(1-t_j)} = 0
\end{aligned}
\label{eq:baseline-zero-cov}
\end{equation}

Injecting values of $\circled{A1}$, $\circled{A2}$, and $\circled{A3}$ in Equation~\ref{eq:eq-A}, we get expression for $\circled{A}$:
\begin{equation}
\begin{aligned}
    \mathbb{E}[\text{Var}(\hat{\Phi}^{\text{SBD}} | \{t_k\}_k)] &= \frac{\mathbb{V}\left(Y_{t=1}\right)}{\mathbb{E}\left[n\hat{\pi}\right]}  + \frac{ \mathbb{V}\left(Y_{t=0}\right)}{\mathbb{E}\left[n(1-\hat{\pi})\right]} = \frac{\mathbb{V}\left(Y_{t=1}\right)}{n\pi}  + \frac{ \mathbb{V}\left(Y_{t=0}\right)}{n(1-\pi)}
\end{aligned}
\label{eq:eq-A-derived}
\end{equation}

\textbf{Computation of $\circled{B}$} \\
Given definition from Equation~\ref{eq:sbd-def}, term $\circled{B}$ decomposes as:
\begin{equation}
\begin{aligned}
    \text{Var}\left(\mathbb{E}\left[\hat{\Phi}^{\text{SBD}} \mid \{t_k\}_k\right]\right) = 
    \text{Var}\left(
        \underbrace{\mathbb{E}\left[\hat{\mathbb{E}}\left(Y_{t=1}\right) \mid \{t_k\}_k\right]}_{\circled{B1}} - 
        \underbrace{\mathbb{E}\left[\hat{\mathbb{E}}\left(Y_{t=0}\right) \mid \{t_k\}_k\right]}_{\circled{B2}}
    \right)
\end{aligned}
\label{eq:eq-B}
\end{equation}

For $\circled{B1}$, we apply similar techniques from computing $\circled{A}$ to get:
\begin{equation}
\begin{aligned}
\mathbb{E}&\left[\hat{\mathbb{E}}\left(Y_{t=1}\right) \mid \{t_k\}_k\right] = \mathbb{E}\left[
    \frac{\sum\limits^n_{i=1}t_iy_i}{\sum\limits_{i=1}^{n}t_i} \mid \{t_k\}_k
\right] = \frac{\mathbb{E}\left[\sum\limits^n_{i=1}t_iy_i \mid \{t_k\}_k\right]}{\sum\limits_{i=1}^{n}t_i} = \frac{\sum\limits^n_{i=1}\mathbb{E}\left[t_iy_i | \{t_k\}_k\right] }{\sum\limits_{i=1}^{n}t_i}\\
&= \frac{\sum\limits^n_{i=1}t_i\mathbb{E}\left[y_i \mid \{t_k\}_k\right]}{\sum\limits_{i=1}^{n}t_i} = \frac{\sum\limits^n_{i=1}t_i\underbrace{\mathbb{E}\left[y_i \mid t_i\right]}_{\mathbb{E}\left(Y_{t=1}\right)}}{\sum\limits_{i=1}^{n}t_i} = \frac{\sum\limits^n_{i=1}t_i}{\sum\limits_{i=1}^{n}t_i}\mathbb{E}\left(Y_{t=1}\right) = \mathbb{E}\left(Y_{t=1}\right)
\end{aligned}
\label{eq:cfd-b1}
\end{equation}

Analogously, $\circled{B2}$ can be written as:
\begin{align*}
    \mathbb{E}\left[\hat{\mathbb{E}}\left(Y_{t=0}\right) \mid \{t_k\}_k\right] = \mathbb{E}\left(Y_{t=0}\right)
\end{align*}

Injecting $\circled{B1}$ and $\circled{B2}$ into Equation~\ref{eq:eq-B}, we get the expression for $\circled{B}$ as:
\begin{equation}
\begin{aligned}
    \text{Var}\left(\mathbb{E}\left[\hat{\Phi}^{\text{SBD}} | \{t_k\}_k\right]\right) = 
    \text{Var}\left(\underbrace{\mathbb{E}\left(Y_{t=1}\right)-\mathbb{E}\left(Y_{t=0}\right)}_{\text{constant}}\right) = 0
\end{aligned}
\label{eq:eq-B-derived}
\end{equation}

Combining expression for $\circled{A}$ (Equation~\ref{eq:eq-A-derived}) and $\circled{B}$ (Equation~\ref{eq:eq-B-derived}), we get:\
\begin{align*}
    \text{Var}\left(\hat{\Phi}^{\text{SBD}}\right) = \frac{\mathbb{V}\left(Y_{t=1}\right)}{n\pi}  + \frac{\mathbb{V}\left(Y_{t=0}\right)}{n\left(1-\pi\right)}
\end{align*}

Given the full mediation assumption, we have the following decompositions:
\begin{align*}
    P\left(y \mid t=0\right) &= \begin{cases} 
      P\left(y \mid a=0\right) & \text{(one-sided non-adherence)} \\
      P\left(y \mid a=0\right)(1-P\left(a \mid t=0\right)) + P\left(y \mid  a=1\right)P\left(a \mid t=0\right) & \text{(two-sided non-adherence)} \\
   \end{cases} \\
    P\left(y\mid t=1\right) &= P\left(y\mid a=0\right)(1-P\left(a\mid t=1\right)) + P\left(y\mid a=1\right)P\left(a\mid t=1\right) 
\end{align*}

Applying the above decompositions allow us to further expand $\mathbb{V}\left(Y_{t=0}\right)$ and $\mathbb{V}\left(Y_{t=1}\right)$.
Under one-sided non-adherence, $\mathbb{V}\left(Y_{t=0}\right)=\mathbb{V}\left(Y_{a=0}\right)$, and under two-sided non-adherence, $\mathbb{V}\left(Y_{t=0}\right)$ decomposes as following:
\begin{equation}
\begin{aligned}
    \mathbb{V}\left(Y_{t=0}\right) &= \mathrm{Var}\left(Y_{a=0}(1-A_{t=0}) + Y_{a=1}A_{t=0}\right) \\
    &= \mathrm{Var}\left(Y_{a=0} - A_{t=0}Y_{a=0}+Y_{a=1}A_{t=0}\right) \\
    &= \mathrm{Var}\left(Y_{a=0}\right) + \mathrm{Var}\left(A_{t=0}Y_{a=0}\right) + \mathrm{Var}\left(A_{t=0}Y_{a=1}\right) - \\
    &\quad\ 2\mathrm{Cov}\left(Y_{a=0}, A_{t=0}Y_{a=0}\right) - 2\mathrm{Cov}\left(A_{t=0}Y_{a=0}, A_{t=0}Y_{a=1}\right) + 2\mathrm{Cov}\left(Y_{a=0}, A_{t=0}Y_{a=1}\right)
\end{aligned}
\label{eq:sbd-var-decompose}
\end{equation}

Leveraging the full ignorability assumption (\textit{i.e.}, all potential outcomes and intakes are independent), we decompose the variances of produces:
\begin{align*}
    \mathrm{Var}\left(A_{t=0}Y_{a=0}\right) &= \mathbb{E}\left(\left(A_{t=0}Y_{a=0}\right)^2\right) - \mathbb{E}\left(A_{t=0}Y_{a=0}\right)^2 \\
    &= \mathbb{E}\left(A_{t=0}^2\right)\mathbb{E}\left(Y_{a=0}^2\right) - \mathbb{E}\left(A_{t=0}\right)^2\mathbb{E}\left(Y_{a=0}\right)^2 \quad \text{(Applying independence)} \\
    &= \mathbb{E}\left(A_{t=0}^2\right)\left(\mathbb{V}\left(Y_{a=0}\right)+\mathbb{E}\left(Y_{a=0}\right)^2\right) - \mathbb{E}\left(A_{t=0}\right)^2\mathbb{E}\left(Y_{a=0}\right)^2 \quad \text{(Definition of variance)} \\
    &= \mathbb{E}\left(\underbrace{A_{t=0}^2}_{A_{t=0}\text{ (binary)}}\right)\mathbb{V}\left(Y_{a=0}\right) + \mathbb{E}\left(Y_{a=0}^2\right)\left(\underbrace{\mathbb{E}\left(A_{t=0}^2\right) - \mathbb{E}\left(A_{t=0}\right)^2}_{\text{Variance by definition}}\right) \\
    &= \mathbb{E}\left(A_{t=0}\right)\mathbb{V}\left(Y_{a=0}\right) + \mathbb{E}\left(Y_{a=0}\right)^2\mathbb{V}\left(A_{t=0}\right) \\
    \mathbb{V}\left(A_{t=0}Y_{a=1}\right) &= \mathbb{E}\left(A_{t=0}\right)\mathbb{V}\left(Y_{a=1}\right) + \mathbb{E}\left(Y_{a=1}\right)^2\mathbb{V}\left(A_{t=0}\right)\quad \text{(Analogous as above)}
\end{align*}

Similarly, leveraging the full ignorability assumption, we decompose the covariance terms:
\begin{align*}
    \mathrm{Cov}\left(Y_{a=0}, A_{t=0}Y_{a=0}\right) &= \mathbb{E}\left(A_{t=0}\right)\mathbb{V}\left(Y_{a=0}\right)\\
    \mathrm{Cov}\left(A_{t=0}Y_{a=0}, A_{t=0}Y_{a=1}\right) &= \mathbb{E}\left(Y_{a=0}\right)\mathbb{E}\left(Y_{a=1}\right)\mathbb{V}\left(A_{t=0}\right)\\
    \mathrm{Cov}\left(Y_{a=0}, A_{t=0}Y_{a=1}\right) &= 0 \quad \text{(Independence)}
\end{align*}

Plug the above results back into the Equation~\ref{eq:sbd-var-decompose}, we get:
\begin{align*}
    \mathbb{V}\left(Y_{t=0}\right) &= \mathbb{V}\left(Y_{a=0}\right) + \mathbb{E}\left(A_{t=0}\right)\mathbb{V}\left(Y_{a=0}\right) + \mathbb{E}\left(Y_{a=0}\right)^2\mathbb{V}\left(A_{t=0}\right) + \mathbb{E}\left(A_{t=0}\right)\mathbb{V}\left(Y_{a=1}\right) + \mathbb{E}\left(Y_{a=1}\right)^2\mathbb{V}\left(A_{t=0}\right) \\
    &\quad\ -2\mathbb{E}\left(A_{t=0}\right)\mathbb{V}\left(Y_{a=0}\right) -2\mathbb{E}\left(Y_{a=0}\right)\mathbb{E}\left(Y_{a=1}\right)\mathbb{V}\left(A_{t=0}\right) \\
    &= \left(1-2\mathbb{E}\left(A_{t=0}\right) + \mathbb{E}\left(A_{t=0}\right)\right)\mathbb{V}\left(Y_{a=0}\right) + \mathbb{E}\left(A_{t=0}\right)\mathbb{V}\left(Y_{a=1}\right) + \left(\mathbb{E}\left(Y_{a=1}\right)-\mathbb{E}\left(Y_{a=1}\right)\right)^2\mathbb{V}\left(A_{t=0}\right) \\
    &= \left(1-\mathbb{E}\left(A_{t=0}\right)\right)\mathbb{V}\left(Y_{a=0}\right) + \mathbb{E}\left(A_{t=0}\right)\mathbb{V}\left(Y_{a=1}\right) + \left(\mathbb{E}\left(Y_{a=1}\right)-\mathbb{E}\left(Y_{a=0}\right)\right)^2\mathbb{V}\left(A_{t=0}\right) \\
    &= \left(1-\mathbb{E}\left(A_{t=0}\right)\right)\mathbb{V}\left(Y_{a=0}\right) + \mathbb{E}\left(A_{t=0}\right)\mathbb{V}\left(Y_{a=1}\right) + \Delta_Y^2\mathbb{V}\left(A_{t=0}\right)
\end{align*}

Analogously, we have $\mathbb{V}\left(Y_{t=1}\right)$ as:
\begin{align*}
    \mathbb{V}\left(Y_{t=1}\right) &= \left(1-\mathbb{E}\left(A_{t=1}\right)\right)\mathbb{V}\left(Y_{a=0}\right) + \mathbb{E}\left(A_{t=1}\right)\mathbb{V}\left(Y_{a=1}\right) + \Delta_Y^2\mathbb{V}\left(A_{t=1}\right)
\end{align*}

Putting everything together, we have the variance of the SBD estimator:
\begin{align*}
    \text{Var}\left(\hat{\Phi}_{\text{OS}}^{\text{SBD}}\right) &= \frac{\left(1-\mathbb{E}\left(A_{t=1}\right)\right)\mathbb{V}\left(Y_{a=0}\right) + \mathbb{E}\left(A_{t=1}\right)\mathbb{V}\left(Y_{a=1}\right) + \Delta_A^2\mathbb{V}\left(A_{t=1}\right)}{n\pi}  + \frac{\mathbb{V}\left(Y_{a=0}\right)}{n(1-\pi)} \\
    \text{Var}\left(\hat{\Phi}_{\text{TS}}^{\text{SBD}}\right) &= \frac{\left(1-\mathbb{E}\left(A_{t=1}\right)\right)\mathbb{V}\left(Y_{a=0}\right) + \mathbb{E}\left(A_{t=1}\right)\mathbb{V}\left(Y_{a=1}\right) + \Delta_A^2\mathbb{V}\left(A_{t=1}\right)}{n\pi}  + \\
    &\quad\ \frac{\left(1-\mathbb{E}\left(A_{t=0}\right)\right)\mathbb{V}\left(Y_{a=0}\right) + \mathbb{E}\left(A_{t=0}\right)\mathbb{V}\left(Y_{a=1}\right) + \Delta_A^2\mathbb{V}\left(A_{t=0}\right)}{n(1-\pi)}
\end{align*}

\subsubsection{CFD estimator’s variance derivation under one-sided non-adherence}
We start by applying the law of total variance to get:
\begin{equation}
\begin{aligned}
    \text{Var}(\hat{\Phi}^{\text{CFD}}_{\text{os}}) = 
    \underbrace{\mathbb{E}\left[\text{Var}\left(\hat{\Phi}^{\text{CFD}}_{\text{os}} | \{a_k,  t_k\}_k\right)\right]}_{\circled{C}} + \underbrace{\text{Var}\left(\mathbb{E}\left[\hat{\Phi}^{\text{CFD}}_{\text{os}} | \{a_k, t_k\}_k\right]\right)}_{\circled{D}}
\end{aligned}
\label{eq:cfd-one-sided-start}
\end{equation}

\textbf{Computation of $\circled{C}$} \\
Given definition from Equation~\ref{eq:one-sided-cfd-def}, we can write term $\circled{C}$ as:

\begin{equation}
\begin{aligned}
    \mathbb{E}&\left[\text{Var}\left(\hat{\Phi}^{\text{CFD}}_{\text{os}} \mid \{a_k,  t_k\}_k\right)\right] = \mathbb{E}\left[\text{Var}((\hat{Y}_{a=1,t=1} - \hat{Y}_{a=0,t=0}(1-\hat{\pi}) - \hat{Y}_{a=0,t=1}\hat{\pi})\hat{\mathbb{E}}\left(A_{t=1}\right) \mid \{a_k,  t_k\}_k)\right] \\
    &= \mathbb{E}[\hat{\mathbb{E}}\left(A_{t=1}\right)^2\text{Var}(\hat{Y}_{a=1,t=1} - \hat{Y}_{a=0,t=0}(1-\hat{\pi}) - \hat{Y}_{a=0,t=1}\hat{\pi} | \{a_k,  t_k\}_k)] \\
    &= \mathbb{E}\left[\hat{\mathbb{E}}\left(A_{t=1}\right)^2\left(
        \underbrace{\text{Var}(\hat{Y}_{a=1,t=1} | \{a_k,  t_k\}_k)}_{\circled{C1}} + 
        \underbrace{(1-\hat{\pi})^2\text{Var}(\hat{Y}_{a=0,t=0} | \{a_k,  t_k\}_k)}_{\circled{C2}} + 
        \underbrace{\hat{\pi}^2\text{Var}(\hat{Y}_{a=0,t=1} | \{a_k,  t_k\}_k)}_{\circled{C3}} \right. \right.\\
        & \quad \left. \left.- 
        \underbrace{(1-\hat{\pi})\text{Cov}(\hat{Y}_{a=1,t=1}, \hat{Y}_{a=0,t=0} | \{a_k,  t_k\}_k)}_{\circled{C4}} - \underbrace{\hat{\pi}\text{Cov}(\hat{Y}_{a=1,t=1}, \hat{Y}_{a=0,t=1} | \{a_k,  t_k\}_k)}_{\circled{C5}} \right. \right.\\
        & \quad \left. \left.+ \underbrace{\hat{\pi}(1-\hat{\pi})\text{Cov}(\hat{Y}_{a=0,t=0}, \hat{Y}_{a=0,t=1} | \{a_k,  t_k\}_k)}_{\circled{C6}}\right)\right]
\end{aligned}
\label{eq:cfd-one-sided-c}
\end{equation}

For $\circled{C1}$, we write:
\begin{equation}
\begin{aligned}
    \text{Var}&(\hat{Y}_{a=1,t=1} | \{a_k,  t_k\}_k) \\
        &= \text{Var}\left(\frac{\sum\limits_{i=1}^n a_i t_i y_i}{\sum\limits_{i=1}^n a_i t_i}  | \{a_k,  t_k\}_k\right) 
        = \left(\frac{1}{\sum\limits_{i=1}^n a_i t_i}\right)^2\text{Var}\left(\sum\limits_{i=1}^n a_i t_i y_i  | \underbrace{\{a_k,  t_k\}_k}_{a_i, t_i, \text{ since samples are \emph{i.i.d.}}}\right) \\
        &= \left(\frac{1}{\sum\limits_{i=1}^n a_i t_i}\right)^2 \sum\limits_{i=1}^n \text{Var}\left(a_i t_i y_i  | a_i,  t_i\right) \quad \text{(samples are \emph{i.i.d.})} \\
        &= \left(\frac{1}{\sum\limits_{i=1}^n a_i t_i}\right)^2 \sum\limits_{i=1}^n \underbrace{(a_i t_i)^2}_{a_it_i \text{(binary)}} \underbrace{\text{Var}\left(y_i  | a_i,  t_i\right)}_{\text{Var}(y|a=1,t=1) \text{, since \emph{i.i.d} and multiplying with $a_i, t_i$}} \\
        &= \left(\frac{1}{\sum\limits_{i=1}^n a_i t_i}\right)^2 \text{Var}(y|a=1,t=1)\sum\limits_{i=1}^n a_i t_i 
        = \frac{1}{\sum\limits_{i=1}^n a_i t_i}\text{Var}(y|a=1,t=1) = \frac{\mathbb{V}\left(Y_{a=1}\right)}{n\hat{\omega}_{a=1,t=1}}
\end{aligned}
\label{eq:cfd-one-sided-c1}
\end{equation}

Analogously, we get $\circled{C2}$ and $\circled{C3}$ as following:
\begin{equation}
\begin{aligned}
    (1-\hat{\pi})^2\text{Var}&(\hat{Y}_{a=0,t=0} | \{a_k,  t_k\}_k) = (1-\hat{\pi})^2\frac{\mathbb{V}\left(Y_{a=0}\right)}{n\hat{\omega}_{a=0,t=0}} \\
    \hat{\pi}^2\text{Var}&(\hat{Y}_{a=0,t=1} | \{a_k,  t_k\}_k) = \hat{\pi}\frac{\mathbb{V}\left(Y_{a=0}\right)}{n\hat{\omega}_{a=0,t=1}}
\end{aligned}
\label{eq:cfd-one-sided-c2-c3}
\end{equation}

For $\circled{C4}$, we start by applying the bi-linearity of covariance:
\begin{equation}
\begin{aligned}
    (1-\hat{\pi})&\text{Cov}(\hat{Y}_{a=1,t=1}, \hat{Y}_{a=0,t=0} | \{a_k,  t_k\}_k) = (1-\hat{\pi})\text{Cov}\left(\frac{\sum\limits_{i=1}^n a_i t_i y_i}{\sum\limits_{i=1}^n a_i t_i}, \frac{\sum\limits_{j=1}^n(1-a_j)(1-t_j)y_j}{\sum\limits_{j=1}^n(1-a_j)(1-t_j)} | \{a_k,  t_k\}_k\right) \\
    &= \frac{(1-\hat{\pi})}{\sum\limits_{i=1}^n a_i t_i \sum\limits_{j=1}^n(1-a_j)(1-t_j)} \text{Cov}\left(\sum\limits_{i=1}^n a_i t_i y_i, \sum\limits_{j=1}^n(1-a_j)(1-t_j)y_j | \{a_k,  t_k\}_k\right) \\
    &= \frac{(1-\hat{\pi})}{\sum\limits_{i=1}^n a_i t_i \sum\limits_{j=1}^n(1-a_j)(1-t_j)} \sum\limits_{i=1}^n\sum\limits_{j=1}^n a_i t_i (1-a_j)(1-t_j)  \underbrace{\text{Cov}\left(y_i, y_j | \{a_k,  t_k\}_k\right)}_{\neq 0 \text{ only if } i=j \text{ because samples are \emph{i.i.d.}}} \\
    &= \frac{(1-\hat{\pi})}{\sum\limits_{i=1}^n a_i t_i \sum\limits_{j=1}^n(1-a_j)(1-t_j)} \sum\limits_{i=1}^n \underbrace{a_i t_i (1-a_i)(1-t_i)}_{0}  \text{Cov}\left(y_i, y_j | \{a_k,  t_k\}_k\right) = 0
\end{aligned}
\label{eq:cfd-one-sided-c4}
\end{equation}

Analogously, $\circled{C5}$ and $\circled{C6}$ are also $0$.

Injecting $\circled{C1}, \circled{C2}, \circled{C3}$ back into $\circled{C}$, we get:
\begin{equation}
\begin{aligned}
     \mathbb{E}&[\text{Var}(\hat{\Phi}^{\text{CFD}}_{\text{os}} | \{a_k,  t_k\}_k)] = \mathbb{E}\left[\hat{\mathbb{E}}\left(A_{t=1}\right)^2\left( 
      \frac{\mathbb{V}\left(Y_{a=1}\right)}{n\hat{\omega}_{a=1,t=1}} +
      (1-\hat{\pi})^2\frac{\mathbb{V}\left(Y_{a=0}\right)}{n\hat{\omega}_{a=0,t=0}} + 
      \hat{\pi}^2\frac{\mathbb{V}\left(Y_{a=0}\right)}{n\hat{\omega}_{a=0,t=1}}
     \right)\right] \\
     &= \frac{1}{n}\left(\mathbb{E}\left[\frac{\hat{\mathbb{E}}\left(A_{t=1}\right)^2}{\hat{\omega}_{a=1,t=1}}\right]\mathbb{V}\left(Y_{a=1}\right) + \left(\mathbb{E}\left[\frac{\hat{\mathbb{E}}\left(A_{t=1}\right)^2(1-\hat{\pi})^2}{\hat{\omega}_{a=0,t=0}}\right] + \mathbb{E}\left[\frac{\hat{\mathbb{E}}\left(A_{t=1}\right)^2\hat{\pi}^2}{\hat{\omega}_{a=0,t=1}}\right]\right)\mathbb{V}\left(Y_{a=0}\right)\right)
\end{aligned}
\label{eq:cfd-one-sided-c-result}
\end{equation}

\textbf{Computation of $\circled{D}$}

Given definition from Equation~\ref{eq:one-sided-cfd-def}, we can write term $\circled{D}$ as:
\begin{equation}
\begin{aligned}
    \text{Var}(\mathbb{E}&[\hat{\Phi}^{\text{CFD}_{\text{os}}} | \{a_k,  t_k\}_k]) = \text{Var}(\mathbb{E}[(\hat{Y}_{a=1,t=1} - \hat{Y}_{a=0,t=0}(1-\hat{\pi}) - \hat{Y}_{a=0,t=1}\hat{\pi})\hat{\mathbb{E}}\left(A_{t=1}\right) \mid \{a_k,  t_k\}_k]) \\
    &= \text{Var}(\hat{\mathbb{E}}\left(A_{t=1}\right)\mathbb{E}[\hat{Y}_{a=1,t=1} - \hat{Y}_{a=0,t=0}(1-\hat{\pi}) - \hat{Y}_{a=0,t=1}\hat{\pi} \mid \{a_k,  t_k\}_k]) \\
    &= \text{Var}\left(\hat{\mathbb{E}}\left(A_{t=1}\right)\left(
        \underbrace{\mathbb{E}[\hat{Y}_{a=1,t=1} \mid \{a_k,t_k\}_k]}_{\circled{D1}} - \underbrace{(1-\hat{\pi})\mathbb{E}[\hat{Y}_{a=0,t=0} \mid \{a_k,t_k\}_k]}_{\circled{D2}} - \underbrace{\hat{\pi}\mathbb{E}[\hat{Y}_{a=0,t=1} \mid \{a_k,t_k\}_k]}_{\circled{D3}}\right)\right)
\end{aligned}
\label{eq:cfd-two-sided-d}
\end{equation}

For $\circled{D1}$, we write:
\begin{equation}
\begin{aligned}
    \mathbb{E}&[\hat{Y}_{a=1,t=1} \mid \{a_k,t_k\}_k] 
    = \mathbb{E}\left[\frac{\sum\limits_{i=1}^{n}a_it_iy_i}{\sum\limits_{i=1}^{n}a_it_i} \mid \{a_k,t_k\}_k\right] = \frac{1}{\sum\limits_{i=1}^{n}a_it_i}\mathbb{E}\left[\sum\limits_{i=1}^{n}a_it_iy_i \mid \{a_k,t_k\}_k\right]
    \\
    &= \frac{1}{\sum\limits_{i=1}^{n}a_it_i}\sum\limits_{i=1}^{n}\mathbb{E}\left[a_it_iy_i | \underbrace{\{a_k,t_k\}_k}_{a_i, t_i \text{, because \emph{i.i.d.} samples}}\right] \\
    &= \frac{1}{\sum\limits_{i=1}^{n}a_it_i}\sum\limits_{i=1}^{n}\mathbb{E}\left[a_it_iy_i | a_i, t_i\right] 
    = \frac{1}{\sum\limits_{i=1}^{n}a_it_i}\sum\limits_{i=1}^{n}a_it_i\underbrace{\mathbb{E}\left[y_i | a_i, t_i\right] }_{\mathbb{E}[y|a=1,t=1] \text{ \emph{i.i.d.} and multiplying by $a_it_i$}} \\
    &= \frac{1}{\sum\limits_{i=1}^{n}a_it_i}\mathbb{E}\left[y | a=1, t=1\right]\sum\limits_{i=1}^{n}a_it_i 
    = \mathbb{E}\left[y | a=1, t=1\right] = \mathbb{E}\left[Y_{a=1,t=1}\right] = \mathbb{E}\left(Y_{a=1}\right)
\end{aligned}
\label{eq:cfd-two-sided-d1}
\end{equation}

Analogously, we get $\circled{D2}$ and $\circled{D3}$ as following:
\begin{equation}
\begin{aligned}
    (1-\hat{\pi})\mathbb{E}[\hat{Y}_{a=0,t=0} | \{a_k,t_k\}_k] &= (1-\hat{\pi})\mathbb{E}\left(Y_{a=0}\right) \\
    \hat{\pi}\mathbb{E}[\hat{Y}_{a=0,t=1} | \{a_k,t_k\}_k] &= \hat{\pi}\mathbb{E}\left(Y_{a=0}\right)
\end{aligned}
\label{eq:cfd-two-sided-d2-d3}
\end{equation}

Injecting $\circled{D1}, \circled{D2}, \circled{D3}$ back into $\circled{D}$ (equation~\ref{eq:cfd-two-sided-d}), we get:
\begin{equation}
\begin{aligned}
\text{Var}(\mathbb{E}&[\hat{\Phi}^{\text{CFD}}_{\text{os}} | \{a_k,  t_k\}_k])
    = \text{Var}\left(\hat{\mathbb{E}}\left(A_{t=1}\right)\left(
        \mathbb{E}\left(Y_{a=1}\right) - (1-\hat{\pi})\mathbb{E}\left(Y_{a=0}\right) - \hat{\pi}\mathbb{E}\left(Y_{a=0}\right)\right)\right) \\
    &= \text{Var}\left(\hat{\mathbb{E}}\left(A_{t=1}\right) (\mathbb{E}\left(Y_{a=1}\right)-\mathbb{E}\left(Y_{a=0}\right)\right) = \Delta_Y^2\underbrace{\text{Var}\left(\hat{\mathbb{E}}\left(A_{t=1}\right)\right)}_{\circled{D7}}
\end{aligned}
\label{eq:cfd-two-sided-d-intermediate}
\end{equation}

To simplify $\circled{D7}$, we again apply the law of total variance to get:
\begin{equation}
\begin{aligned}
\text{Var}\left(\hat{\mathbb{E}}\left(A_{t=1}\right)\right) = 
\underbrace{\mathbb{E}[\text{Var}(\hat{\mathbb{E}}\left(A_{t=1}\right) \mid \{t_j\}_k)]}_{\circled{D7.1}} + \underbrace{\text{Var}(\mathbb{E}[\hat{\mathbb{E}}\left(A_{t=1}\right) \mid \{t_j\}_k])}_{\circled{D7.2}}
\end{aligned}
\label{eq:cfd-two-sided-d7}
\end{equation}

Analogous to the derivation of $\circled{A1}$ (Equation~\ref{eq:cfd-a1}), $\circled{D7.1}$ can be written as:
\begin{equation}
\begin{aligned}
\mathbb{E}&[\text{Var}(\hat{\mathbb{E}}\left(A_{t=1}\right) \mid \{t_j\}_k)] = \frac{\mathbb{V}\left(A_{t=1}\right)}{n\pi}
\end{aligned}
\label{eq:cfd-two-sided-d7.1}
\end{equation}

Analogous to the derivation of $\circled{B1}$ (Equation~\ref{eq:cfd-b1}),
$\circled{D7.2}$ can be written as:
\begin{align*}
\text{Var}\left(\mathbb{E}[\hat{\mathbb{E}}\left(A_{t=1}\right) \mid \{t_j\}_k]\right) = \text{Var}\left(\mathbb{E}\left(A_{t=1}\right)\right) = 0
\end{align*}

Injecting $\circled{D7.1}$ and $\circled{D7.2}$ back to $\circled{D}$, we get:
\begin{equation}
\begin{aligned}
\text{Var}\left(\mathbb{E}[\hat{\Phi}^{\text{CFD}}_{\text{os}} | \{a_k,  t_k\}_k]\right) &= \Delta_Y^2\frac{\mathbb{V}\left(A_{t=1}\right)}{n\pi}
\end{aligned}
\label{eq:cfd-two-sided-d-final}
\end{equation}

Now injecting $\circled{C}$ and $\circled{D}$ back into Equation~\ref{eq:cfd-one-sided-start}, we get variance for $\hat{\Phi}^{\text{CFD}_{\text{os}}}$ as:
\begin{align*}
    \text{Var}(\hat{\Phi}^{\text{CFD}}_{\text{os}}) &= \frac{1}{n}\left(\mathbb{V}\left(Y_{a=0}\right)\left(\mathbb{E}\left[\frac{\hat{\mathbb{E}}\left(A_{t=1}\right)^2(1-\hat{\pi})^2}{\hat{\omega}_{a=0,t=0}}\right] + \mathbb{E}\left[\frac{\hat{\mathbb{E}}\left(A_{t=1}\right)^2\hat{\pi}^2}{\hat{\omega}_{a=0,t=1}}\right]\right) + \right. \\
    &\left. \quad +\mathbb{V}\left(Y_{a=1}\right)\mathbb{E}\left[\frac{\hat{\mathbb{E}}\left(A_{t=1}\right)^2}{\hat{\omega}_{a=1,t=1}}\right] 
      + \mathbb{V}\left(A_{t=1}\right)\frac{\Delta_Y^2}{\pi}\right)
\end{align*}

\subsubsection{CFD estimator’s variance derivation under two-sided non-adherence}

We start by applying the law of total variance to get:
\begin{equation}
\begin{aligned}
    \text{Var}(\hat{\Phi}^{\text{CFD}}_{\text{ts}}) = 
    \underbrace{\mathbb{E}\left[\text{Var}\left(\hat{\Phi}^{\text{CFD}}_{\text{ts}} \mid \{a_k,  t_k\}_k\right)\right]}_{\circled{E}} + \underbrace{\text{Var}\left(\mathbb{E}\left[\hat{\Phi}^{\text{CFD}}_{\text{ts}} \mid \{a_k, t_k\}_k\right]\right)}_{\circled{F}}
\end{aligned}
\label{eq:cfd-two-sided-start}
\end{equation}

\textbf{Computation of $\circled{E}$} \\
Given definition from Equation~\ref{eq:two-sided-cfd-def}, we can write term $\circled{E}$ as:
\begin{equation}
\begin{aligned}
    \mathbb{E}&[\text{Var}(\hat{\Phi}^{\text{CFD}}_{\text{ts}} | \{a_k,  t_k\}_k)] \\
    &= \mathbb{E}[\text{Var}(((\hat{\mathbb{E}}\left(Y_{a=1, t=0}\right) - \hat{\mathbb{E}}\left(Y_{a=0, t=0}\right))(1-\hat{\pi}) + (\hat{\mathbb{E}}\left(Y_{a=1, t=1}\right) - \hat{\mathbb{E}}\left(Y_{a=0, t=1}\right))\hat{\pi})(\hat{\mathbb{E}}\left(A_{t=1}\right)-\hat{\mathbb{E}}\left(A_{t=0}\right)) | \{a_k,  t_k\}_k)] \\
    &= \mathbb{E}[(\hat{\mathbb{E}}\left(A_{t=1}\right)-\hat{\mathbb{E}}\left(A_{t=0}\right))^2\text{Var}((\hat{\mathbb{E}}\left(Y_{a=1, t=0}\right) - \hat{\mathbb{E}}\left(Y_{a=0, t=0}\right))(1-\hat{\pi}) + (\hat{\mathbb{E}}\left(Y_{a=1, t=1}\right) - \hat{\mathbb{E}}\left(Y_{a=0, t=1}\right))\hat{\pi} | \{a_k,  t_k\}_k)] \\
    &= \mathbb{E}[(\hat{\mathbb{E}}\left(A_{t=1}\right)-\hat{\mathbb{E}}\left(A_{t=0}\right))^2(
        2\underbrace{\text{Cov}((\hat{\mathbb{E}}\left(Y_{a=1, t=0}\right) - \hat{\mathbb{E}}\left(Y_{a=0, t=0}\right))(1-\hat{\pi}), (\hat{\mathbb{E}}\left(Y_{a=1, t=1}\right) - \hat{\mathbb{E}}\left(Y_{a=0, t=1}\right))\hat{\pi}| \{a_k,  t_k\}_k)}_{\circled{E1}} + \\
        &\ \quad\underbrace{\text{Var}((\hat{\mathbb{E}}\left(Y_{a=1, t=0}\right) - \hat{\mathbb{E}}\left(Y_{a=0, t=0}\right))(1-\hat{\pi})| \{a_k,  t_k\}_k)}_{\circled{E2}} + 
        \underbrace{\text{Var}((\hat{\mathbb{E}}\left(Y_{a=1, t=1}\right) - \hat{\mathbb{E}}\left(Y_{a=0, t=1}\right))\hat{\pi}| \{a_k,  t_k\}_k)}_{\circled{E3}}
    )] 
\end{aligned}
\label{eq:cfd-two-sided-e}
\end{equation}

For $\circled{E1}$, start by applying the bi-linearity of covariance:
\begin{align*}
    \text{Cov}&((\hat{\mathbb{E}}\left(Y_{a=1, t=0}\right) - \hat{\mathbb{E}}\left(Y_{a=0, t=0}\right))(1-\hat{\pi}), (\hat{\mathbb{E}}\left(Y_{a=1, t=1}\right) - \hat{\mathbb{E}}\left(Y_{a=0, t=1}\right))\hat{\pi}| \{a_k,  t_k\}_k) \\
    & = (1-\hat{\pi})\hat{\pi}\text{Cov}((\hat{\mathbb{E}}\left(Y_{a=1, t=0}\right) - \hat{\mathbb{E}}\left(Y_{a=0, t=0}\right)), (\hat{\mathbb{E}}\left(Y_{a=1, t=1}\right) - \hat{\mathbb{E}}\left(Y_{a=0, t=1}\right))| \{a_k,  t_k\}_k) \\
    & = (1-\hat{\pi})\hat{\pi}
        (\underbrace{\text{Cov}(\hat{\mathbb{E}}\left(Y_{a=1, t=0}\right), \hat{\mathbb{E}}\left(Y_{a=1, t=1}\right)| \{a_k,  t_k\}_k)}_{\circled{E1.1}} + \underbrace{\text{Cov}(\hat{\mathbb{E}}\left(Y_{a=0, t=0}\right), \hat{\mathbb{E}}\left(Y_{a=0, t=1}\right)| \{a_k,  t_k\}_k)}_{\circled{E1.2}} \\
        &\ \quad - \underbrace{\text{Cov}(\hat{\mathbb{E}}\left(Y_{a=1, t=0}\right), \hat{\mathbb{E}}\left(Y_{a=0, t=1}\right)| \{a_k,  t_k\}_k)}_{\circled{E1.3}} - \underbrace{\text{Cov}(\hat{\mathbb{E}}\left(Y_{a=0, t=0}\right), \hat{\mathbb{E}}\left(Y_{a=1, t=1}\right)| \{a_k,  t_k\}_k)}_{\circled{E1.4}})
\end{align*}

For $\circled{E1.1}$, also applying bi-linearity of covariance, we write:
\begin{equation}
\begin{aligned}
    \text{Cov}&(\hat{\mathbb{E}}\left(Y_{a=1, t=0}\right), \hat{\mathbb{E}}\left(Y_{a=1, t=1}\right)| \{a_k,  t_k\}_k) = \text{Cov}\left(\frac{\sum\limits^n_{i=1}a_i(1-t_i)y_i}{\sum\limits_{j=1}^{n}a_i(1-t_i)}, \frac{\sum\limits^n_{j=1}a_jt_jy_j}{\sum\limits_{j=1}^{n}a_jt_j} | \{a_k,  t_k\}_k\right) \\
    &= \frac{1}{\sum\limits_{i=1}^{n}a_i(1-t_i)}\frac{1}{\sum\limits_{j=1}^{n}a_jt_j}\text{Cov}\left(\sum\limits^n_{i=1}a_i(1-t_i)y_i, \sum\limits^n_{j=1}a_jt_jy_j | \{a_k,  t_k\}_k \right) \\
    &= \frac{1}{\sum\limits_{i=1}^{n}a_i(1-t_i)}\frac{1}{\sum\limits_{j=1}^{n}a_jt_j} \sum\limits_{i=1}^n\sum\limits_{j=1}^na_ia_j(1-t_i)t_j\underbrace{\text{Cov}(y_i, y_j | \{a_k,  t_k\}_k)}_{\neq 0 \text{ only if } i=j \text{ (since } \textit{i.i.d,)}} \\
    &= \frac{1}{\sum\limits_{i=1}^{n}a_i(1-t_i)}\frac{1}{\sum\limits_{j=1}^{n}a_jt_j} \sum\limits_{i=1}^na_ia_i\underbrace{(1-t_i)t_i}_{0}\text{Cov}(y_i, y_i | \{a_k,  t_k\}_k) = 0
\end{aligned}
\label{eq:cfd-two-sided-cov}
\end{equation}

Analogously, we have $\circled{E1.2}$, $\circled{E1.3}$, $\circled{E1.4}$ also equal to 0. Therefore, we have $\circled{E1}$:
\begin{align*}
    \text{Cov}((\hat{\mathbb{E}}\left(Y_{a=1, t=0}\right) - \hat{\mathbb{E}}\left(Y_{a=0, t=0}\right))(1-\hat{\pi}), (\hat{\mathbb{E}}\left(Y_{a=1, t=1}\right) - \hat{\mathbb{E}}\left(Y_{a=0, t=1}\right))\hat{\pi}| \{a_k,  t_k\}_k) = 0    
\end{align*}

For $\circled{E2}$, we write:
\begin{equation}
\begin{aligned}
    \text{Var}&((\hat{\mathbb{E}}\left(Y_{a=1, t=0}\right) - \hat{\mathbb{E}}\left(Y_{a=0, t=0}\right))(1-\hat{\pi})| \{a_k,  t_k\}_k) = (1-\hat{\pi})^2\text{Var}((\hat{\mathbb{E}}\left(Y_{a=1, t=0}\right) - \hat{\mathbb{E}}\left(Y_{a=0, t=0}\right))| \{a_k,  t_k\}_k) \\
    &= (1-\hat{\pi})^2(
    \underbrace{\text{Var}(\hat{\mathbb{E}}\left(Y_{a=1, t=0}\right)| \{a_k,  t_k\}_k)}_{\circled{E2.1}} + 
    \underbrace{\text{Var}(\hat{\mathbb{E}}\left(Y_{a=0, t=0}\right)| \{a_k,  t_k\}_k)}_{\circled{E2.2}} \\
    &\quad - 
    2\underbrace{\text{Cov}(\hat{\mathbb{E}}\left(Y_{a=1, t=0}\right), \hat{\mathbb{E}}\left(Y_{a=0, t=0}\right)| \{a_k,  t_k\}_k)}_{\circled{E2.3}})
\end{aligned}
\label{eq:e2}
\end{equation}

For $\circled{E2.1}$, we write:
\begin{align*}
    \text{Var}&(\hat{\mathbb{E}}\left(Y_{a=1, t=0}\right)| \{a_k,  t_k\}_k) = \text{Var}\left(\frac{\sum\limits^n_{i=1}a_i(1-t_i)y_i}{\sum\limits_{i=1}^{n}a_i(1-t_i)} | \{a_k,  t_k\}_k\right) \\
    &= \left(\frac{1}{\sum\limits_{i=1}^{n}a_i(1-t_i)}\right)^2\text{Var}\left(\sum\limits^n_{i=1}\underbrace{a_i(1-t_i)y_i}_{\textit{i.i.d.}} | \underbrace{\{a_k,  t_k\}_k}_{a_i, t_i \text{ since  \textit{i.i.d.}}}\right)= \left(\frac{1}{{\sum\limits_{i=1}^{n}a_i(1-t_i)}}\right)^2\sum\limits^n_{i=1}\text{Var}\left(a_i(1-t_i)y_i | a_i, t_i\right) \\
    &= \left(\frac{1}{{\sum\limits_{i=1}^{n}a_i(1-t_i)}}\right)^2\sum\limits^n_{i=1}\underbrace{(a_i(1-t_i))^2}_{a_i(1-t_i)\text{ since binary}.}\underbrace{\text{Var}\left(y_i | a_i, t_i\right)}_{\text{Var}(y | a=1, t=0) \text{ (because \textit{i.i.d.} and multiply by $a_i(1-t_i)$)} } \\
    &= \left(\frac{1}{{\sum\limits_{i=1}^{n}a_i(1-t_i)}}\right)^2\text{Var}(y|a=1, t=0)\sum\limits^n_{i=1}a_i(1-t_i) = \frac{\mathbb{V}\left(Y_{a=1,t=0}\right)}{\sum\limits^n_{i=1}a_i(1-t_i)} = \frac{\mathbb{V}\left(Y_{a=1}\right)}{n\hat{\omega}_{a=1,t=0}}
\end{align*}

Analogously, we get expression for $\circled{E2.2}$ as:
\begin{align*}
    \text{Var}(\hat{\mathbb{E}}\left(Y_{a=0, t=0}\right)| \{a_k,  t_k\}_k) = \frac{\mathbb{V}\left(Y_{a=0}\right)}{n\hat{\omega}_{a=0,t=0}}
\end{align*}

For $\circled{E2.3}$, analogous to Equation~\ref{eq:cfd-two-sided-cov}, we can get:
\begin{align*}
    \text{Cov}(\hat{\mathbb{E}}\left(Y_{a=1, t=0}\right), \hat{\mathbb{E}}\left(Y_{a=0, t=0}\right)| \{a_k,  t_k\}_k) = 0
\end{align*}

Injecting $\circled{E2.1}, \circled{E2.2}, \circled{E2.3}$ back to Equation~\ref{eq:e2}, we get expression for $\circled{E2}$: 
\begin{align*}
    \text{Var}((\hat{\mathbb{E}}\left(Y_{a=1, t=0}\right) - \hat{\mathbb{E}}\left(Y_{a=0, t=0}\right))(1-\hat{\pi}) mid \{a_k,  t_k\}_k) = (1-\hat{\pi})^2\left(\frac{\mathbb{V}\left(Y_{a=1}\right)}{n\hat{\omega}_{a=2,t=0}} + \frac{\mathbb{V}\left(Y_{a=0}\right)}{n\hat{\omega}_{a=0,t=0}}\right)
\end{align*}

Analogously, we can get expression for $\circled{E3}$:
\begin{align*}
    \text{Var}&((\hat{\mathbb{E}}\left(Y_{a=1, t=1}\right) - \hat{\mathbb{E}}\left(Y_{a=0, t=1}\right))\hat{\pi} \mid \{a_k,  t_k\}_k) = \hat{\pi}^2\left(\frac{\mathbb{V}\left(Y_{a=1}\right)}{n\hat{\omega}_{a=1,t=1}} + \frac{\mathbb{V}\left(Y_{a=0}\right)}{n\hat{\omega}_{a=0,t=1}}\right)
\end{align*}

Injecting $\circled{E1}, \circled{E2}, \circled{E3}$ into Equation~\ref{eq:cfd-two-sided-e}, we get expression for $\circled{E}$:
\begin{align*}
    \mathbb{E}&[\text{Var}(\hat{\Phi}^{\text{CFD}}_{\text{ts}} | \{a_k,  t_k\}_k)] = \mathbb{E}\left[\left(\hat{\mathbb{E}}\left(A_{t=1}\right)-\hat{\mathbb{E}}\left(A_{t=0}\right)\right)^2\left( (1-\hat{\pi})^2\left(\frac{\mathbb{V}\left(Y_{a=1}\right)}{n\hat{\omega}_{a=1,t=0}} + \frac{\mathbb{V}\left(Y_{a=0}\right)}{n\hat{\omega}_{a=0,t=0}}\right) \right.\right. \\ 
    &\left.\left.+ \hat{\pi}^2\left(\frac{\mathbb{V}\left(Y_{a=1}\right)}{n\hat{\omega}_{a=1,t=1}} + \frac{\mathbb{V}\left(Y_{a=0}\right)}{n\hat{\omega}_{a=0,t=1}}\right)\right) \right] \\
    &= \mathbb{E}\left[\frac{(\hat{\mathbb{E}}\left(A_{t=1}\right)-\hat{\mathbb{E}}\left(A_{t=0}\right))^2(1-\hat{\pi})^2}{n\hat{\omega}_{a=1,t=0}}\right]\mathbb{V}\left(Y_{a=1}\right)
    + \mathbb{E}\left[\frac{(\hat{\mathbb{E}}\left(A_{t=1}\right)-\hat{\mathbb{E}}\left(A_{t=0}\right))^2(1-\hat{\pi})^2}{n\hat{\omega}_{a=0,t=0}}\right]\mathbb{V}\left(Y_{a=0}\right) \\
    &\quad+ \mathbb{E}\left[\frac{(\hat{\mathbb{E}}\left(A_{t=1}\right)-\hat{\mathbb{E}}\left(A_{t=0}\right))^2\hat{\pi}^2}{n\hat{\omega}_{a=1,t=1}}\right]\mathbb{V}\left(Y_{a=1}\right)
    + \mathbb{E}\left[\frac{(\hat{\mathbb{E}}\left(A_{t=1}\right)-\hat{\mathbb{E}}\left(A_{t=0}\right))^2\hat{\pi}^2}{n\hat{\omega}_{a=0,t=1}}\right]\mathbb{V}\left(Y_{a=0}\right) \\
    &= \frac{1}{n}\left(\mathbb{V}\left(Y_{a=1}\right)\left(\mathbb{E}\left[\frac{(\hat{\mathbb{E}}\left(A_{t=1}\right)-\hat{\mathbb{E}}\left(A_{t=0}\right))^2(1-\hat{\pi})^2}{\hat{\omega}_{a=1,t=0}}\right] + \mathbb{E}\left[\frac{(\hat{\mathbb{E}}\left(A_{t=1}\right) -\hat{\mathbb{E}}\left(A_{t=0}\right))^2\hat{\pi}^2}{\hat{\omega}_{a=1,t=1}}\right]\right) \right. \\
    &\left. \quad + \mathbb{V}\left(Y_{a=0}\right)\left(\mathbb{E}\left[\frac{(\hat{\mathbb{E}}\left(A_{t=1}\right)-\hat{\mathbb{E}}\left(A_{t=0}\right))^2(1-\hat{\pi})^2}{\hat{\omega}_{a=0,t=0}}\right] + \mathbb{E}\left[\frac{(\hat{\mathbb{E}}\left(A_{t=1}\right) -\hat{\mathbb{E}}\left(A_{t=0}\right))^2\hat{\pi}^2}{\hat{\omega}_{a=0,t=1}}\right]\right)\right)
\end{align*}

\textbf{Computation of $\circled{F}$} \\
Given definition from Equation~\ref{eq:two-sided-cfd-def}, we can write term $\circled{F}$ as:

\begin{equation}
\begin{aligned}
    \text{Var}&\left(\mathbb{E}\left[\hat{\Phi}^{\text{CFD}}_{\text{ts}} | \{a_k, t_k\}_k\right]\right) \\
    &=\text{Var}\left(\mathbb{E}\left[((\hat{\mathbb{E}}\left(Y_{a=1, t=0}\right) - \hat{\mathbb{E}}\left(Y_{a=0, t=0}\right))(1-\hat{\pi}) + (\hat{\mathbb{E}}\left(Y_{a=1, t=1}\right) - \hat{\mathbb{E}}\left(Y_{a=0, t=1}\right))\hat{\pi})(\hat{\mathbb{E}}\left(A_{t=1}\right)-\hat{\mathbb{E}}\left(A_{t=0}\right)) | \{a_k,  t_k\}_k)\right]\right) \\
    &=\text{Var}\left((\hat{\mathbb{E}}\left(A_{t=1}\right)-\hat{\mathbb{E}}\left(A_{t=0}\right))\left(
    \underbrace{\mathbb{E}\left[(\hat{\mathbb{E}}\left(Y_{a=1, t=0}\right) - \hat{\mathbb{E}}\left(Y_{a=0, t=0}\right))(1-\hat{\pi}) | \{a_k,  t_k\}_k\right]}_{\circled{F1}} + \right. \right. \\
     & \quad \left. \left. \underbrace{\mathbb{E}\left[(\hat{\mathbb{E}}\left(Y_{a=1, t=1}\right) - \hat{\mathbb{E}}\left(Y_{a=0, t=1}\right))\hat{\pi} | \{a_k,  t_k\}_k\right]}_{\circled{F2}}\right)\right)
\end{aligned}
\label{eq:cfd-two-sided-f}
\end{equation}

For $\circled{F1}$ we write:
\begin{equation}
\begin{aligned}
    \mathbb{E}&\left[(\hat{\mathbb{E}}\left(Y_{a=1, t=0}\right) - \hat{\mathbb{E}}\left(Y_{a=0, t=0}\right))(1-\hat{\pi}) | \{a_k,  t_k\}_k\right] 
    = (1-\hat{\pi})\mathbb{E}\left[\hat{\mathbb{E}}\left(Y_{a=1, t=0}\right) - \hat{\mathbb{E}}\left(Y_{a=0, t=0}\right) | \{a_k,  t_k\}_k\right] \\
    &= (1-\hat{\pi})\left(\underbrace{\mathbb{E}\left[\hat{\mathbb{E}}\left(Y_{a=1, t=0}\right) | \{a_k,  t_k\}_k\right]}_{\circled{F1.1}} - \underbrace{\mathbb{E}\left[\hat{\mathbb{E}}\left(Y_{a=0, t=0}\right) | \{a_k,  t_k\}_k\right]}_{\circled{F1.2}}\right)
\end{aligned}
\label{eq:cfd-two-sided-f1}
\end{equation}

For $\circled{F1.1}$ we write:
\begin{align*}
    \mathbb{E}&\left[\hat{\mathbb{E}}\left(Y_{a=1, t=0}\right) | \{a_k,  t_k\}_k\right] = \mathbb{E}\left[\frac{\sum\limits^n_{i=1}a_i(1-t_i)y_i}{\sum\limits_{i=1}^{n}a_i(1-t_i)} | \{a_k,  t_k\}_k\right] = \frac{1}{\sum\limits_{i=1}^{n}a_i(1-t_i)}\mathbb{E}\left[\sum\limits^n_{i=1}a_i(1-t_i)y_i | \{a_k,  t_k\}_k\right] \\
    &= \frac{1}{\sum\limits_{i=1}^{n}a_i(1-t_i)}\sum\limits^n_{i=1}\mathbb{E}\left[a_i(1-t_i)y_i | \{a_k,  t_k\}_k\right] = \frac{1}{\sum\limits_{i=1}^{n}a_i(1-t_i)}\sum\limits^n_{i=1}a_i(1-t_i)\mathbb{E}\left[y_i | \{a_k,  t_k\}_k\right] \\
    &= \frac{1}{\sum\limits_{i=1}^{n}a_i(1-t_i)}\sum\limits^n_{i=1}a_i(1-t_i)\underbrace{\mathbb{E}\left[y_i | a_i,  t_i\right]}_{Y_{a=1, t=0} \text{ (\textit{i.i.d.} and multiply with } a_i(1-t_i)\text{)}} \\
    &= \frac{\mathbb{E}\left(Y_{a=1, t=0}\right)}{\sum\limits_{i=1}^{n}a_i(1-t_i)}\sum\limits^n_{i=1}a_i(1-t_i) = \mathbb{E}\left(Y_{a=1, t=0}\right) = \mathbb{E}\left(Y_{a=1}\right)
\end{align*}

Analogously, we get expression for $\circled{F1.2}$:
\begin{align*}
    \mathbb{E}\left[\hat{\mathbb{E}}\left(Y_{a=0, t=0}\right) | \{a_k,  t_k\}_k\right] = \mathbb{E}\left(Y_{a=0}\right)
\end{align*}

Injecting $\circled{F1.1}$ and $\circled{F1.2}$ into Equation~\ref{eq:cfd-two-sided-f1}, we get expression for $\circled{F1}$ as:
\begin{align*}
    \mathbb{E}&\left[(\hat{\mathbb{E}}\left(Y_{a=1, t=0}\right)-\hat{\mathbb{E}}\left(Y_{a=0, t=0}\right))(1-\hat{\pi}) | \{a_k,  t_k\}_k\right] = (1-\hat{\pi})\left(\mathbb{E}\left(Y_{a=1}\right)-\mathbb{E}\left(Y_{a=0}\right)\right)
\end{align*}

Analogously, we get expression for $\circled{F2}$ as:
\begin{align*}
    \mathbb{E}\left[(\hat{\mathbb{E}}\left(Y_{a=1, t=1}\right) - \hat{\mathbb{E}}\left(Y_{a=0, t=1}\right))\hat{\pi} | \{a_k,  t_k\}_k\right] = \hat{\pi}\left(\mathbb{E}\left(Y_{a=1}\right)-\mathbb{E}\left(Y_{a=0}\right)\right)
\end{align*}

Injecting $\circled{F1}, \circled{F2}$ into Equation~\ref{eq:cfd-two-sided-f}, we get expression for $\circled{F}$ as:
\begin{equation}
\begin{aligned}
\text{Var}&\left(\mathbb{E}\left[\hat{\Phi}^{\text{CFD}}_{\text{ts}} | \{a_k, t_k\}_k\right]\right) \\
&= \text{Var}\left((\hat{\mathbb{E}}\left(A_{t=1}\right)-\hat{\mathbb{E}}\left(A_{t=0}\right))\left(
    (1-\hat{\pi})\left(\mathbb{E}\left(Y_{a=1}\right)-\mathbb{E}\left(Y_{a=0}\right)\right) + \hat{\pi}\left(Y_{a=1, t=1}-\mathbb{E}\left(Y_{a=0}\right)\right)\right)\right) \\
    &= \text{Var}\left((\hat{\mathbb{E}}\left(A_{t=1}\right)-\hat{\mathbb{E}}\left(A_{t=0}\right))\left(\mathbb{E}\left(Y_{a=1}\right)-\mathbb{E}\left(Y_{a=0}\right)\right)\right) = \Delta_Y^2\underbrace{\text{Var}\left(\hat{\mathbb{E}}\left(A_{t=1}\right)-\hat{\mathbb{E}}\left(A_{t=0}\right)\right)}_{\circled{F3}}
\end{aligned}
\label{eq:cfd-two-sided-D-term}
\end{equation}

Analogous to the derivation for $\text{Var}\left(\hat{\Phi^{\text{SBD}}}\right)$ (Section~\ref{sec:baseline-variance}), we can write expression for $\circled{F3}$ as: 
\begin{align*}
    \text{Var}\left(\hat{\mathbb{E}}\left(A_{t=1}\right)-\hat{\mathbb{E}}\left(A_{t=0}\right)\right) = \frac{\mathbb{V}\left(A_{t=1}\right)}{n\pi}  + \frac{\mathbb{V}\left(A_{t=0}\right)}{n(1-\pi)}
\end{align*}

Injecting $\circled{F3}$ into Equation~\ref{eq:cfd-two-sided-D-term}, we get expression for $\circled{F}$ as:
\begin{align*}
    \text{Var}&\left(\mathbb{E}\left[\hat{\Phi}^{\text{CFD}}_{\text{ts}} | \{a_k, t_k\}_k\right]\right) = \left(\mathbb{E}\left(Y_{a=1}\right)-\mathbb{E}\left(Y_{a=0}\right)\right)^2\left(\frac{\mathbb{V}\left(A_{t=1}\right)}{n\pi}  + \frac{\mathbb{V}\left(A_{t=0}\right)}{n(1-\pi)}\right)
\end{align*}

Injecting $\circled{E}, \circled{F}$ back into Equation~\ref{eq:cfd-two-sided-start}, we get variance for $\hat{\Phi}^{\text{CFD}}_{\text{ts}}$ as:
\begin{align*}
    \text{Var}(\hat{\Phi}^{\text{CFD}}_{\text{ts}})=& \frac{1}{n}\left(
    \mathbb{V}\left(Y_{a=1}\right)\left(\mathbb{E}\left[\frac{(\hat{\mathbb{E}}\left(A_{t=1}\right)-\hat{\mathbb{E}}\left(A_{t=0}\right))^2(1-\hat{\pi})^2}{\hat{\omega}_{a=1, t=0}}\right] + \mathbb{E}\left[\frac{(\hat{\mathbb{E}}\left(A_{t=1}\right) -\hat{\mathbb{E}}\left(A_{t=0}\right))^2\hat{\pi}^2}{\hat{\omega}_{a=1, t=1}}\right]\right) \right.\\
    &\left. + \mathbb{V}\left(Y_{a=0}\right)\left(\mathbb{E}\left[\frac{(\hat{\mathbb{E}}\left(A_{t=1}\right)-\hat{\mathbb{E}}\left(A_{t=0}\right))^2(1-\hat{\pi})^2}{\hat{\omega}_{a=0, t=0}}\right] + \mathbb{E}\left[\frac{(\hat{\mathbb{E}}\left(A_{t=1}\right) -\hat{\mathbb{E}}\left(A_{t=0}\right))^2\hat{\pi}^2}{\hat{\omega}_{a=0, t=1}}\right]\right) \right.\\
    &\left. + \Delta_Y^2\left(\frac{\mathbb{V}\left(A_{t=1}\right)}{\pi}  + \frac{\mathbb{V}\left(A_{t=0}\right)}{1-\pi}\right)\right)
\end{align*}

\subsection{Derive asymptotic variance upper and lower bounds}\label{sec:appendix:variance:bound}

From previous sections, we get variance for $\hat{\Phi}^{\text{SBD}}_{\text{os}}$, 
$\hat{\Phi}^{\text{SBD}}_{\text{ts}}$, $\hat{\Phi}^{\text{CFD}}_{\text{os}}$, and $\hat{\Phi}^{\text{CFD}}_{\text{ts}}$ as;
\begin{align*}
    \text{Var}\left(\hat{\Phi}_{\text{OS}}^{\text{SBD}}\right) &= \frac{\left(1-\mathbb{E}\left(A_{t=1}\right)\right)\mathbb{V}\left(Y_{a=0}\right) + \mathbb{E}\left(A_{t=1}\right)\mathbb{V}\left(Y_{a=1}\right) + \Delta_Y^2\mathbb{V}\left(A_{t=1}\right)}{n\pi}  + \frac{\mathbb{V}\left(Y_{a=0}\right)}{n(1-\pi)} \\
    \text{Var}\left(\hat{\Phi}_{\text{TS}}^{\text{SBD}}\right) &= \frac{\left(1-\mathbb{E}\left(A_{t=1}\right)\right)\mathbb{V}\left(Y_{a=0}\right) + \mathbb{E}\left(A_{t=1}\right)\mathbb{V}\left(Y_{a=1}\right) + \Delta_Y^2\mathbb{V}\left(A_{t=1}\right)}{n\pi}  + \\
    &\quad\ \frac{\left(1-\mathbb{E}\left(A_{t=0}\right)\right)\mathbb{V}\left(Y_{a=0}\right) + \mathbb{E}\left(A_{t=0}\right)\mathbb{V}\left(Y_{a=1}\right) + \Delta_Y^2\mathbb{V}\left(A_{t=0}\right)}{n(1-\pi)}
\end{align*}
\begin{align*}
    \text{Var}(\hat{\Phi}^{\text{CFD}}_{\text{os}}) &= \frac{1}{n}\left(\mathbb{V}\left(Y_{a=0}\right)\left(\mathbb{E}\left[\frac{\hat{\mathbb{E}}\left(A_{t=1}\right)^2(1-\hat{\pi})^2}{\hat{\omega}_{a=0,t=0}}\right] + \mathbb{E}\left[\frac{\hat{\mathbb{E}}\left(A_{t=1}\right)^2\hat{\pi}^2}{\hat{\omega}_{a=0,t=1}}\right]\right) + \mathbb{V}\left(Y_{a=1}\right)\mathbb{E}\left[\frac{\hat{\mathbb{E}}\left(A_{t=1}\right)^2}{\hat{\omega}_{a=1,t=1}}\right] 
      + \right. \\
    &\left. \quad \mathbb{V}\left(A_{t=1}\right)\frac{\Delta_Y^2}{\pi}\right) \\
    \text{Var}(\hat{\Phi}^{\text{CFD}}_{\text{ts}})=& \frac{1}{n}\left(
    \mathbb{V}\left(Y_{a=1}\right)\left(\mathbb{E}\left[\frac{(\hat{\mathbb{E}}\left(A_{t=1}\right)-\hat{\mathbb{E}}\left(A_{t=0}\right))^2(1-\hat{\pi})^2}{\hat{\omega}_{a=1, t=0}}\right] + \mathbb{E}\left[\frac{(\hat{\mathbb{E}}\left(A_{t=1}\right) -\hat{\mathbb{E}}\left(A_{t=0}\right))^2\hat{\pi}^2}{\hat{\omega}_{a=1, t=1}}\right]\right) \right.\\
    &\left. + \mathbb{V}\left(Y_{a=0}\right)\left(\mathbb{E}\left[\frac{(\hat{\mathbb{E}}\left(A_{t=1}\right)-\hat{\mathbb{E}}\left(A_{t=0}\right))^2(1-\hat{\pi})^2}{\hat{\omega}_{a=0, t=0}}\right] + \mathbb{E}\left[\frac{(\hat{\mathbb{E}}\left(A_{t=1}\right) -\hat{\mathbb{E}}\left(A_{t=0}\right))^2\hat{\pi}^2}{\hat{\omega}_{a=0, t=1}}\right]\right) \right.\\
    &\left. + \Delta_Y^2\left(\frac{\mathbb{V}\left(A_{t=1}\right)}{\pi}  + \frac{\mathbb{V}\left(A_{t=0}\right)}{1-\pi}\right)\right)
\end{align*}

\subsubsection{Asymptotic lower bounds of $\text{Var}(\hat{\Phi}^{\text{SBD}})$}

Based on the derivation for $\hat{\Phi}^{\text{SBD}}_{\text{os}}$ and
$\hat{\Phi}^{\text{SBD}}_{\text{ts}}$, we get following with a slight rewriting:
\begin{align*}
    n\text{Var}\left(\hat{\Phi}_{\text{OS}}^{\text{SBD}}\right) &= \frac{\left(1-\mathbb{E}\left(A_{t=1}\right)\right)\mathbb{V}\left(Y_{a=0}\right) + \mathbb{E}\left(A_{t=1}\right)\mathbb{V}\left(Y_{a=1}\right) + \Delta_Y^2\mathbb{V}\left(A_{t=1}\right)}{\pi}  + \frac{\mathbb{V}\left(Y_{a=0}\right)}{(1-\pi)} \\
    n\text{Var}\left(\hat{\Phi}_{\text{TS}}^{\text{SBD}}\right) &= \frac{\left(1-\mathbb{E}\left(A_{t=1}\right)\right)\mathbb{V}\left(Y_{a=0}\right) + \mathbb{E}\left(A_{t=1}\right)\mathbb{V}\left(Y_{a=1}\right) + \Delta_Y^2\mathbb{V}\left(A_{t=1}\right)}{\pi}  + \\
    &\quad\ \frac{\left(1-\mathbb{E}\left(A_{t=0}\right)\right)\mathbb{V}\left(Y_{a=0}\right) + \mathbb{E}\left(A_{t=0}\right)\mathbb{V}\left(Y_{a=1}\right) + \Delta_Y^2\mathbb{V}\left(A_{t=0}\right)}{(1-\pi)}
\end{align*}

Then, by the law of large numbers, we get:
\begin{align*}
    \lim_{n\rightarrow \infty}n\text{Var}(\hat{\Phi}^{\text{SBD}}_{\text{os}}) &= \frac{1}{\pi} \left(\left(1-\mathbb{E}\left(A_{t=1}\right)\right)\mathbb{V}\left(Y_{a=0}\right) + \mathbb{E}\left(A_{t=1}\right)\mathbb{V}\left(Y_{a=1}\right) + \Delta_Y^2\mathbb{V}\left(A_{t=1}\right)\right) 
     +  \frac{1}{1-\pi} \mathbb{V}\left(Y_{a=0}\right) \\
    \lim_{n\rightarrow \infty}n\text{Var}(\hat{\Phi}^{\text{SBD}}_{\text{ts}}) &= \frac{1}{\pi} \left(\left(1-\mathbb{E}\left(A_{t=1}\right)\right)\mathbb{V}\left(Y_{a=0}\right) + \mathbb{E}\left(A_{t=1}\right)\mathbb{V}\left(Y_{a=1}\right) + \Delta_Y^2\mathbb{V}\left(A_{t=1}\right)\right) +\\
    & \quad  \frac{1}{1-\pi}\left(\left(1-\mathbb{E}\left(A_{t=0}\right)\right)\mathbb{V}\left(Y_{a=0}\right) + \mathbb{E}\left(A_{t=0}\right)\mathbb{V}\left(Y_{a=1}\right) + \Delta_Y^2\mathbb{V}\left(A_{t=0}\right)\right) \\
\end{align*}

Because of the positivity assumption (Assumption~\ref{asup:positivity}), we have:
\begin{align*}
    \pi < 1, \quad 1-\pi < 1
\end{align*}

Substituting above inequalities back to $ \lim\limits_{n\rightarrow \infty}n\text{Var}(\hat{\Phi}^{\text{SBD}}_{\text{os}})$ and $ \lim\limits_{n\rightarrow \infty}n\text{Var}(\hat{\Phi}^{\text{SBD}}_{\text{ts}})$, we get following asymptotic variance lower bounds:
\begin{align*}
    \lim_{n\rightarrow \infty}n\text{Var}(\hat{\Phi}^{\text{SBD}}_{\text{os}}) &>  \left(2-\mathbb{E}\left(A_{t=1}\right)\right)\mathbb{V}\left(Y_{a=0}\right) + \mathbb{E}\left(A_{t=1}\right)\mathbb{V}\left(Y_{a=1}\right) + \Delta_Y^2\mathbb{V}\left(A_{t=1}\right) \\
    \lim_{n\rightarrow \infty}n\text{Var}(\hat{\Phi}^{\text{SBD}}_{\text{ts}}) &> \left(\left(1-\mathbb{E}\left(A_{t=1}\right)\right)\mathbb{V}\left(Y_{a=0}\right) + \mathbb{E}\left(A_{t=1}\right)\mathbb{V}\left(Y_{a=1}\right) + \Delta_Y^2\mathbb{V}\left(A_{t=1}\right)\right) +\\
    & \quad \left(\left(1-\mathbb{E}\left(A_{t=0}\right)\right)\mathbb{V}\left(Y_{a=0}\right) + \mathbb{E}\left(A_{t=0}\right)\mathbb{V}\left(Y_{a=1}\right) + \Delta_Y^2\mathbb{V}\left(A_{t=0}\right)\right) \\
    &= \left(2-\mathbb{E}\left(A_{t=1}\right) -\mathbb{E}\left(A_{t=0}\right)\right)\mathbb{V}\left(Y_{a=0}\right) + \left(\mathbb{E}\left(A_{t=0}\right)+\mathbb{E}\left(A_{t=1}\right)\right)\mathbb{V}\left(Y_{a=1}\right) + \\
    &\quad \Delta_Y^2\left(\mathbb{V}\left(A_{t=0}\right) + \mathbb{V}\left(A_{t=1}\right)\right)
\end{align*}

\subsubsection{Asymptotic variance upper bounds of $\text{Var}(\hat{\Phi}^{\text{CFD}})$}

Based on the derivation for $\hat{\Phi}^{\text{CFD}}_{\text{os}}$ and
$\hat{\Phi}^{\text{CFD}}_{\text{ts}}$, we get following with a slight rewriting:
\begin{align*}
    n\text{Var}(\hat{\Phi}^{\text{CFD}}_{\text{os}}) &= \mathbb{V}\left(Y_{a=0}\right)\left(\mathbb{E}\left[\frac{\hat{\mathbb{E}}\left(A_{t=1}\right)^2(1-\hat{\pi})^2}{\hat{\omega}_{a=0,t=0}}\right] + \mathbb{E}\left[\frac{\hat{\mathbb{E}}\left(A_{t=1}\right)^2\hat{\pi}^2}{\hat{\omega}_{a=0,t=1}}\right]\right) + \mathbb{V}\left(Y_{a=1}\right)\mathbb{E}\left[\frac{\hat{\mathbb{E}}\left(A_{t=1}\right)^2}{\hat{\omega}_{a=1,t=1}}\right] 
      +  \\
    & \quad \mathbb{V}\left(A_{t=1}\right)\frac{\Delta_Y^2}{\pi} \\
    n\text{Var}(\hat{\Phi}^{\text{CFD}}_{\text{ts}})=& 
    \mathbb{V}\left(Y_{a=1}\right)\left(\mathbb{E}\left[\frac{(\hat{\mathbb{E}}\left(A_{t=1}\right)-\hat{\mathbb{E}}\left(A_{t=0}\right))^2(1-\hat{\pi})^2}{\hat{\omega}_{a=1, t=0}}\right] + \mathbb{E}\left[\frac{(\hat{\mathbb{E}}\left(A_{t=1}\right) -\hat{\mathbb{E}}\left(A_{t=0}\right))^2\hat{\pi}^2}{\hat{\omega}_{a=1, t=1}}\right]\right) \\
    & + \mathbb{V}\left(Y_{a=0}\right)\left(\mathbb{E}\left[\frac{(\hat{\mathbb{E}}\left(A_{t=1}\right)-\hat{\mathbb{E}}\left(A_{t=0}\right))^2(1-\hat{\pi})^2}{\hat{\omega}_{a=0, t=0}}\right] + \mathbb{E}\left[\frac{(\hat{\mathbb{E}}\left(A_{t=1}\right) -\hat{\mathbb{E}}\left(A_{t=0}\right))^2\hat{\pi}^2}{\hat{\omega}_{a=0, t=1}}\right]\right) \\
    & + \Delta_Y^2\left(\frac{\mathbb{V}\left(A_{t=1}\right)}{\pi}  + \frac{\mathbb{V}\left(A_{t=0}\right)}{1-\pi}\right)
\end{align*}

Then, by the law of large numbers, we get:
\begin{equation}
\begin{aligned}
    \lim_{n\rightarrow \infty}n\text{Var}(\hat{\Phi}^{\text{CFD}}_{\text{os}}) &= 
    \mathbb{V}\left(Y_{a=0}\right)\left(\frac{\Delta_A^2(1-\pi)^2}{\omega_{a=0,t=0}} + \frac{\Delta_A^2\pi^2}{\omega_{a=0, t=1}}\right) + +\mathbb{V}\left(Y_{a=1}\right)
      \frac{\Delta_A^2}{\omega_{a=1, t=1}}
      + \mathbb{V}\left(A_{t=1}\right)\frac{\Delta_Y^2}{\pi} \\
    \lim_{n\rightarrow \infty}n\text{Var}(\hat{\Phi}^{\text{CFD}}_{\text{ts}}) &= \mathbb{V}\left(Y_{a=1}\right)\left(\frac{\Delta_A^2(1-\pi)^2}{\omega_{a=1, t=0}} + \frac{\Delta_A^2\pi^2}{\omega_{a=1,t=1}}\right) + \mathbb{V}\left(Y_{a=0}\right)\left(\frac{\Delta_A^2(1-\pi)^2}{\omega_{a=0, t=0}}
    + \frac{\Delta_A^2\pi^2}{\omega_{a=0,t=1}}\right) \\
    &\quad + \Delta_Y^2\left(\frac{\mathbb{V}\left(A_{t=1}\right)}{\pi} + \frac{\mathbb{V}\left(A_{t=0}\right)}{1-\pi}\right)
\end{aligned}
\label{eq:cfd-as-var}
\end{equation}

As shown in Equation~\ref{eq:cfd-as-var}, the asymptotic variance of CFD estimations depends on $\pi$, $1-\pi$, $\omega_{a=0, t=0}$, $\omega_{a=1, t=0}$, $\omega_{a=0, t=1}$, and $\omega_{a=1, t=1}$ as their denominators. The larger these quantities are, the smaller the asymptotic variances are, which intuitively means more samples for each nuisance parameter will allow better estimation.

To ensure reasonable estimation quality, we define the minimum probability for the \textit{possible} treatment assignment and intake combinations as $\rho$. Formally, we write: 

\begin{equation}
\begin{aligned}
    \rho &= \begin{cases}
    \min\limits_{\substack{t',a' \in \{0,1\} \\ (a', t') \neq (1,0)}}\left\{ P(a', t' \mid x_0) \right\} 
    & \text{if one-sided non-adherence} \\[2mm]
    \min\limits_{t',a' \in \{0,1\}}\left\{ P(a', t' \mid x_0) \right\} 
    & \text{if two-sided non-adherence}
    \end{cases}
\end{aligned}
\label{eq:epsilon}
\end{equation}

Substitute Equation~\ref{eq:epsilon} into Equation~\ref{eq:cfd-as-var},  we get the following asymptotic variance upper bounds:
\begin{align*}
    \lim_{n\rightarrow \infty}n\text{Var}(\hat{\Phi}^{\text{CFD}}_{\text{os}}) &\le 
    \frac{1}{\rho}\left(\mathbb{V}\left(Y_{a=0}\right)\left(\Delta_A^2((1-\pi)^2 + \pi^2)\right) +\mathbb{V}\left(Y_{a=1}\right) \Delta_A^2 + \Delta_Y^2 \mathbb{V}\left(A_{t=1}\right)\right) \\
    \lim_{n\rightarrow \infty}n\text{Var}(\Phi^{\text{CFD}}_{\text{ts}}) &\le \frac{1}{\rho}\left(\left(\mathbb{V}\left(Y_{a=1}\right)+ \mathbb{V}\left(Y_{a=0}\right)\right)\Delta_A^2((1-\pi)^2 +\pi^2) + \left(\mathbb{V}\left(A_{t=1}\right) + \mathbb{V}\left(A_{t=0}\right)\right)\Delta_Y^2\right)
\end{align*}

Lastly, we simplify the upper bound by using the fact that $(1-\pi)^2+\pi^2 < 1$ when $0<\pi<1$ to get the following upper bounds:
\begin{align*}
    \lim_{n\rightarrow \infty}n\text{Var}(\hat{\Phi}^{\text{CFD}}_{\text{os}}) &< 
    \frac{1}{\rho}\left(\left(\mathbb{V}\left(Y_{a=0}\right) + \mathbb{V}\left(Y_{a=1}\right)\right)\Delta_A^2 + \Delta_Y^2 \mathbb{V}\left(A_{t=1}\right)\right) \\
    \lim_{n\rightarrow \infty}n\text{Var}(\hat{\Phi}^{\text{CFD}}_{\text{ts}}) &< \frac{1}{\rho}\left(\left(\mathbb{V}\left(Y_{a=0}\right) + \mathbb{V}\left(Y_{a=1}\right)\right)\Delta_A^2 + \left(\mathbb{V}\left(A_{t=1}\right) + \mathbb{V}\left(A_{t=0}\right)\right)\Delta_Y^2\right)
\end{align*}

\subsection{Compare SBD and CFD's variance bounds directly}\label{sec:appendix:variance:compare}
From the previous sections, we derive SBD's variance lower bounds under one-sided and two-sided non-adherence as follows:
\begin{align*}
    \lim_{n\rightarrow \infty}n\text{Var}(\hat{\Phi}^{\text{SBD}}_{\text{os}}) &>  \left(2-\mathbb{E}\left(A_{t=1}\right)\right)\mathbb{V}\left(Y_{a=0}\right) + \mathbb{E}\left(A_{t=1}\right)\mathbb{V}\left(Y_{a=1}\right) + \Delta_Y^2\mathbb{V}\left(A_{t=1}\right) \\
    \lim_{n\rightarrow \infty}n\text{Var}(\hat{\Phi}^{\text{SBD}}_{\text{ts}}) &> \left(2-\mathbb{E}\left(A_{t=1}\right) -\mathbb{E}\left(A_{t=0}\right)\right)\mathbb{V}\left(Y_{a=0}\right) + \left(\mathbb{E}\left(A_{t=0}\right)+\mathbb{E}\left(A_{t=1}\right)\right)\mathbb{V}\left(Y_{a=1}\right) + \\
    &\quad \Delta_Y^2\left(\mathbb{V}\left(A_{t=0}\right) + \mathbb{V}\left(A_{t=1}\right)\right)
\end{align*}

And CFD's variance upper bounds under one-sided and two-sided non-adherence are as follows:
\begin{align*}
    \lim_{n\rightarrow \infty}n\text{Var}(\hat{\Phi}^{\text{CFD}}_{\text{os}}) &< 
    \frac{1}{\rho}\left(\left(\mathbb{V}\left(Y_{a=0}\right) + \mathbb{V}\left(Y_{a=1}\right)\right) \Delta_A^2 + \Delta_Y^2 \mathbb{V}\left(A_{t=1}\right)\right) \\
    \lim_{n\rightarrow \infty}n\text{Var}(\hat{\Phi}^{\text{CFD}}_{\text{ts}}) &< \frac{1}{\rho}\left(\left(\mathbb{V}\left(Y_{a=0}\right) + \mathbb{V}\left(Y_{a=1}\right)\right)\Delta_A^2 + \left(\mathbb{V}\left(A_{t=1}\right) + \mathbb{V}\left(A_{t=0}\right)\right)\Delta_Y^2\right)
\end{align*}

To simplify the comparison, we start by assuming that treatment intake under different treatment assignments, and outcome under different treatment intakes have the same variance:
\begin{equation}
\begin{aligned}
    \mathbb{V}\left(A_{t=0}\right) = \mathbb{V}\left(A_{t=1}\right) = V_A \\
    \mathbb{V}\left(Y_{a=0}\right) = \mathbb{V}\left(Y_{a=1}\right) = V_Y \\
\end{aligned}
\label{eq:equal_variance}
\end{equation}

Which leads to the following simplified lower bounds for SBD:
\begin{equation}
\begin{aligned}
\lim_{n\rightarrow \infty}n\text{Var}(\hat{\Phi}^{\text{SBD}}_{\text{os}}) &> 2V_Y + V_A\Delta_Y^2 \\
\lim_{n\rightarrow \infty}n\text{Var}(\hat{\Phi}^{\text{SBD}}_{\text{ts}}) &>  2V_Y + 2V_A\Delta_Y^2
\end{aligned}\label{eq:sbd-bounds}
\end{equation}

And the following simplified upper bounds for SBD:
\begin{equation}
\begin{aligned}
    \lim_{n\rightarrow \infty}n\text{Var}(\hat{\Phi}^{\text{CFD}}_{\text{os}}) &< 
    \frac{1}{\rho}\left(2V_Y\Delta_A^2 + V_A\Delta_Y^2\right) \\
    \lim_{n\rightarrow \infty}n\text{Var}(\hat{\Phi}^{\text{CFD}}_{\text{ts}}) &< \frac{1}{\rho}(2V_Y\Delta_A^2 + 2V_A\Delta_Y^2)
\end{aligned}\label{eq:cfd-bounds}
\end{equation}

By taking the difference between SBD's lower bound and CFD's upper bound, we get the following:
\begin{align*}
    \lim_{n\rightarrow \infty}n\left(\mathrm{Var}(\hat{\Phi}^{\mathrm{SBD}}_{\mathrm{os}})-\mathrm{Var}(\hat{\Phi}^{\mathrm{CFD}}_{\mathrm{os}})\right)  &> 2V_Y - \frac{2}{\rho}V_Y\Delta_A^2 - \frac{1-\rho}{\rho}V_A\Delta_Y^2 \\
    \lim_{n\rightarrow \infty}n\left(\mathrm{Var}(\hat{\Phi}^{\mathrm{SBD}}_{\mathrm{ts}})-\mathrm{Var}(\hat{\Phi}^{\mathrm{CFD}}_{\mathrm{ts}})\right) &> 2V_Y - \frac{2}{\rho}V_Y\Delta_A^2 - \frac{2(1-\rho)}{\rho}V_A\Delta_Y^2
\end{align*}

Finally, applying the big-O notation, we get the following asymptotic variance difference bound for both one-sided and two-sided non-adherence:
\begin{align*}
    \lim_{n\rightarrow \infty}n\left(\mathrm{Var}(\hat{\Phi}^{\mathrm{SBD}})-\mathrm{Var}(\hat{\Phi}^{\mathrm{CFD}})\right)  &> \mathcal{O}\left(V_Y-\frac{V_Y\Delta_A^2}{\rho} - \frac{(1-\rho)V_A\Delta_Y^2}{\rho}\right)
\end{align*}

\section{Variance Reduction Bound Numerical Visualization}\label{sec:appendix:num_viz}
To compare the variance bounds difference between SBD and CFD without assuming Equation~\ref{eq:equal_variance}, we consider a special case where outcome $Y$ is binary. 
In such a scenario, we apply a simple trick to represent variance in terms of expectation.
For instance:
\begin{align*}
    \mathbb{V}\left(Y_{a=0}\right) = \mathbb{E}\left(Y_{a=0}^2\right) - \mathbb{E}\left(Y_{a=0}\right)^2 = \mathbb{E}\left(Y_{a=0}\right) - \mathbb{E}\left(Y_{a=0}\right)^2 = \mathbb{E}\left(Y_{a=0}\right)\left(1-\mathbb{E}\left(Y_{a=0}\right)\right)
\end{align*}
This trick allows us to replace all variance terms in the variance bounds with their corresponding expectations, effectively reducing the number of variables that our visualization needs to consider.

\section{Implementation Details}\label{sec:appendix:implementation}

\begin{figure*}
  \centering
  \includegraphics[width=1.0\linewidth,alt={lobster_appendix}]{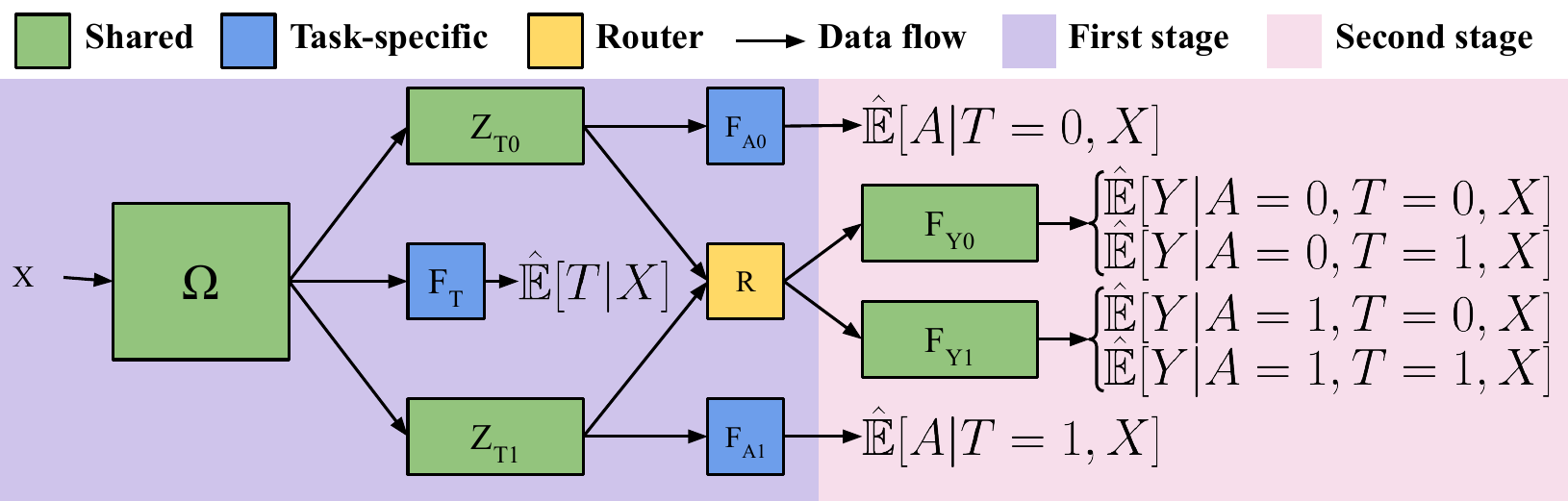}
  \captionof{figure}{Illustration of our proposed \methodname\ architecture,which jointly models all required nuisance parameters}\vspace{-4mm}
  \label{fig:appendix:cfd:lobster}
\end{figure*}

\subsection{T-learner Details}\label{sec:appendix:implementation:tlearner}
Each T-learner is implemented with two independent multi-layer perceptrons (MLPs) with three hidden ELU-activated layers of dimensions 300. 
Only one T-learner (two MLPs) is used for the SBD-based method. Each MLP in the T-learner estimates one type of potential outcome. For the CFD-based method, seven MLPs are used (one for propensity score, two for treatment intake prediction, and four for outcome prediction).

\subsection{\methodname\ Details}\label{sec:appendix:implementation:lobster}
As shown in Figure~\ref{fig:appendix:cfd:lobster}, \methodname\ is composed of five shared neural network modules and three task-specific neural network modules.
Each shared module is implemented using an MLP with three hidden ELU-activated layers of dimensions 300. 
Each task-specific module is implemented using an MLP with three hidden ELU-activated layers of dimensions 100. 
The router module takes in two different hidden representations from the first stage (\textit{i.e.}, representations for predicting $\mathbb{E}[A | T=0, X]$ or $\mathbb{E}[A | T=1, X]$), and dynamically determines which representation to output for the second stage based on which outcome is being predicted. 
For example, when predicting $\mathbb{E}[Y | A=0, T=0, X]$ and $\mathbb{E}[Y | A=1, T=0, X]$, the router module outputs hidden representation for predicting $\mathbb{E}[A | T=0, X]$.
When predicting $\mathbb{E}[Y | A=0, T=1, X]$ and $\mathbb{E}[Y | A=1, T=1, X]$, the router module outputs hidden representations for predicting $\mathbb{E}[A | T=1, X]$.
This design allows for sharing the same hidden MLP representations when predicting outcomes conditioned on the same type of treatment assignment ($T$).
Lastly, we conceptually break down \methodname\ into first stage and second stage in the illustration; however, the entire model is end-to-end trained with a multi-task objective function. 
This design enables efficient optimization in practice.

\subsection{Training Details}\label{sec:appendix:training}
80\% of all data is randomly selected for training. Among all tthe raining data, another 20\% is randomly selected as the validation set. All neural networks are trained using Adam optimizer~\citep{kingma2014adam} with a fixed learning rate of $10^{-4}$ for up to 1000 epochs. L2 regularization is selected via grid search over $\{10^{-2}, 5 \times 10^{-3}, 10^{-3}, 5 \times 10^{-4}, 10^{-4}, 5 \times 10^{-5}, 10^{-5}, 0.0\}$ by selecting the parameter resulting in the lowest validation loss. Early stop is used with a patience of 5 epochs based on the validation loss. We also applied a learning rate schedule, which reduces the learning rate by 0.5 after five epochs of non-improvement to the validation loss.

When the outcome prediction is a regression task, the $\alpha$ and $\beta$ in \methodname's training objective are set to the root mean square of the outcomes in the training data to ensure a similar magnitude between regression and classification loss. The $\alpha$ and $\beta$ are set to 1 when the outcome prediction is a classification task. Experiments were conducted on a server with 256 CPU cores and 8 NVIDIA RTX A6000 GPUs for training and evaluation.

\section{Experiments Details}\label{sec:appendix:exp-details}
\subsection{Data Access}\label{sec:appendix:exp-details:data}
We accessed the IHDP dataset via the NPCI open-source package (https://github.com/vdorie/npci). Specifically, the IHDP data can be located under npci/examples/ihdp\_sim
/data/ihdp.RData in the NPCI GitHub repository. 
The data were accessed from the latest version of the repository, updated on Mar 16, 2022. No license is associated with this repository. 

The AMR-UTI data was accessed via PhysioNet (https://physionet.org/content/antimicrobial-resistance-uti/1.0.0/). We used the 1.0.0 version of AMR-UTI data, which is shared under the PhysioNet Credentialed Health Data License 1.5.0 (https://physionet.org/content/antimicrobial-resistance-uti/view-license/1.0.0/). During our research, we followed the data use agreement required by AMR-UTI, PhysioNet Credentialed Health Data Use Agreement 1.5.0 (https://physionet.org/content/antimicrobial-resistance-uti/view-dua/1.0.0/). 

\section{Additional Experimental Results}\label{sec:appendix:add-exp}

\subsection{Sensitivity Analysis with Synthetic Datasets}

\begin{figure*}[t]
     \centering
     \subfigure[Synthetic Dataset A]{\includegraphics[width=0.48\linewidth,alt={synthetic_a_appendix_p50}]{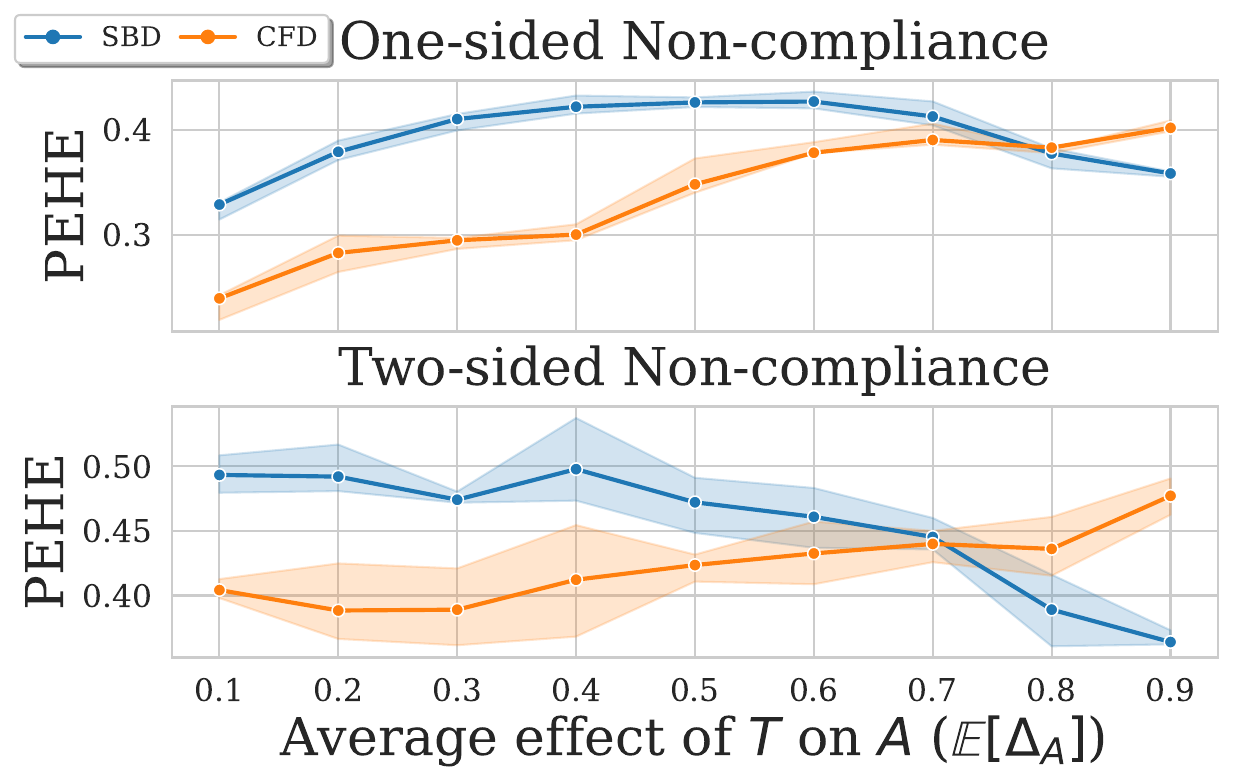}}\label{fig:sim-p-50:sim_a}
    \subfigure[Synthetic Dataset B]{\includegraphics[width=0.48\linewidth,alt={synthetic_b_appendix_p50}]{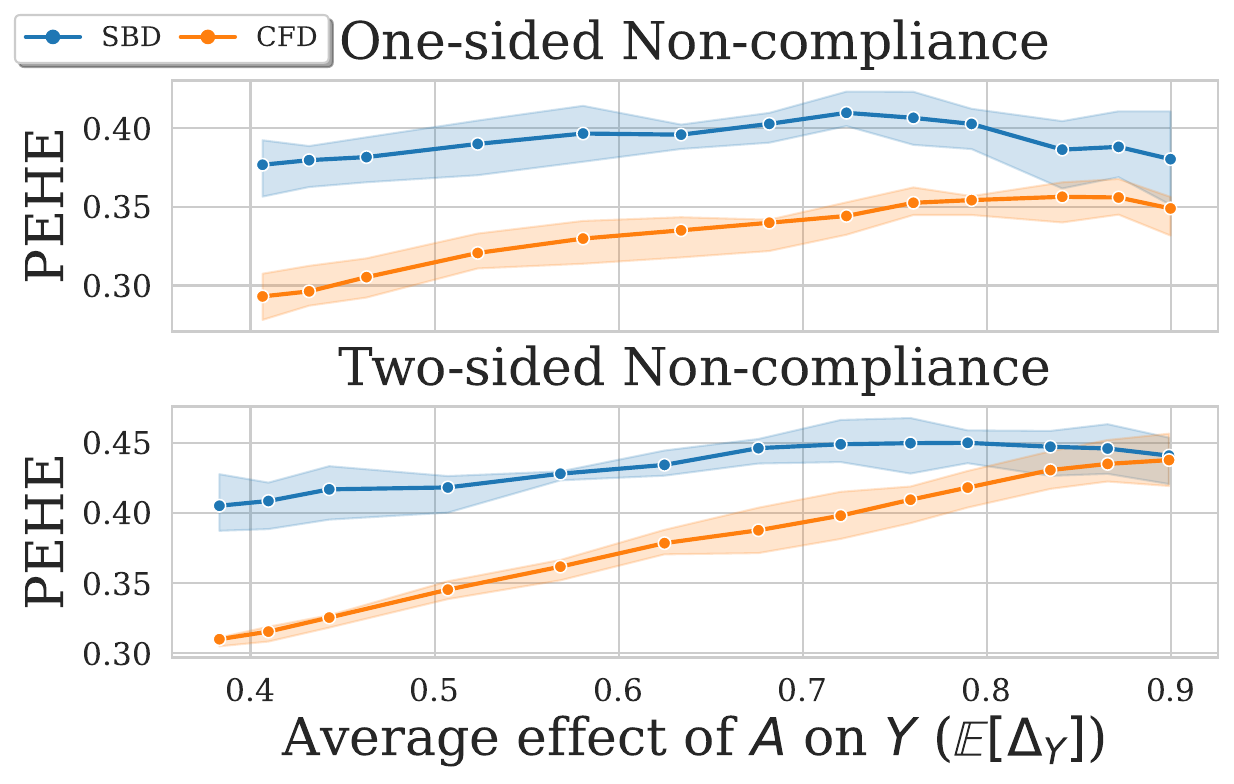}}\label{fig:sim-p-50:sim_b}
    \caption{CATEA estimation PEHE achieved by SBD and CFD on synthetic datasets with the number of features set to 50 instead of 30.}
    \label{fig:sim-p-50}
\end{figure*}

\begin{figure*}[t]
     \centering
     \subfigure[Synthetic Dataset A]{\includegraphics[width=0.48\linewidth,alt={synthetic_a_sppendix_amp5}]{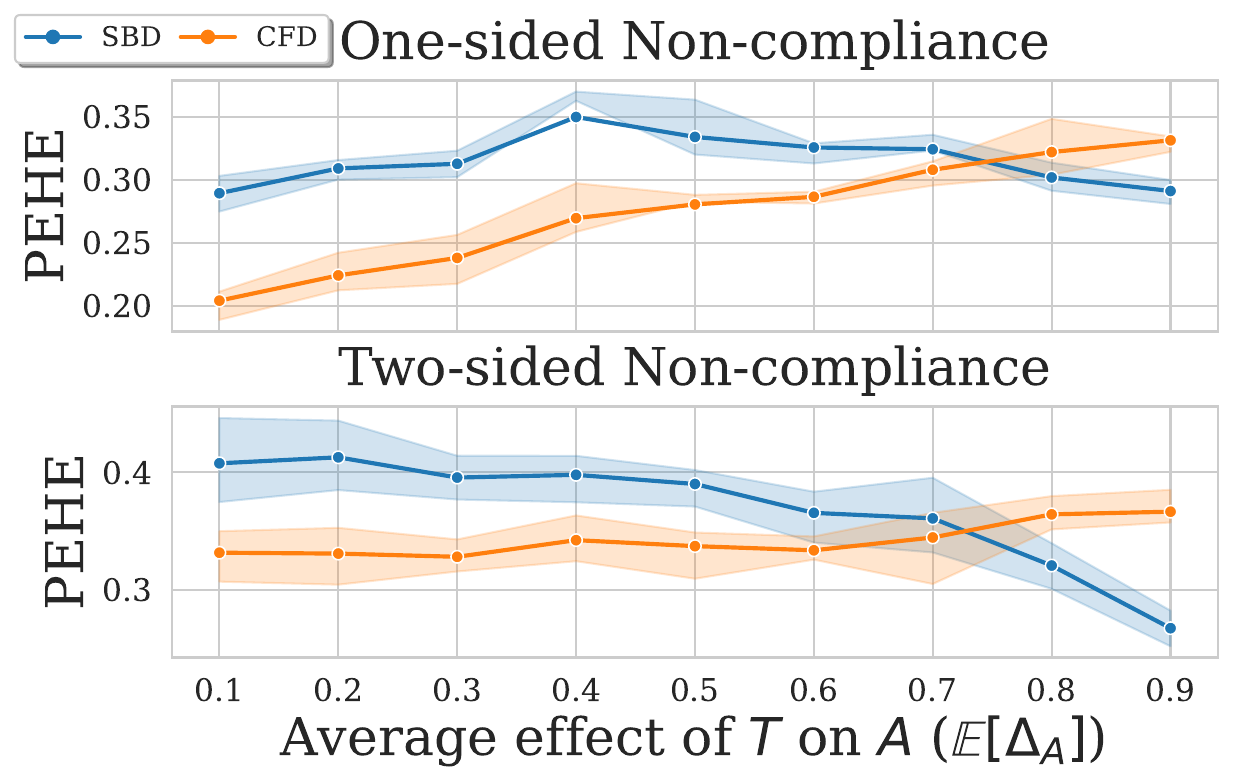}}\label{fig:sim-amp-5:sim_a}
    \subfigure[Synthetic Dataset B]{\includegraphics[width=0.48\linewidth,alt={synthetic_b_sppendix_amp5}]{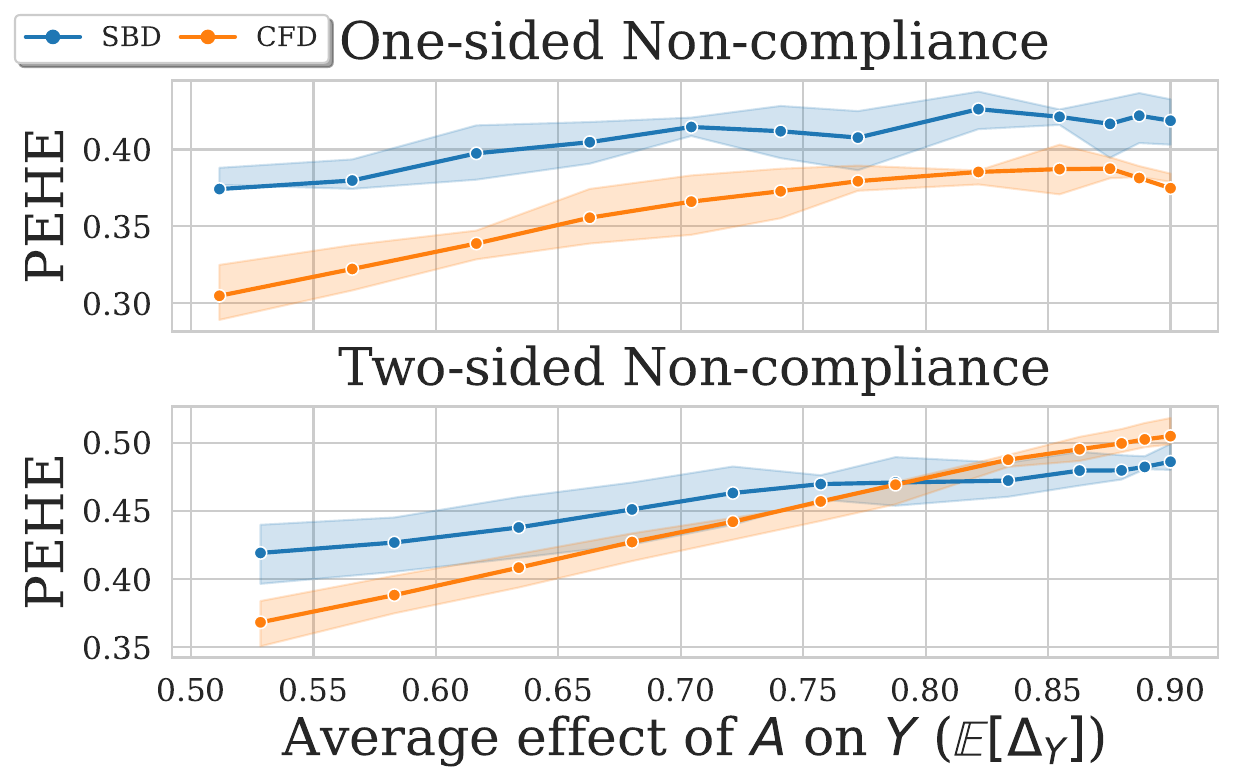}}\label{fig:sim-amp-5:sim_b}
    \caption{CATEA estimation PEHE achieved by SBD and CFD on synthetic datasets with random weights' sampling distribution changes to  $\unif(-5, 5)$ instead of $\unif(-10, 10)$.}
    \label{fig:sim-amp-5}
\end{figure*}

We repeated our experiments with different simulation parameters to assess the sensitivity of our synthetic experimental results (Section~\ref{sec:experiments:cfd-sbd-comp}). 
In particular, we first modified the number of features from 30 to 50. 
As shown in Figure~\ref{fig:sim-p-50}, the overall trend in both Synthetic Dataset A and Synthetic Dataset B remains the same as Section~\ref{sec:experiments:cfd-sbd-comp}: CFD is more advantageous than SBD when effect sizes are smaller.
We also modified the random weights' sampling parameter from $\unif(-10, 10)$ to $\unif(-5, 5)$ and repeated the experiments. 
As shown in Figure~\ref{fig:sim-amp-5}, the overall trend remains the same, CFD is advantageous over SBD when effect sizes are smaller.

\subsection{DragonNet vs. LobsterNet}
\begin{figure*}[t]
     \centering
     \subfigure[IHDP]{\includegraphics[width=0.48\linewidth,alt={ihdp_dragon_lobster}]{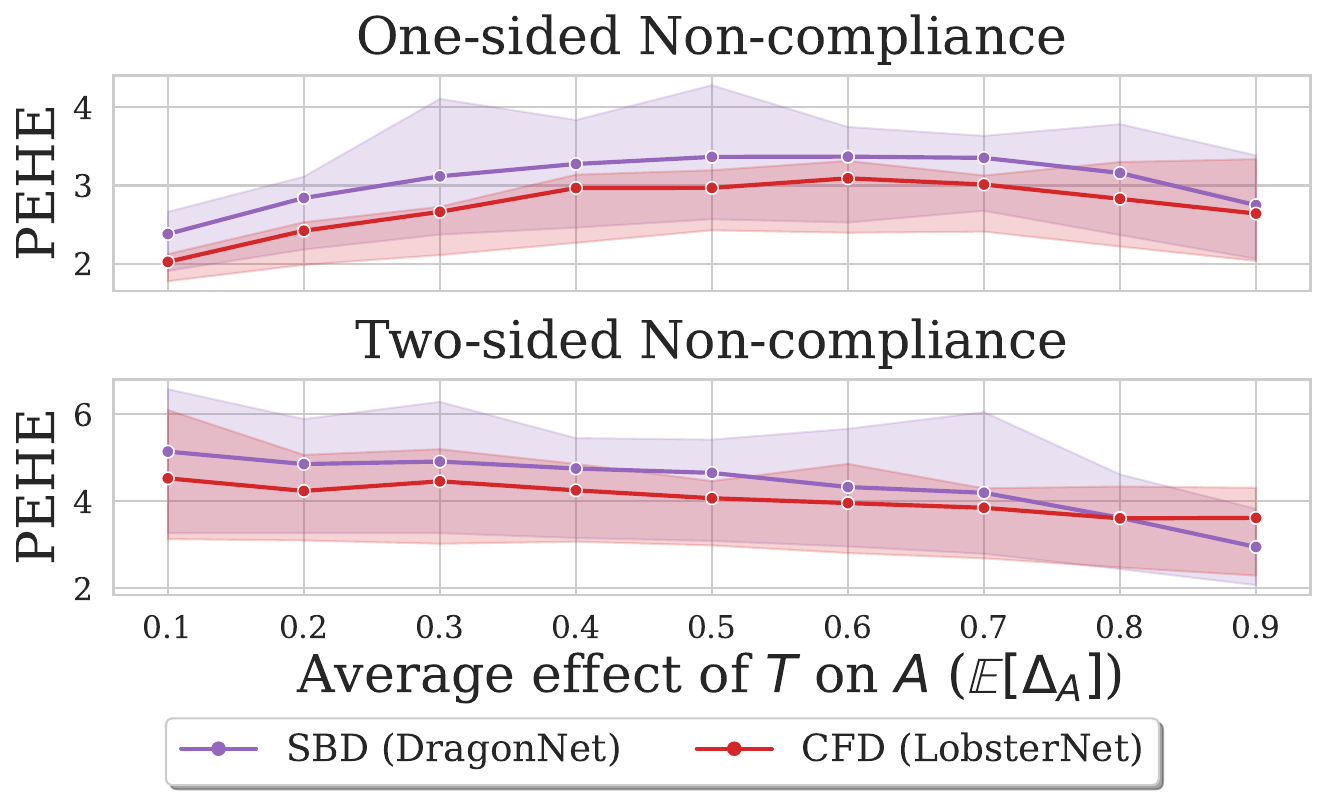}}\label{fig:real-dragon-lobster:ihdp_a}
    \subfigure[AMR-UTI]{\includegraphics[width=0.48\linewidth,alt={amruti_dragon_lobster}]{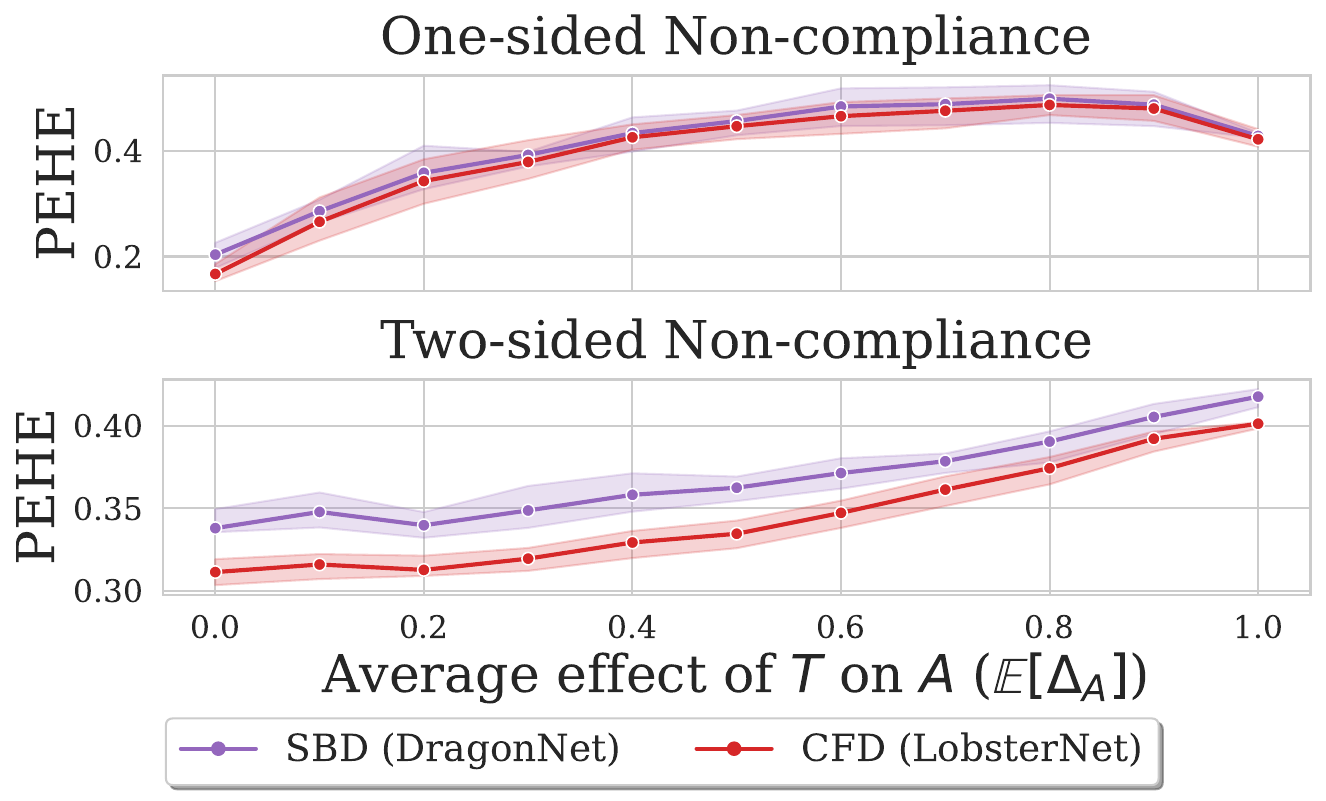}}\label{fig:real-dragon-lobster:amruti}
    \caption{CATEA estimation PEHE achieved by DragonNet and \methodname\ on IHDP and AMR-UTI datasets. 
    Across the majority of the settings, \methodname\ either matches or outperforms DragonNet.}
    \label{fig:real-dragon-lobster}
\end{figure*}

We additionally compare DragonNet~\citep{shi2019adapting} against our proposed \methodname\ using IHDP and AMR-UTI. 
As shown in Figure~\ref{fig:real-dragon-lobster}, \methodname\ either matches or outperforms DragonNet in terms of CATAE estimation PEHE, despite both methods jointly modeling all of their nuisance parameters, illustrating the practical advantage of CFD over SBD.

\subsection{Relative PEHE Improvement on IHDP and AMR-UTI}\label{sec:appendix:relative-improvement}
\begin{figure*}[t]
     \centering
     \subfigure[IHDP]{\includegraphics[width=0.48\linewidth,alt={ihdp_appendix_improve}]{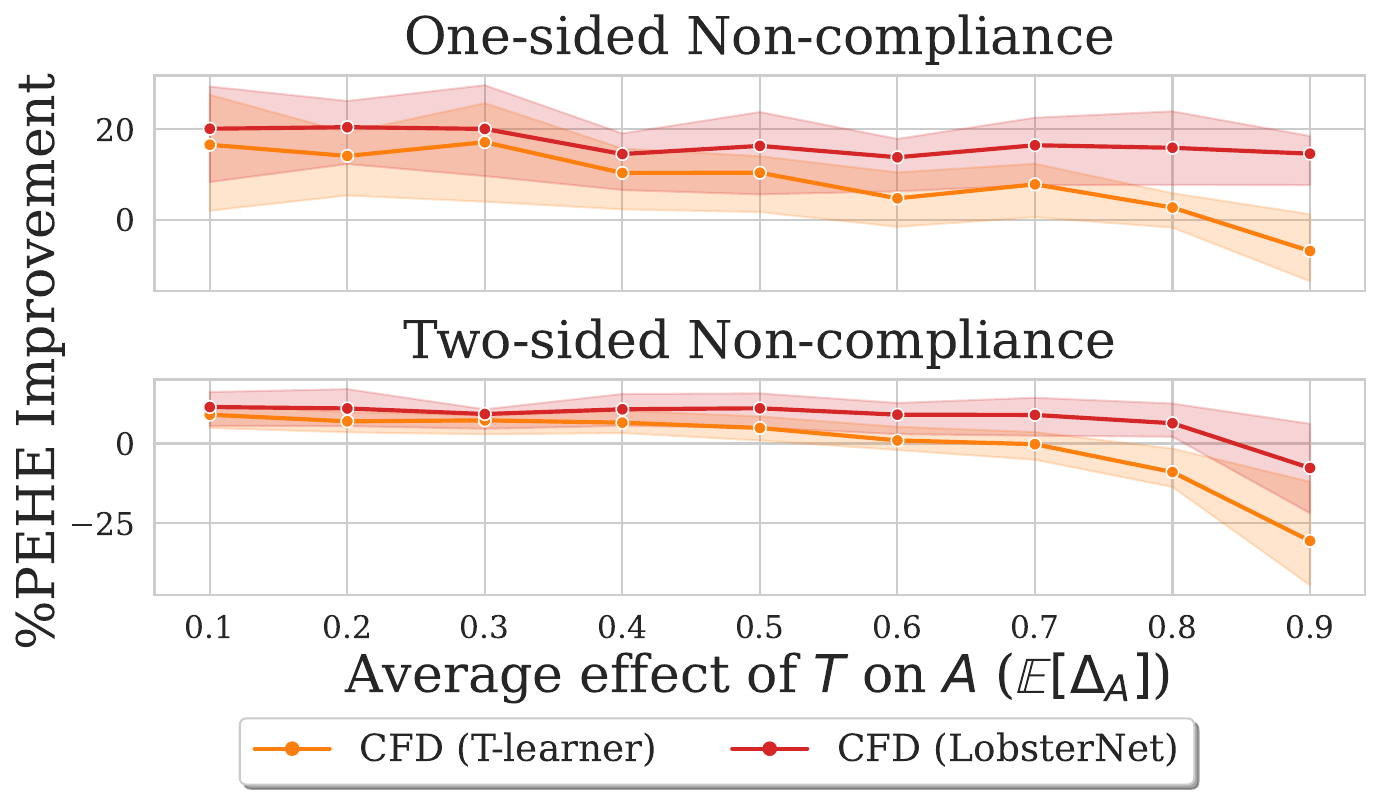}}\label{fig:real-improve:ihdp_a}
    \subfigure[AMR-UTI]{\includegraphics[width=0.48\linewidth,alt={amruti_appendix_improve}]{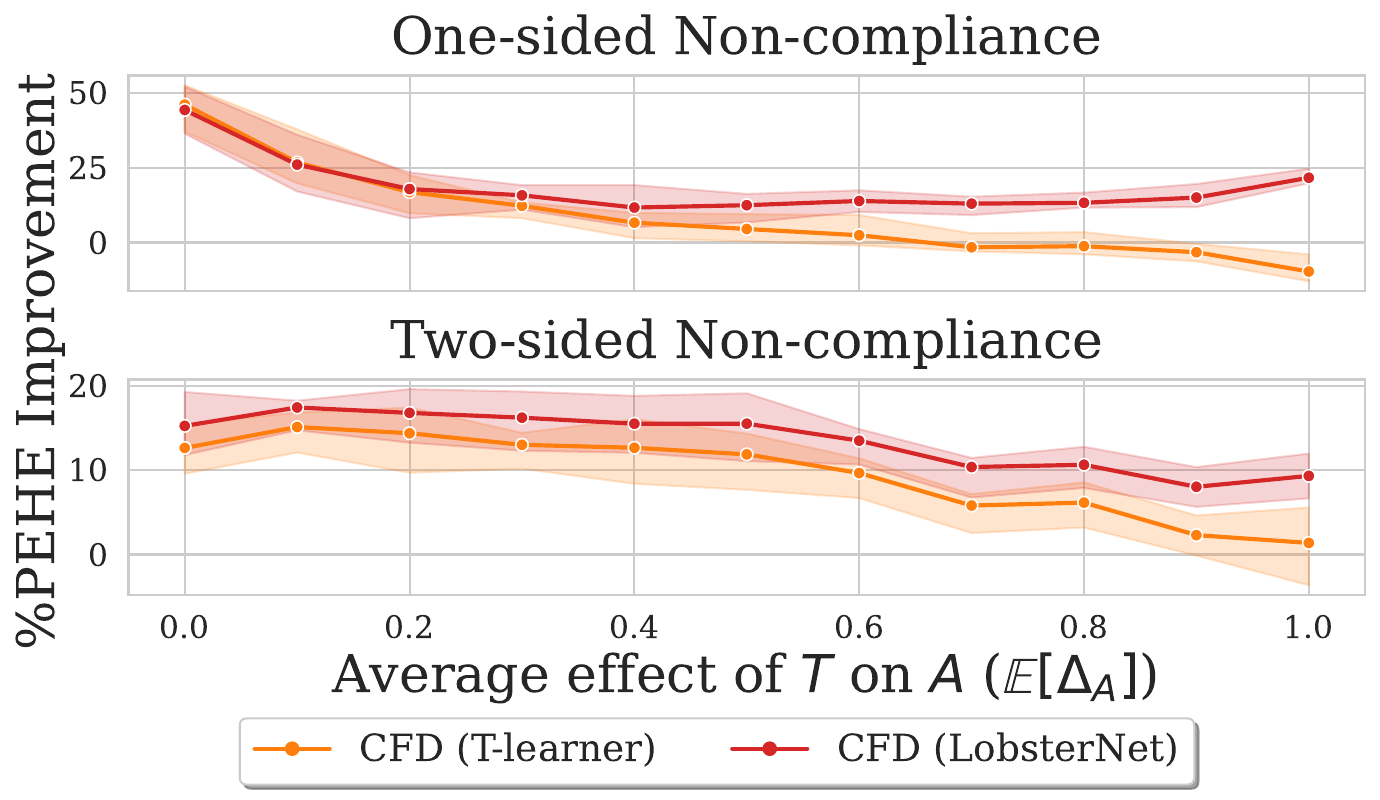}}\label{fig:real-improve:amruti}
    \caption{Relative CATEA estimation PEHE improvement achieved by CFD methods compared to the SBD method on the IHDP and AMR-UTI datasets. 
    Across the majority of the settings, CFD methods achieve a significant amount of improvements.}
    \label{fig:real-improve}
\end{figure*}

We also evaluated CFD methods in terms of their relative performance improvements compared to the SBD method.
We measure the relative improvement with $\frac{\text{PEHE}_{\text{SBD}}-\text{PEHE}_{\text{CFD}}}{\text{PEHE}_{\text{SBD}}} \times 100$.
As shown in Figure~\ref{fig:real-improve}, across most settings, CFD methods deliver significant improvements in PEHE.
In particular, under one-sided non-adherence, \methodname consistently delivers close to 20\% PEHE reduction across all settings.

\end{appendix}
\end{document}